\documentclass[acmtog]{acmart}
\usepackage{amsmath}
\usepackage{caption}
\usepackage{subcaption}
\usepackage{graphicx}
\usepackage{enumitem}
\usepackage{multirow}

\definecolor{Qing}{rgb}{0,0,0}
\newcommand{\qing}[1]{{\color{Qing} {#1}}}

\definecolor{zrj}{rgb}{0,0,0}
\newcommand{\zrj}[1]{{\color{zrj} {#1}}}
\AtBeginDocument{%
  }

\begin{document}

\title{Generative Texture Filtering}



\author{Rongjia Zheng}
\affiliation{%
  \institution{Sun Yat-sen University}
  \city{Guangzhou}
  \country{China}}
\email{zhengrj23@mail2.sysu.edu.cn}

\author{Shangwei Huang}
\affiliation{%
  \institution{Sun Yat-sen University}
  \city{Guangzhou}
  \country{China}}
\email{huangshw36@mail2.sysu.edu.cn}

\author{Lei Zhu}
\affiliation{%
  \institution{Hong Kong University of Science and Technology (Guangzhou)}
  \city{Guangzhou}
  \country{China}}
\email{leizhu@hkust-gz.edu.cn}

\author{Wei-Shi Zheng}
\affiliation{%
  \institution{Sun Yat-sen University}
  \city{Guangzhou}
  \country{China}}
\email{wszheng@ieee.org}

\author{Qing Zhang}
\authornote{Corresponding author.}
\affiliation{%
  \institution{Sun Yat-sen University}
  \city{Guangzhou}
  \country{China}}
\email{zhangqing.whu.cs@gmail.com}







\begin{abstract}
We present a generative method for texture filtering, which exhibits surprisingly good performance and generalizability. Our core idea is to empower texture filtering by taking full advantage of the strong learned image prior of pre-trained generative models. To this end, we propose to fine-tune a pre-trained generative model via a two-stage strategy. Specifically, we first conduct supervised fine-tuning on a very small set of paired images, and then perform reinforcement fine-tuning on a large-scale unlabeled dataset under the guidance of a reward function that quantifies the quality of texture removal and structure preservation. Extensive experiments show that our method clearly outperforms previous methods, and is effective to deal with previously challenging cases. \zrj{Our code is available at \href{https://github.com/OnlyZZZZ/Generative_Texture_Filtering}{https://github.com/OnlyZZZZ/Generative\_Texture\_Filtering}.}
\end{abstract}



\begin{teaserfigure}
    \centering
    \includegraphics[width=0.196\linewidth]{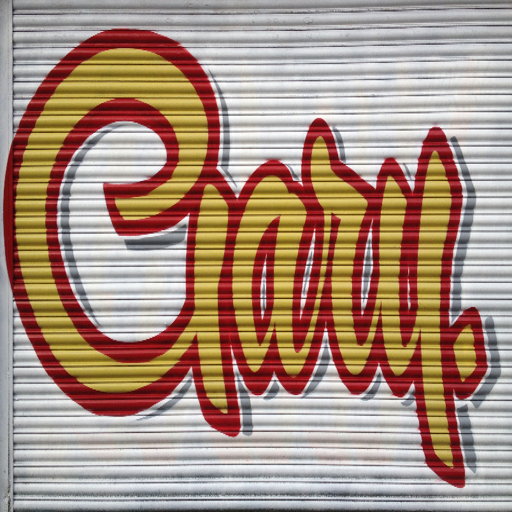}
    \includegraphics[width=0.196\linewidth]{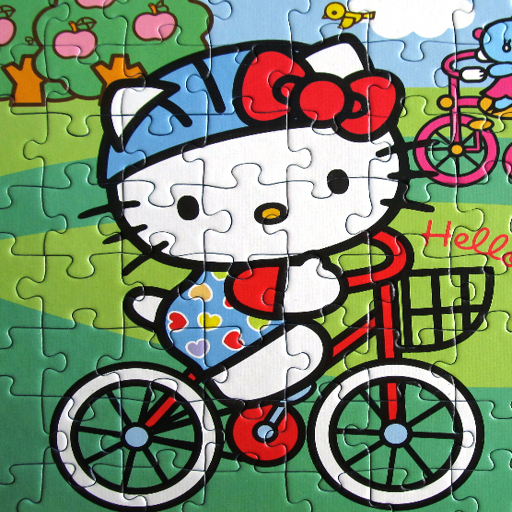}
    \includegraphics[width=0.196\linewidth]{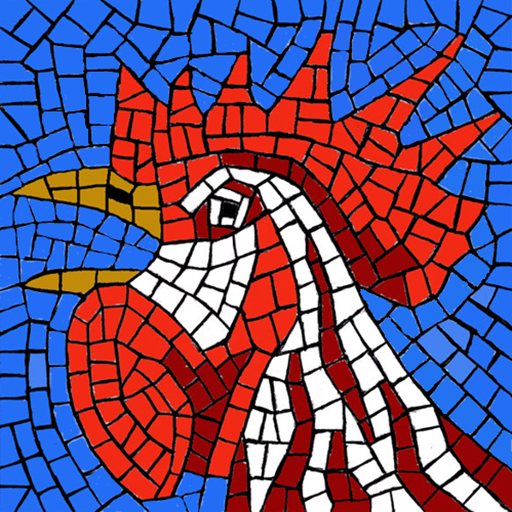}
    \includegraphics[width=0.196\linewidth]{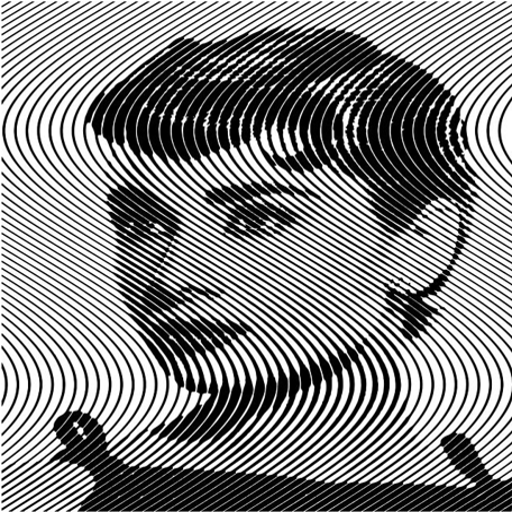} 
    \includegraphics[width=0.196\linewidth]{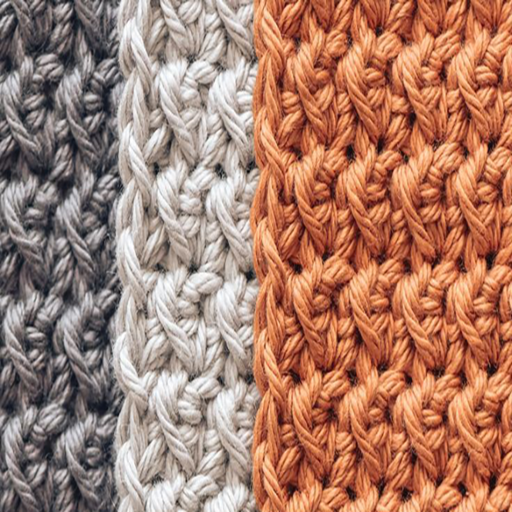} \vspace{1pt} \\
    \includegraphics[width=0.196\linewidth]{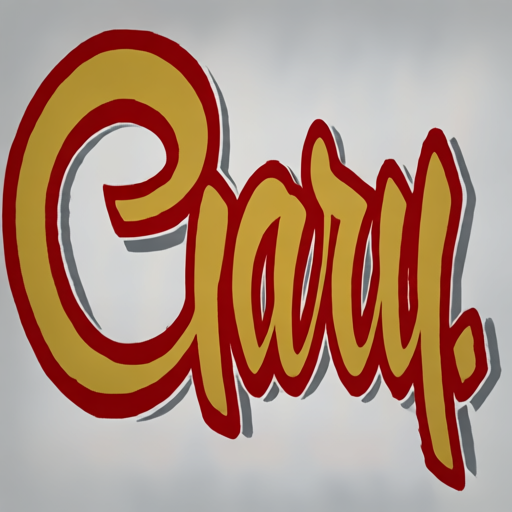}
    \includegraphics[width=0.196\linewidth]{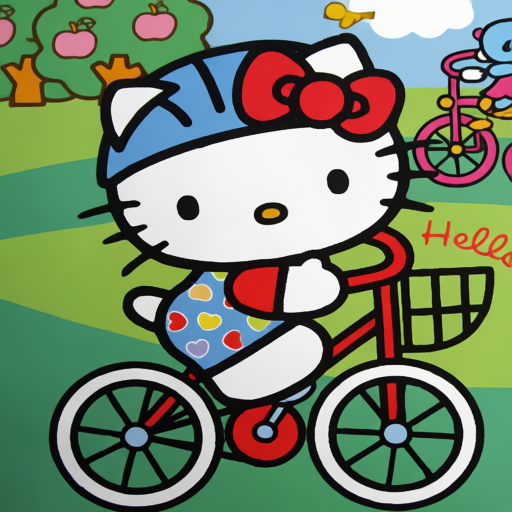}
    \includegraphics[width=0.196\linewidth]{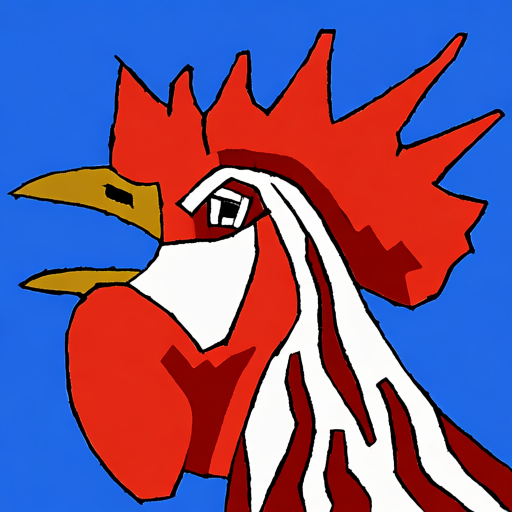}
    \includegraphics[width=0.196\linewidth]{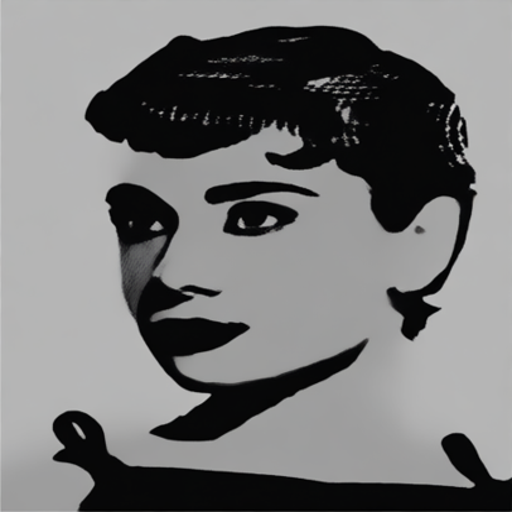} 
    \includegraphics[width=0.196\linewidth]{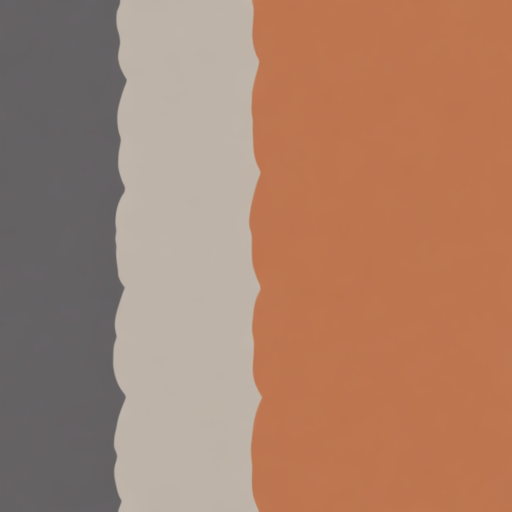}
    \vspace{-3mm}
    \caption{\qing{We achieve generative texture filtering with strong performance and generalization ability by fine-tuning a pre-trained generative model. Top and bottom are the input images and our texture filtering results, respectively.}}
    \label{fig:teaser}
\end{teaserfigure}
\maketitle

\begin{figure*}[!t]
    \captionsetup[subfigure]{labelformat=empty}
	\centering
	\begin{subfigure}[c]{0.161\textwidth}
		\centering
        \includegraphics[width=\linewidth]{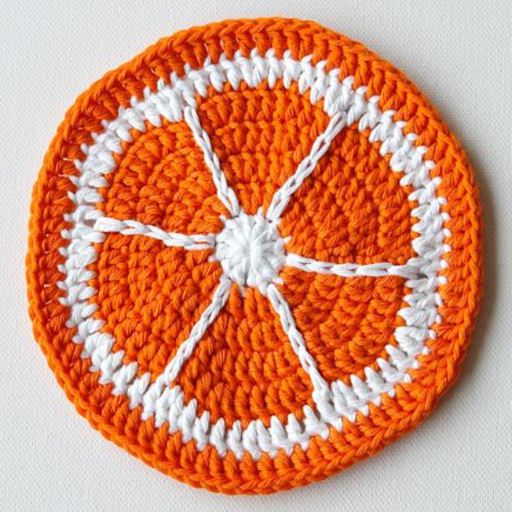} \\
        \vspace{-2mm}
		\caption{Input}
	\end{subfigure}
    \begin{subfigure}[c]{0.161\textwidth}
		\centering
        \includegraphics[width=\linewidth]{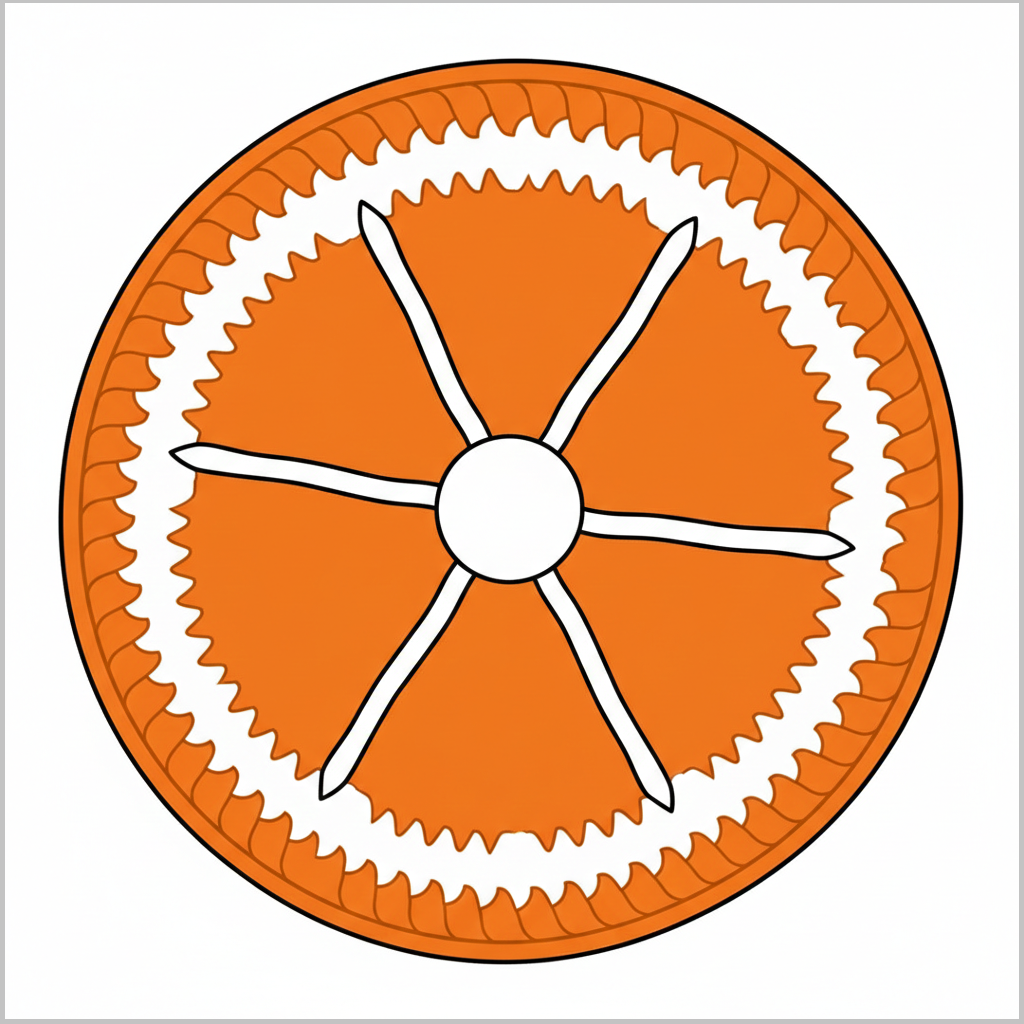} \\
        \vspace{-2mm}
		\caption{Nano Banana}
	\end{subfigure}
    \begin{subfigure}[c]{0.161\textwidth}
		\centering
        \includegraphics[width=\linewidth]{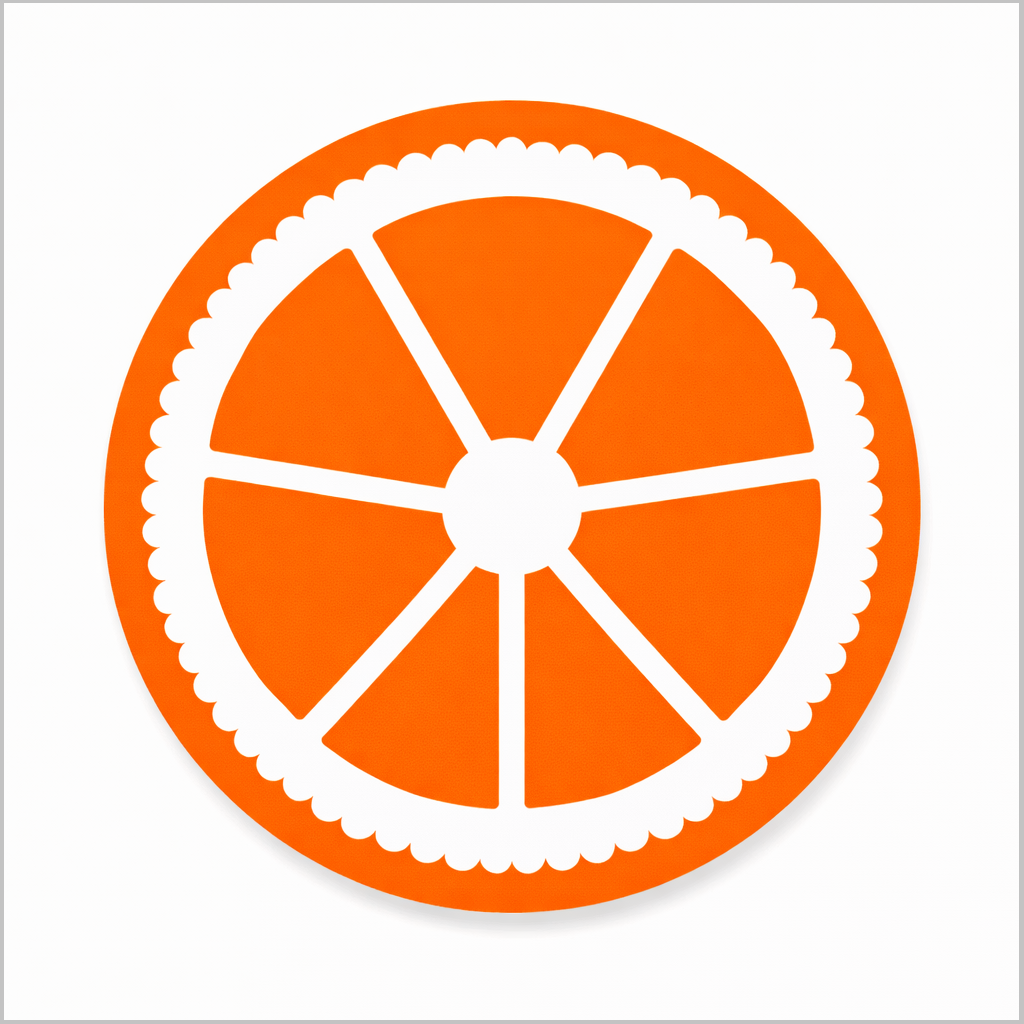} \\
        \vspace{-2mm}
		\caption{GPT-4o}
	\end{subfigure}
    \begin{subfigure}[c]{0.161\textwidth}
		\centering
        \includegraphics[width=\linewidth]{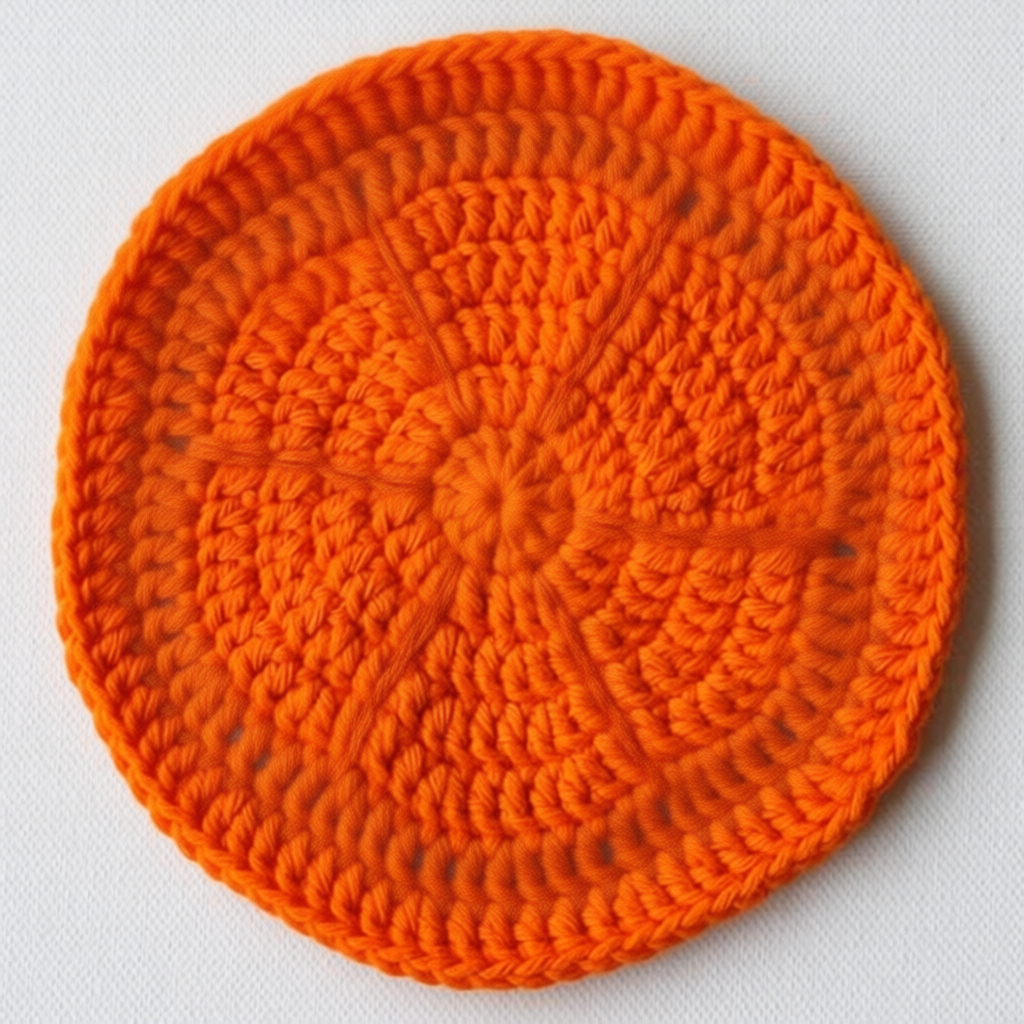} \\
        \vspace{-2mm}
		\caption{FLUX.1-Kontext}
	\end{subfigure}
    \begin{subfigure}[c]{0.161\textwidth}
		\centering
        \includegraphics[width=\linewidth]{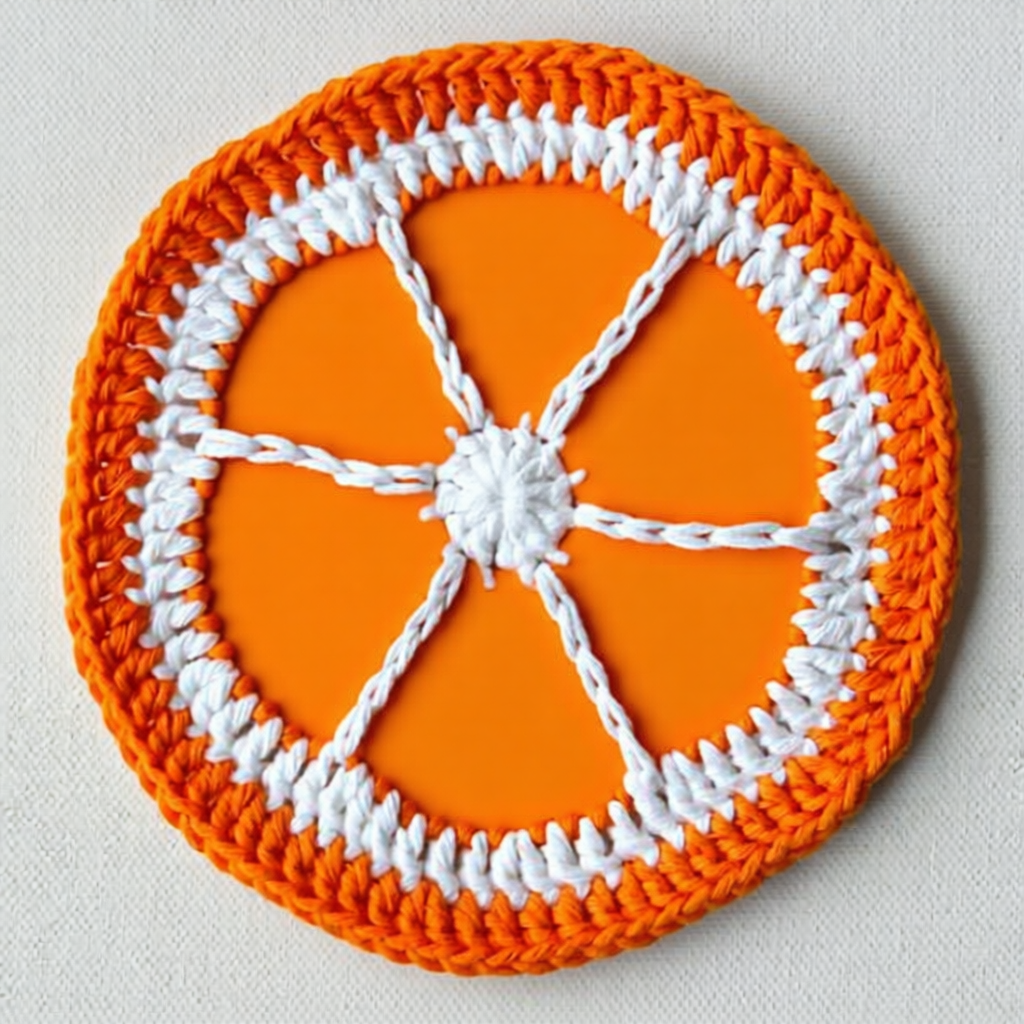} \\
        \vspace{-2mm}
		\caption{Qwen-Image-Edit}
	\end{subfigure}
    \begin{subfigure}[c]{0.161\textwidth}
		\centering
        \includegraphics[width=\linewidth]{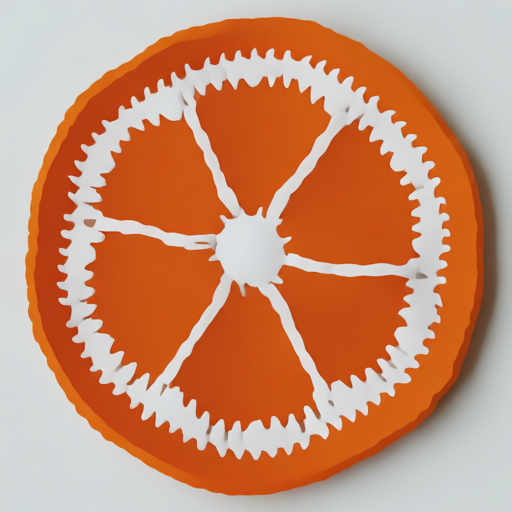} \\
        \vspace{-2mm}
		\caption{Ours}
	\end{subfigure}
    \vspace{-2mm}
\caption{\qing{Results of directly applying popular image generation models for texture filtering. For fair comparison, we produce all the results using the same text prompt: ``remove texture but preserve structure, keep color and structure faithful to the original image''. Note, we tried various other text prompts, but found that they did not work as well as the one we used.}}
\label{fig:big_model}
\end{figure*}

\begin{figure}[!t]
    \centering
    \includegraphics[width=1.0\linewidth]{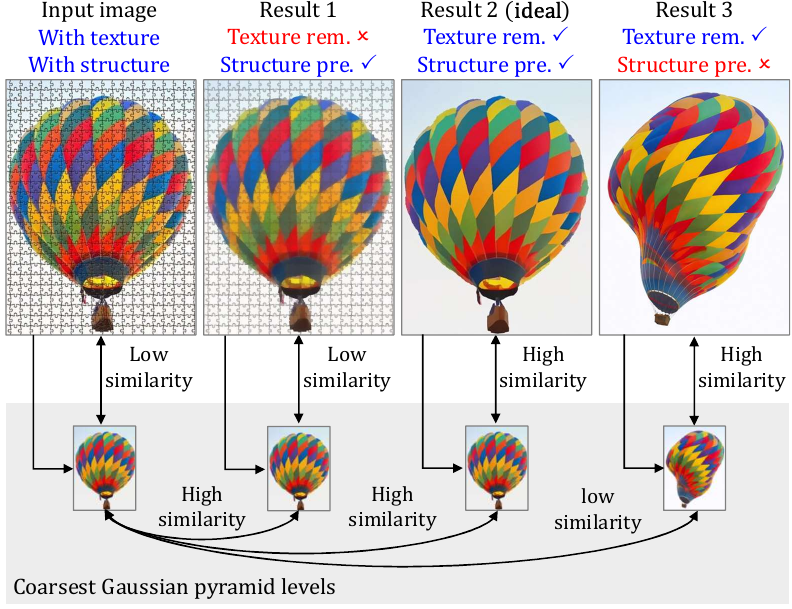}\\
    \vspace{-3mm}
    \caption{\qing{Visual Illustration of cues that inspire us to quantify the texture removal and structure preservation performance of a given texture filtering output. Texture rem. and structure pre. refer to texture removal and structure preservation, respectively. The four low-resolution images at the bottom are the  corresponding coarsest Gaussian pyramid levels of the top images. }}
    \label{fig:observation}
    \vspace{-2mm}
\end{figure}

\section{Introduction}
\qing{Texture filtering has long been a fundamental problem in computational photography and image processing. As shown in Figure~\ref{fig:teaser}, it aims to smooth out textures while preserving image structures. Because of allowing structure and texture separation, texture filtering finds various applications towards precise manipulation of structure and texture elements in an image, e.g., detail enhancement, image abstraction, and inverse halftoning. }

\qing{Despite remarkable progress over the past decades, current texture filtering methods \citep{xu2012structure-rtv,karacan2013structure,cho2014bilateral-btf,fan2018image,zhang2023pyramid,zhang2025stf} still suffer from the following limitations: 
\begin{itemize}[leftmargin=1em]
\setlength\itemsep{0.5em} 
    \item \textbf{Generalizability}: due to the lack of large-scale training datasets, existing learning-based methods often struggle to generalize to images with unseen textures. On the other hand, conventional non-learning-based methods basically require a tedious image-specific parameter tuning process to produce satisfactory results. 
    \item \textbf{Effectiveness}: as previous methods are mostly built upon observations on local image statistics or strong image assumptions, their effectiveness usually deteriorates significantly on images that beyond their representation expressiveness. 
\end{itemize}
}

\qing{We believe that the above limitations encountered by previous methods can be attributed to their insufficient ability to represent structure and texture. To address their limitations, we propose to embrace recent generative models, as they have demonstrated very powerful image generation capability, which is generally considered to require a holistic and accurate understanding of image representation as a foundation. However, as shown in Figure~\ref{fig:big_model}, while a naive application of popular generative models (e.g., \zrj{Nano banana~\cite{comanici2025gemini}, GPT-4o~\cite{hurst2024gpt},} Flux.1-Kontext \citep{labs2025flux}, and Qwen-Image-Edit~\cite{wu2025qwenimagetechnicalreport}) shows considerable potential in texture filtering, they fail to generate high-quality results that are faithful to the original image in terms of color and structure preservation because the models are not taught to generate outputs strictly aligned with the task of texture filtering. 
}

\qing{In this work, we propose to formulate texture filtering as an image generation problem. Following this perspective, we develop a texture filtering method with superior performance as well as strong generalizability by fine-tuning an open-source pre-trained generative model, Qwen-Image-Edit. To enable more stable, efficient, and effective learning of texture filtering, we introduce a two-stage fine-tuning strategy, where we first conduct supervised fine-tuning on a very small set of paired images to provide a stable starting point for texture filtering, and then perform large-scale reinforcement fine-tuning on unlabeled data to further refine the performance. To do so, we devise a reward function for guiding the reinforcement fine-tuning, which is able to quantify the quality of the texture filtering output from two aspects: texture removal and structure preservation.

We define the reward function based on the observation utilized in \citep{zhang2023pyramid}, i.e., textures in an image will be gradually eliminated along with the Gaussian pyramid construction, eventually resulting in a coarsest pyramid level that is nearly texture-free while summarizing the image structures. As shown in Figure~\ref{fig:observation}, this observation provides two important cues to evaluate the texture removal and structure preservation performance of a texture filtering result: (i) if the result still contains texture, the similarity between its coarsest Gaussian pyramid level and the result itself will be low due to the inherent loss of textures during pyramid construction, while on the contrary, if no texture contained in the result, the similarity between the two should be relatively high. That is, higher similarity between the texture filtering result and its coarsest Gaussian pyramid level generally corresponds to better texture removal. (ii) as the coarsest Gaussian pyramid level summarizes the image structures, the similarity between the coarsest levels of the texture filtering result and the original image can be an indicator for performance of structure preservation, namely higher similarity means more effective preservation of structure.

\begin{figure*}[!t]
    \centering
    \includegraphics[width=1.0\textwidth]{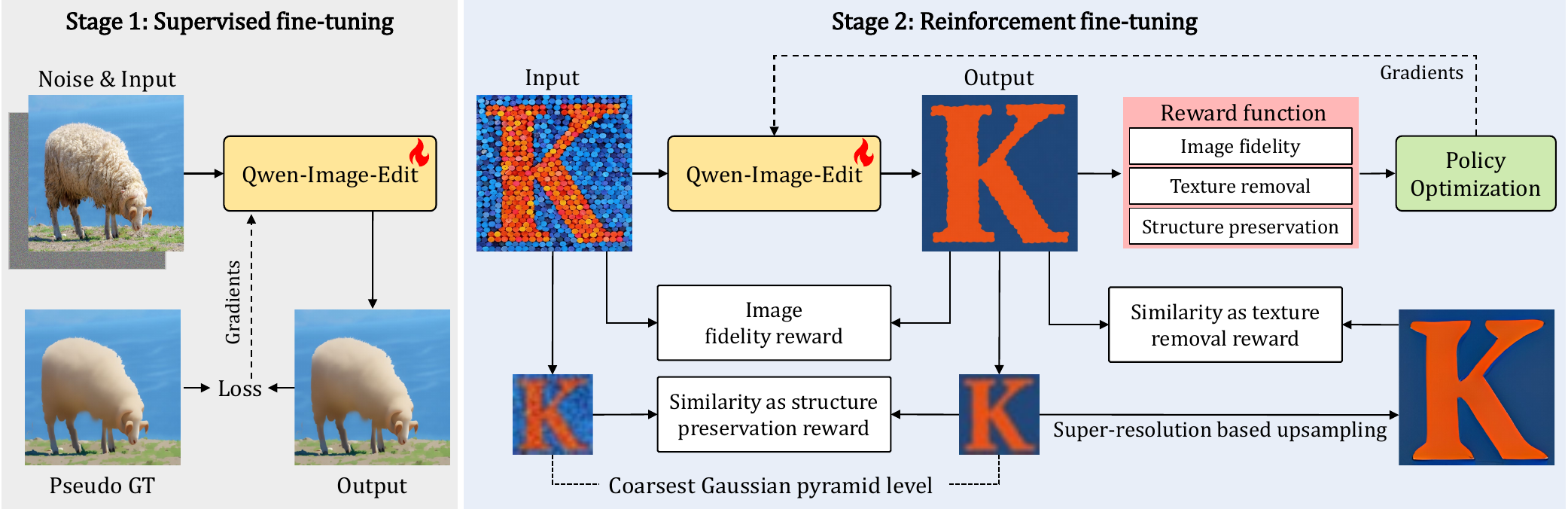}\\
    \vspace{-4mm}
    \caption{Overview of our generative texture filtering framework.}
    \label{fig:overview}
    \vspace{-2mm}
\end{figure*}

To enable large-scale reinforcement fine-tuning, we employ a text-to-image model to synthesize a total of 10,000 images that cover a very diverse types of texture. Experiments on both synthetic and real-world data demonstrate that our method outperforms previous methods, and exhibits strong performance and generalizability. In summary, our work makes the following contributions:
\vspace{1mm}
\begin{itemize}[leftmargin=1em]
\setlength\itemsep{0.5em} 
    \item We propose to cast texture filtering as an image generation problem, and present the first generative method for texture filtering. 
    \item We introduce a two-stage fine-tuning strategy for adapting a pre-trained generative model as a texture filtering method with strong performance and generalizability.
    \item We develop a reward function that can quantify the quality of the texture filtering output from the perspective of texture removal and structure preservation.   
\end{itemize}
}

\section{Related Work}
\qing{\textbf{Edge-aware filtering.} Existing methods can be roughly classified into three main categories: local filtering based methods, optimization-based methods, and learning-based methods. Among them, local filtering based methods work by performing weighted average over a local spatial neighborhood, with a large number of works falling into this category \citep{perona1990scale,tomasi1998bilateral,chen2007real,durand2002fast,weiss2006fast,paris2006fast,fattal2009edge,kass2010smoothed,aubry2014fast,paris2011local,criminisi2010geodesic,gastal2011domain,gastal2012adaptive}. Optimization-based methods, on the other hand, achieve edge-preserving filtering by solving a gradient-aware global linear system \cite{farbman2008edge,bi20151,liu2020real,xu2011image}. Finally, some learning-based methods propose to address the problem in a data-driven fashion \cite{xu2015deep,fan2018image,liu2016learning}. Although demonstrating impressive results, they are not suitable for texture filtering because of lacking the ability to classify structure and texture edges. }

\vspace{0.5em}
\noindent \textbf{Texture filtering.} As the main difficulty of this problem lies in separating structure from texture, early methods mostly focus on developing texture-structure separation measures, e.g., local extrema \citep{subr2009edge}, relative total variation \citep{xu2012structure-rtv}, region covariance \citep{karacan2013structure}, graph segment \citep{zhang2015segment}, and minimum spanning tree \citep{bao2013tree}. Instead of relying on hand-crafted measures, some later works are built upon the scale difference between texture and structure \citep{jeon2016scale,du2016two,zhang2023pyramid,zhang2014rolling}. There are also some deep learning based methods in the literature, 
where \citet{kim2018structure} propose to utilize deep variational priors while \citet{lu2018deep} achieve supervised learning of texture filtering by creating paired data through blending
textures with clean structure-only images. More recently, \citet{zhang2025stf} present a self-supervised texture filtering framework based on a texture-inversion observation. 

\section{Method}
\qing{We aim to achieve texture filtering from a generative perspective, by unleashing the powerful image priors learned by recent image generation foundation models. To this end, we introduce a two-stage strategy to fine-tune an open-source pre-trained generative model, Qwen-Image-Edit~\cite{wu2025qwenimagetechnicalreport}, as a strong texture filtering method. Specifically, we first perform supervised fine-tuning (SFT) on a small set of paired images to learn basic texture filtering ability, and then largely refine its effectiveness by conducting reinforcement fine-tuning (RFT) on large-scale unlabeled data, under the guidance of a reward function that can quantify the quality of texture removal and structure preservation of a texture filtering output. Figure~\ref{fig:overview} gives an overview of our approach. }

\subsection{Supervised Fine-tuning}

\begin{figure*}[!t]
	\centering
	\begin{subfigure}[c]{0.195\textwidth}
		\centering
        \includegraphics[width=\linewidth]{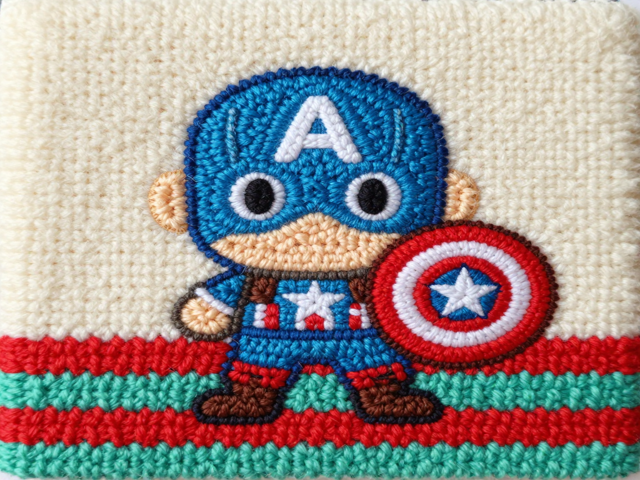}\\
        \vspace{-2mm}
		\caption*{Input}
	\end{subfigure}
    \begin{subfigure}[c]{0.195\textwidth}
		\centering
        \includegraphics[width=\linewidth]{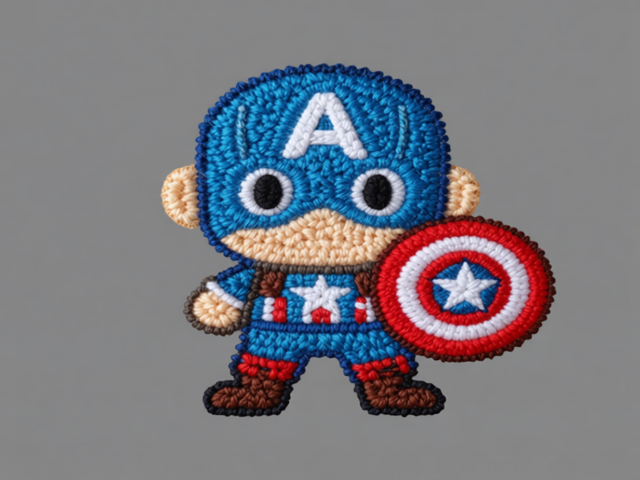}\\
        \vspace{-2mm}
		\caption*{Vanilla Qwen-Image-Edit}
	\end{subfigure}
    \begin{subfigure}[c]{0.195\textwidth}
		\centering
        \includegraphics[width=\linewidth]{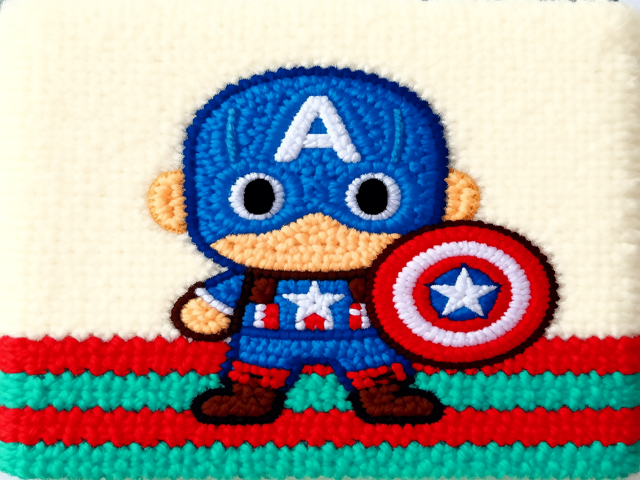}\\
        \vspace{-2mm}
		\caption*{With only SFT}
	\end{subfigure}
    \begin{subfigure}[c]{0.195\textwidth}
		\centering
        \includegraphics[width=\linewidth]{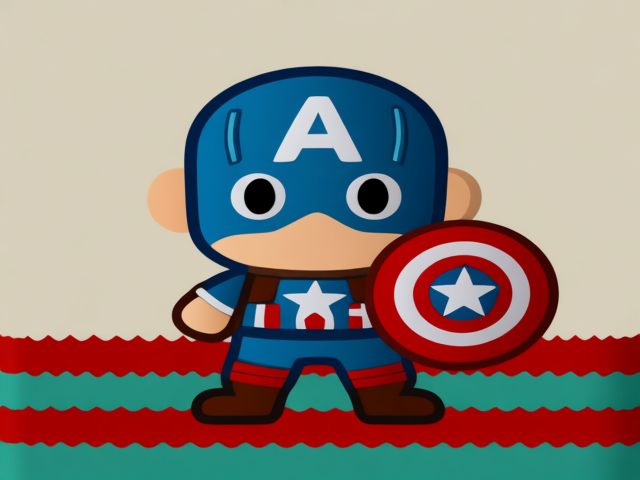}\\ 
        \vspace{-2mm}
		\caption*{With only RFT}
	\end{subfigure}
    \begin{subfigure}[c]{0.195\textwidth}
		\centering
        \includegraphics[width=\linewidth]{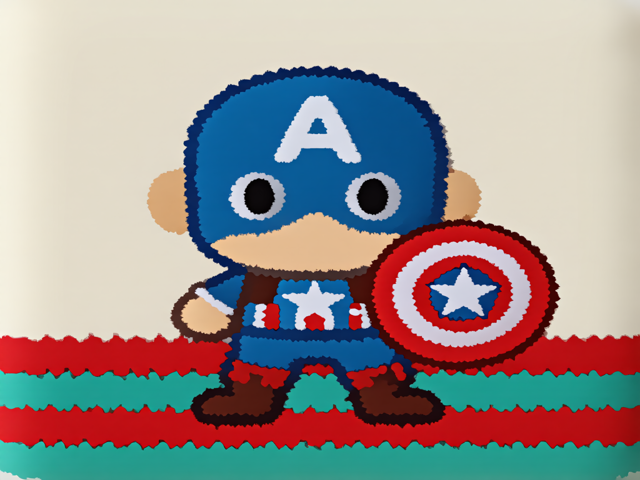}\\ 
        \vspace{-2mm}
		\caption*{Full method (SFT + RFT)}
	\end{subfigure}
    \vspace{-3mm}
	\caption{Effectiveness of our two-stage fine-tuning strategy. Note the texture residuals and distorted structures in the results of merely performing supervised fine-tuning (SFT) or reinforcement fine-tuning (RFT). In comparison, we produce better result by combining SFT and RFT.}
	\label{fig:two-stage-ablation}
    \vspace{-2mm}
\end{figure*}

\begin{figure}[!t]
    \centering
    \begin{subfigure}[c]{0.24\linewidth}
		\centering
		\includegraphics[width=\textwidth]{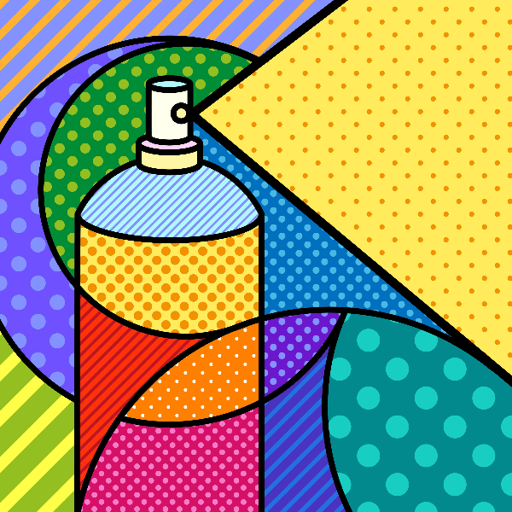} \\ \vspace{1pt}
        \includegraphics[width=\textwidth]{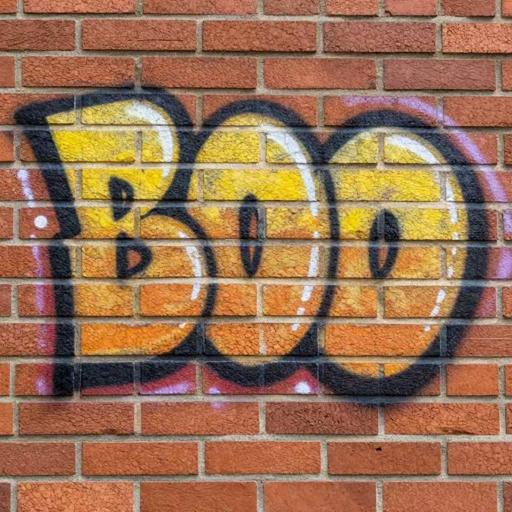} \\
    \end{subfigure}
    \begin{subfigure}[c]{0.24\linewidth}
		\centering
		\includegraphics[width=\textwidth]{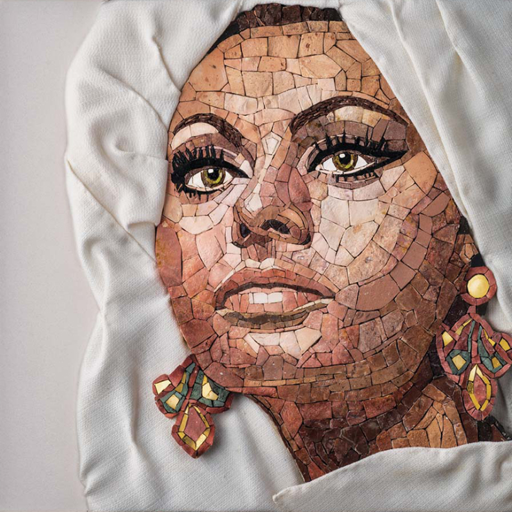} \\ \vspace{1pt}
        \includegraphics[width=\textwidth]{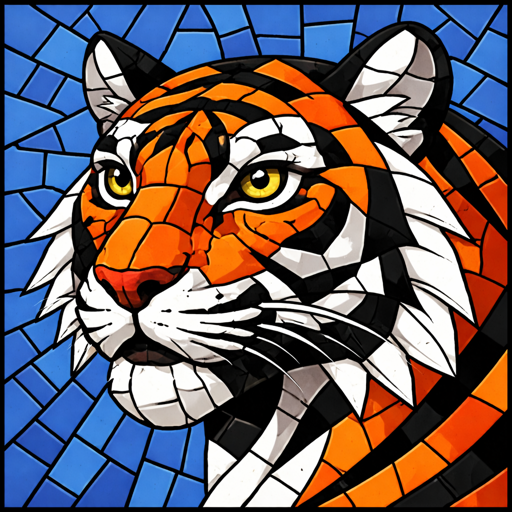} 
    \end{subfigure}
    \begin{subfigure}[c]{0.24\linewidth}
		\centering
		\includegraphics[width=\textwidth]{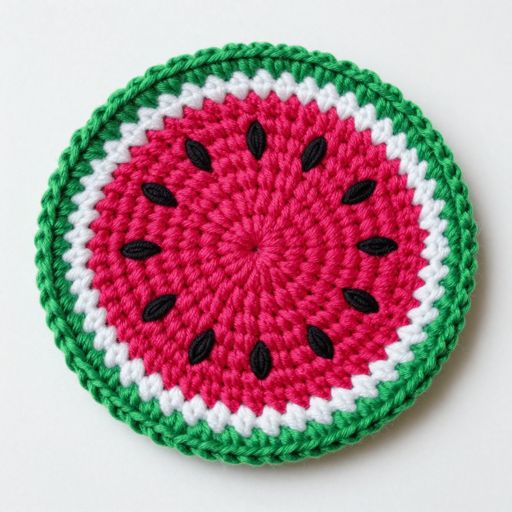} \\ \vspace{1pt}
        \includegraphics[width=\textwidth]{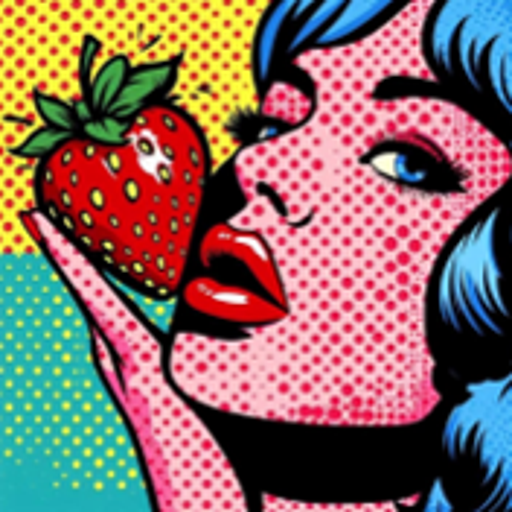} \\
    \end{subfigure}
    \begin{subfigure}[c]{0.24\linewidth}
		\centering
		\includegraphics[width=\textwidth]{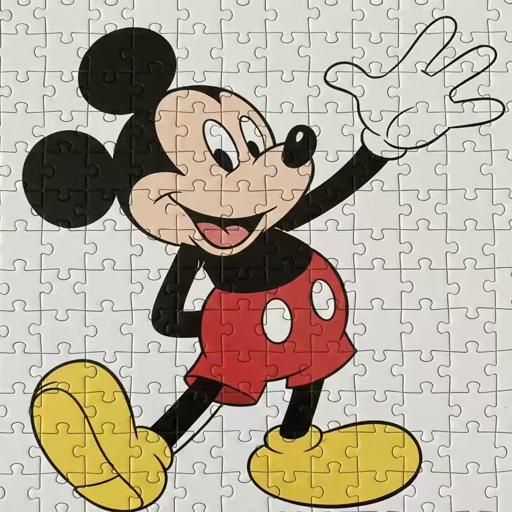} \\ \vspace{1pt}
        \includegraphics[width=\textwidth]{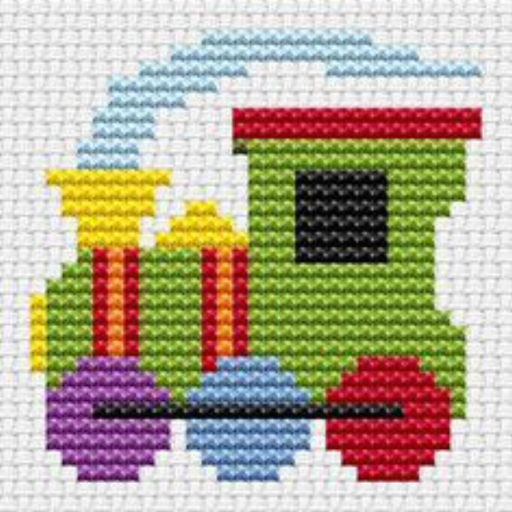} \\
    \end{subfigure}
    \vspace{-4mm}
    \caption{Some example images in our synthesized dataset.}
    \label{fig:train_dataset}
    \vspace{-3mm}
\end{figure}

\begin{figure*}[!t]
	\centering
	\begin{subfigure}[c]{0.195\textwidth}
		\centering
        \includegraphics[width=\linewidth]{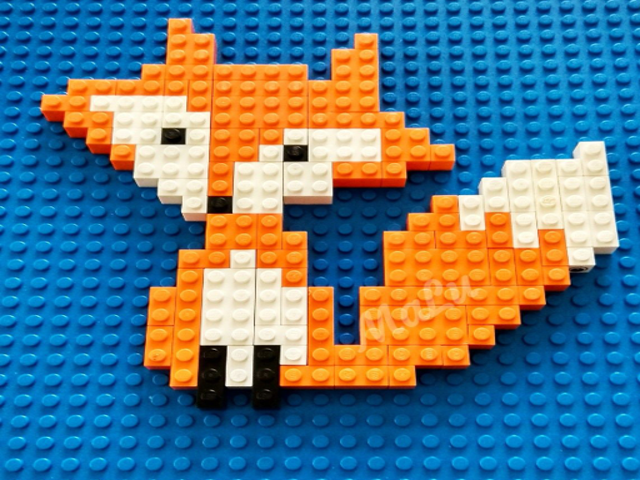}\\ 
        \vspace{-2mm}
		\caption*{Input}
	\end{subfigure}
    \begin{subfigure}[c]{0.195\textwidth}
		\centering
        \includegraphics[width=\linewidth]{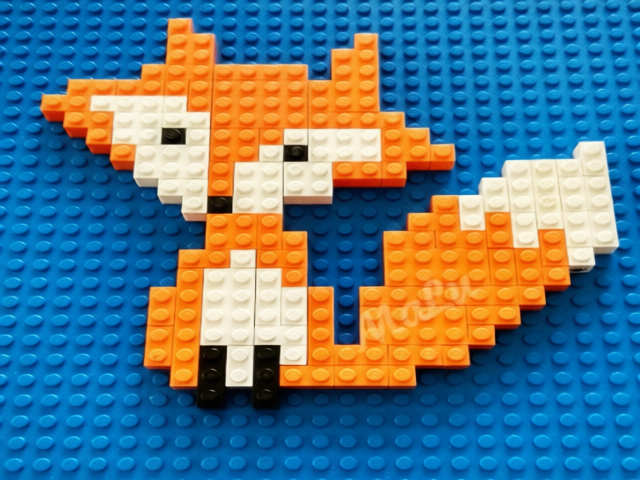}\\ 
        \vspace{-2mm}
		\caption*{Without texture removal}
	\end{subfigure}
    \begin{subfigure}[c]{0.195\textwidth}
		\centering
        \includegraphics[width=\linewidth]{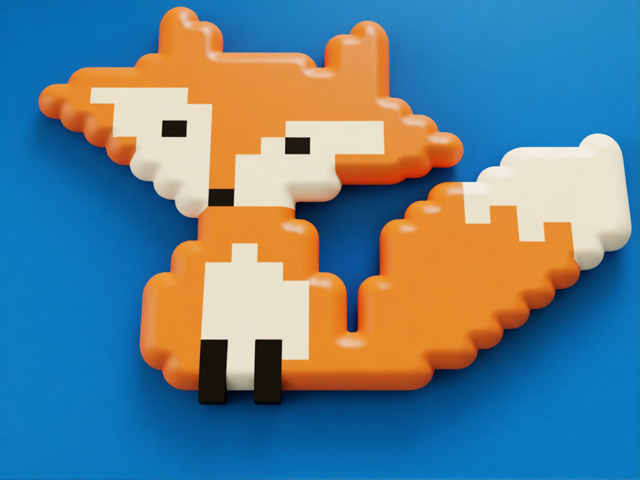} \\ 
        \vspace{-2mm}
		\caption*{Without structure preservation}
	\end{subfigure}
    \begin{subfigure}[c]{0.195\textwidth}
		\centering
        \includegraphics[width=\linewidth]{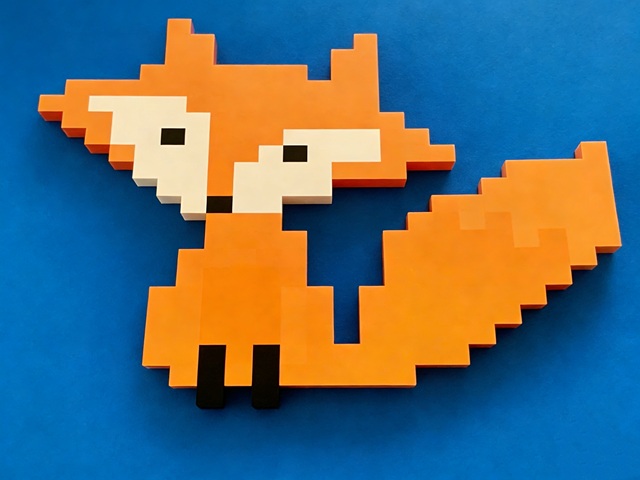}\\ 
        \vspace{-2mm}
		\caption*{Without image fidelity}
	\end{subfigure}
    \begin{subfigure}[c]{0.195\textwidth}
		\centering
        \includegraphics[width=\linewidth]{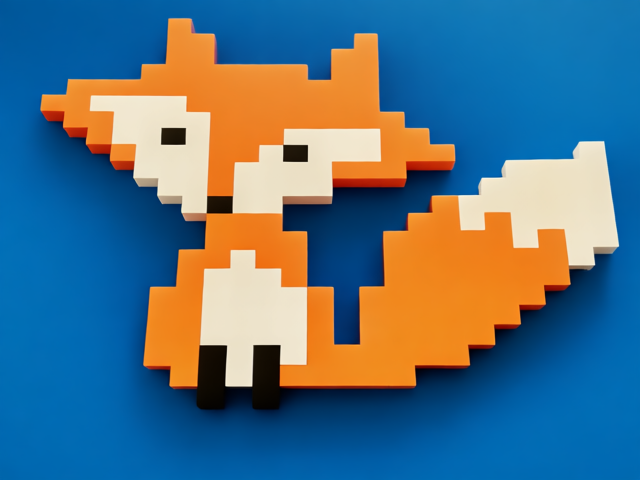}\\ 
        \vspace{-2mm}
		\caption*{Full method}
	\end{subfigure}
    \vspace{-3mm}
	\caption{Effectiveness of each component in our reward function. As shown, removing the texture removal reward would make the model to take a shortcut to directly generate the input image, while omitting the structure preservation reward produces result with distorted structures, and ignoring the image fidelity reward results in clear color and structure inconsistencies. }
	\label{fig:reward_function}
    \vspace{-1mm}
\end{figure*}

\begin{figure}[!t]
    \centering
    \includegraphics[width=1.0\linewidth]{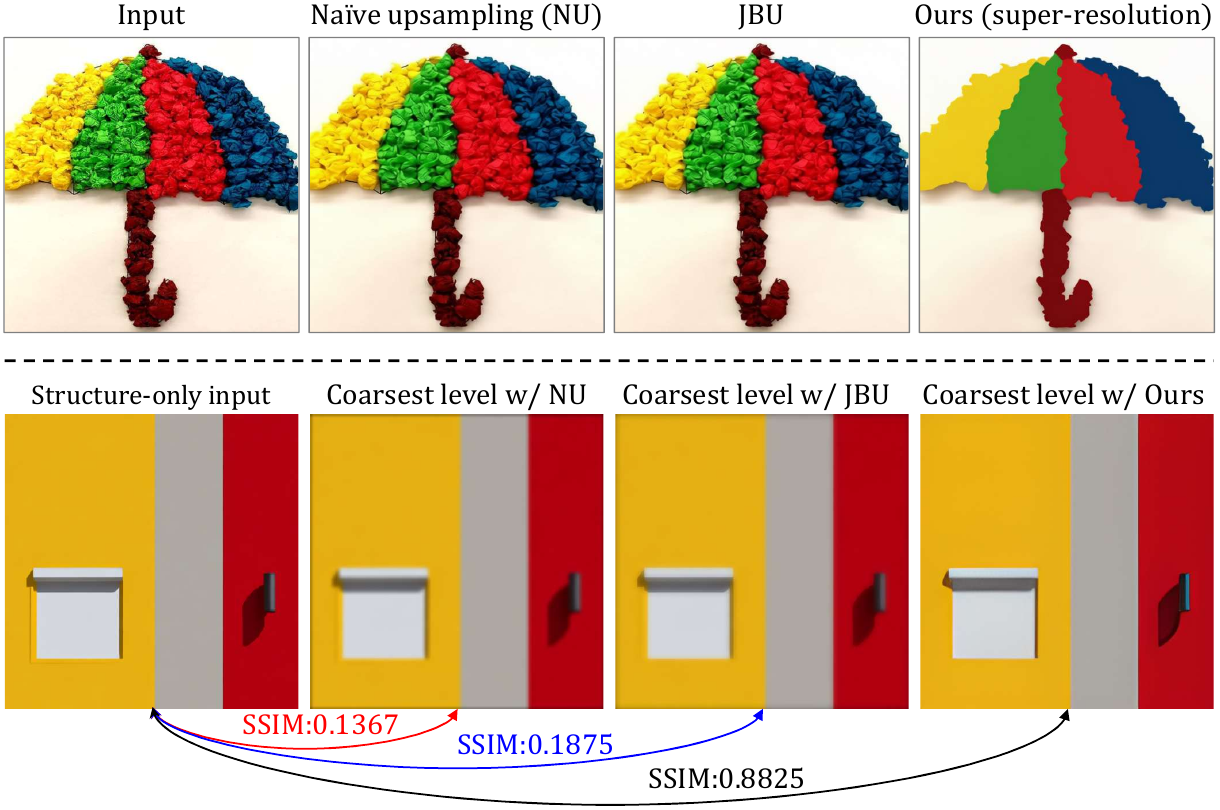}\\
    \vspace{-4mm}
    \caption{Effect of using different image upsampling methods for texture removal reward computation. The top row shows that we can produce significantly better texture filtering result by using our adopted super-resolution model, while the use of naive upsampling and joint bilateral upsampling (JBU) \citep{kopf2007joint} is ineffective for texture removal. The reason behind is that, as shown in the bottom row, even upsampling the coarsest Gaussian pyramid level of a structure-only input with NU and JBU will also produce structure blurry images with low SSIM similarities (0.1367 and 0.1875) to the input, making the texture removal reward an invalid indicator. }
    \label{fig:different_upsample}
    \vspace{-2mm}
\end{figure}

\qing{To narrow down the exploration space of reinforcement fine-tuning, we first fine-tune the Qwen-Image-Edit model on a small paired set with only 10 input images, where the texture filtering results are produced by the method of \citep{zhang2023pyramid} though we find that any source of texture filtered images by existing methods would work with our method (see Figure~\ref{fig:ablation_dataset_methods} for demonstration). While it intuitively seems unlikely to work, we show in Figure~\ref{fig:ablation_size_dataset} that a number of 10 paired images is sufficient to provide a feasible initialization for subsequent reinforcement fine-tuning, and largely increasing the number of paired images does not lead to clear performance improvement because reinforcement fine-tuning is our key to obtain strong texture filtering ability (see Figure~\ref{fig:two-stage-ablation}). Specifically, we perform supervised fine-tuning with the flow matching objective~\citep{liu2022flow,lipman2022flow}:
\begin{equation} 
    \begin{aligned}
    x_t &= (1 - t)x_0 + t x_1, t \in [0, 1], \\
    \mathcal{L}_{FM}(\theta) &= \mathbb{E}_{t, x_0 \sim X_0, x_1 \sim X_1} \Big[ \big\| v - v_\theta(x_t, t, c) \big\|_2^2 \Big],
    \end{aligned}
\end{equation}
where $x_0 \sim X_0$ denotes the target texture filtering image, and $x_1 \sim X_1$ refers to the Gaussian noise. $c$ represents the input image and the related text prompt. Note, we use ``remove texture but preserve structure, keep color and structure faithful to the original image'' as the default prompt throughout the paper. $v=x_1-x_0$ is the target velocity field, and $v_\theta$ indicates the trainable model. As shown in Figure~\ref{fig:two-stage-ablation}, supervised fine-tuning is necessary and beneficial to structure preservation because it helps reduce the possibility to sample texture filtering results with poor structures during reinforcement fine-tuning.
}

\subsection{Reinforcement Fine-tuning}
\qing{After performing supervised fine-tuning as a warm-up, we then perform large-scale reinforcement fine-tuning to gain the ability to generate high-quality texture filtering results. To do so, we synthesize a large-scale unlabeled dataset for model training, and develop a reward function to guide reinforcement learning. }

\vspace{0.5em}
\qing{\noindent \textbf{Training dataset synthesis.} To construct a large-scale unlabeled dataset covering a wide range of textures, instead of relying on laborious manual collection, we employ a text-to-image diffusion model \citep{team2025zimage} to create a dataset containing 10,000 images of $512 \times 512$ resolution in a generative manner, by feeding the model with texts including keywords such as ``texture'', ```mosaic'', ``cross-stitch'', ``brick'', ``lego'', ``jigsaw'', ``pattern'', and ``pebble''. Figure~\ref{fig:train_dataset} shows some example images in the dataset.}

\vspace{0.5em}
\qing{\noindent \textbf{Reward function.} To guide reinforcement fine-tuning, we develop a reward function to quantify the quality of texture filtering output from the perspective of texture removal and structure preservation. As illustrated in Figure~\ref{fig:observation}, given a texture filtering result, higher similarity between the result itself and its coarsest Gaussian pyramid level means better texture removal performance, while higher similarity between the coarsest pyramid levels of the result and the original image indicates better structure preservation performance. Based on this finding, we define the texture removal reward as:
\begin{equation}
\mathcal{R}_{texture} = \text{SSIM}\Big(I_{res}, \; \phi_{sr}\big(G_{res}^{N} \big) \Big),
\label{eq:reward_texture}
\end{equation}
where $I_{res}$ denotes a texture filtering result. $G_{res}^{N}$ is the coarsest Gaussian pyramid level of $I_{res}$ obtained by repeatedly applying Gaussian filtering and then downsampling (reducing size by half) over $I_{res}$, where $N$ denotes the number of downsampling operations. $\phi_{\mathrm{sr}}$ denotes a frozen super-resolution model~\cite{wang2021real} used to upsample $G_{res}^{N}$ to the resolution of $I_{res}$. $\mathrm{SSIM}(\cdot)$ means the SSIM metric \citep{wang2004image}, and we set $N=4$ in our experiments. Figure~\ref{fig:different_upsample} demonstrates the superiority of our employed super-resolution model over other upsampling alternatives. Note, though the super-resolution model works well in structure-preserving upsampling, naively applying it for upsampling the coarsest Gaussian pyramid level of the original image struggles to generate satisfactory texture filtering result (see Figure~\ref{fig:ablation_sr}). The reason is that, as analyzed in \citep{zhang2023pyramid}, the coarsest level of the original image may contain a small amount of texture residuals and may also fail to record small-scale structures.}

Besides, we define the following structure preservation reward:
\begin{equation} 
\mathcal{R}_{structure}
= 1- \left\| G_{src}^{N} - G_{res}^{N} \right\|_1,
\label{eq:structure_reward}
\end{equation}
where $G_{src}^N$ is the coarsest Gaussian pyramid level of the input image $I_{src}$. Note, here we do not similarly adopt the SSIM metric. \zrj{The reason is that SSIM is relatively less discriminative to measure the structure similarity of two very low-resolution images, and thus may generate result with incomplete structures (see Figure~\ref{fig:structure_preservation}).} 

Akin to previous methods~\cite{xu2012structure-rtv,zhang2025stf}, we also define an image fidelity reward to ensure that color and structure in the result $I_{res}$ are consistent to the input image $I_{src}$:
\begin{equation}
\mathcal{R}_{fidelity} = 1- \left\| I_{src} - I_{res} \right\|_2,
\label{eq:reward_fidelity}
\end{equation}
With all the rewards, the final reward function is defined as:
\begin{equation}
\mathcal{R}_{total} = \lambda_1{\mathcal{R}_{texture}} + \lambda_2{\mathcal{R}_{structure}} + \lambda_3{\mathcal{R}_{fidelity}},
\label{eq:reward_final}
\end{equation}
where $\lambda_1$, $\lambda_2$, and $\lambda_3$ are empirically set as $0.2$, $0.6$, and $0.2$. 

\vspace{0.5em}
\qing{\noindent \textbf{Policy optimization.} Considering that the reward computation requires time-consuming iterative diffusion steps to sample texture filtering results from the generative model, in order to trade off the training efficiency and the sampling quality, we follow \cite{zheng2025diffusionnft} to perform policy optimization directly on the forward diffusion process using the flow matching objective. This optimization strategy enables training with arbitrary black-box solvers, allowing us to leverage high-order solvers such as DPM-Solver~\cite{lu2022dpm} for fast sampling of high-quality results. Given an input image and a prompt, to optimize the policy, we first use DPM-Solver to sample multiple filtering results from the generative model. We then compute their rewards using Eq.~\eqref{eq:reward_final} and encourage the model to favor high-reward policies while discouraging low-reward ones. Accordingly, the policy optimization loss is formulated as follows:
\begin{equation}
\mathcal{L}(\theta) = \mathbb{E}_{c, \pi^{\text{old}}(x_0) | c, t} \left[ r \left\| v_\theta^+(x_t, c, t) - v \right\|_2^2 + (1 - r) \left\| v_\theta^-(x_t, c, t) - v \right\|_2^2 \right],
\end{equation}
where $v$ is the target velocity field and $r$ is the normalized reward score. $\pi^{\text{old}}(x_0)$ denotes filtering result $x_0$ sampled from the distribution of the old policy $v^{\text{old}}$ (i.e., original generative model). The implicit positive and negative policies $v_\theta^+$ and $v_\theta^-$ are combinations of the old policy $v^{\text{old}}$ and the training policy $v_\theta$, weighted by a hyperparameter $\beta$:
\begin{equation}
\begin{aligned}
v_\theta^+(x_t, c, t) = (1 - \beta) v^{\text{old}}(x_t, c, t) + \beta v_\theta(x_t, c, t), \\
v_\theta^-(x_t, c, t) = (1 + \beta) v^{\text{old}}(x_t, c, t) - \beta v_\theta(x_t, c, t).
\end{aligned}
\end{equation}
\zrj{The normalized reward $r \in [0, 1]$ is obtained by:
\begin{equation}
r= \frac{1}{2} + \frac{1}{2} \text{clip} \left[ \frac{\mathcal{R}_{total}(x_0, c) - \mathbb{E}_{\pi^{\mathrm{old}}(\cdot | c)} \mathcal{R}_{total}(x_0, c)}{Z_c}, -1, 1 \right],
\end{equation}
where $Z_c$ is the standard deviation of the rewards, and clip[$\cdot$, -1, 1] is a truncation function that limits the value to the range of [-1, 1].}}

\subsection{Implementation Details}
\qing{\noindent \textbf{Supervised fine-tuning.} We apply LoRA~\cite{hu2022lora} to fine-tune the Qwen-Image-Edit model at a resolution of $512 \times 512$, using the AdamW optimizer with a learning rate of 3e-4 and a weight decay of 1e-2. \zrj{The LoRA layers are applied to the key, query, and value projection layers, as well as the output projection layers within each attention block. The rank of each LoRA layer is set as 32, and its weights are initialized from a Gaussian distribution. We use the default setting of the original super-resolution model~\cite{wang2021real} in our experiments.} We finetune the model for 200 steps (taking about 4 hours) on four 24GB 4090 GPUs.

\begin{figure}[!t]
    \centering
    \begin{subfigure}[c]{0.32\linewidth}
		\centering
		\includegraphics[width=\textwidth]{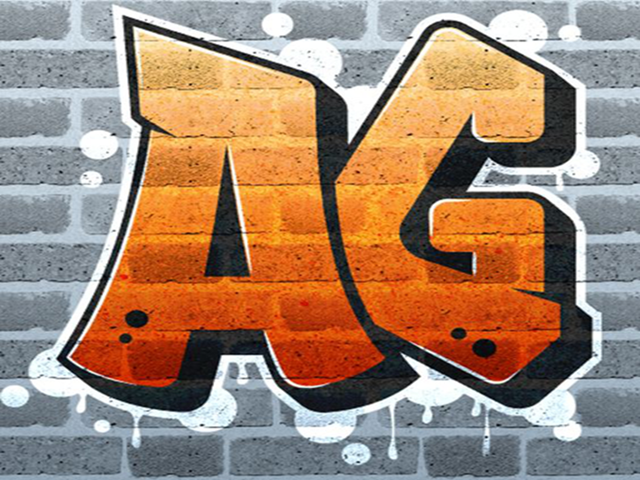} \\
        \vspace{-2mm}
		\caption*{Input}
    \end{subfigure}
    \begin{subfigure}[c]{0.32\linewidth}
		\centering
		\includegraphics[width=\textwidth]{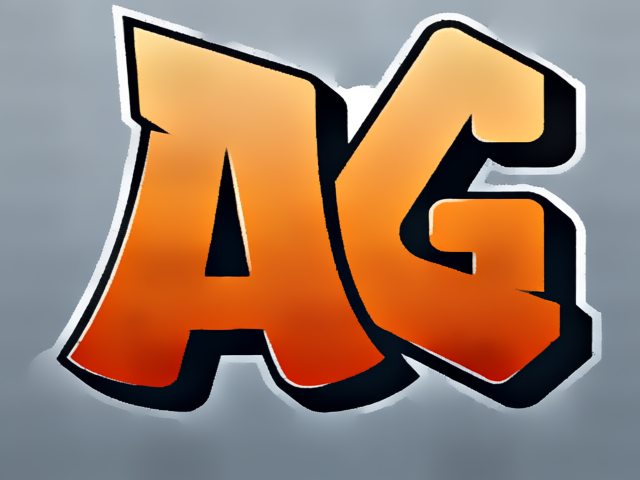} \\
        \vspace{-2mm}
		\caption*{Result with SSIM metric}
    \end{subfigure}
    \begin{subfigure}[c]{0.32\linewidth}
		\centering
		\includegraphics[width=\textwidth]{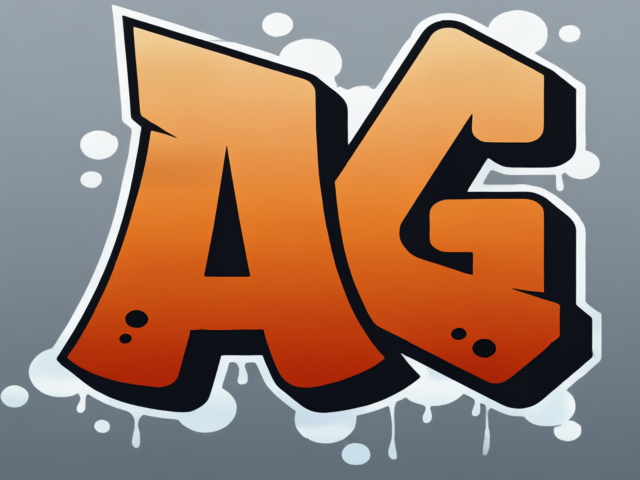} \\ 
        \vspace{-2mm}
		\caption*{Ours with Eq.~\eqref{eq:structure_reward}}
    \end{subfigure}
    \vspace{-3mm}
    \caption{\zrj{Effect of using different metrics for structure preservation reward computation. As shown, our metric helps obtain result with better structures.}}
    \label{fig:structure_preservation}
    \vspace{-2mm}
\end{figure}

\vspace{0.5em}
\noindent \textbf{Reinforcement fine-tuning.} We use DPM-Solver with 6 denoising steps for sampling. The LoRA configuration is the same as supervised fine-tuning. For each training image, we sample 12 texture filtering results using the prompt ``remove texture but preserve structure, keep color and structure faithful to the original image''. We then perform policy optimization for 15 epochs using the AdamW optimizer with a learning rate of 3e-4 and a weight decay of 1e-2. The whole training takes about 10 hours on eight 80GB A800 GPUs.}

\begin{figure*}[!t]
	\centering
	\begin{subfigure}[c]{0.161\textwidth}
		\centering
        \includegraphics[width=\linewidth]{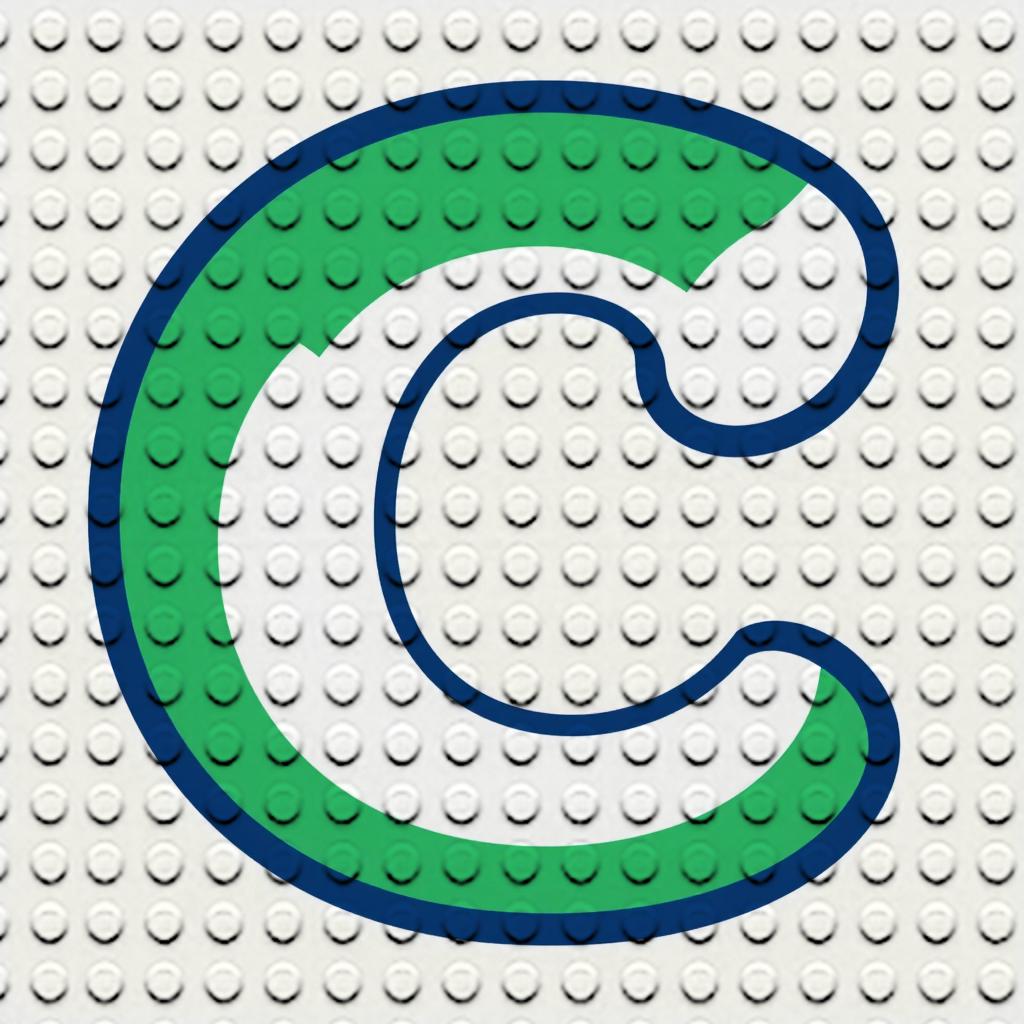} \\ \vspace{0.5mm}
        \includegraphics[width=\linewidth]{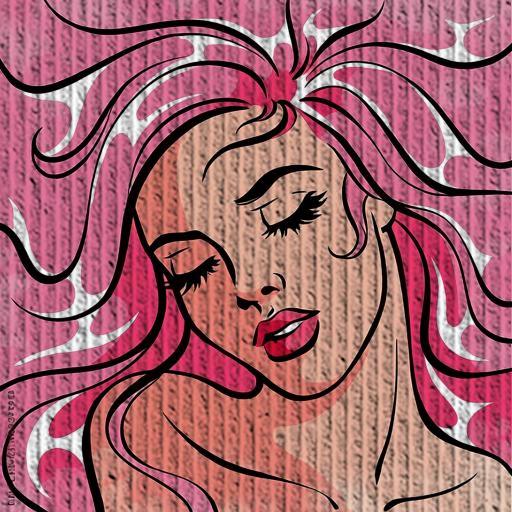} \\ 
        \vspace{-2mm}
		\caption*{Input}
	\end{subfigure}
    \begin{subfigure}[c]{0.161\textwidth}
		\centering
        \includegraphics[width=\linewidth]{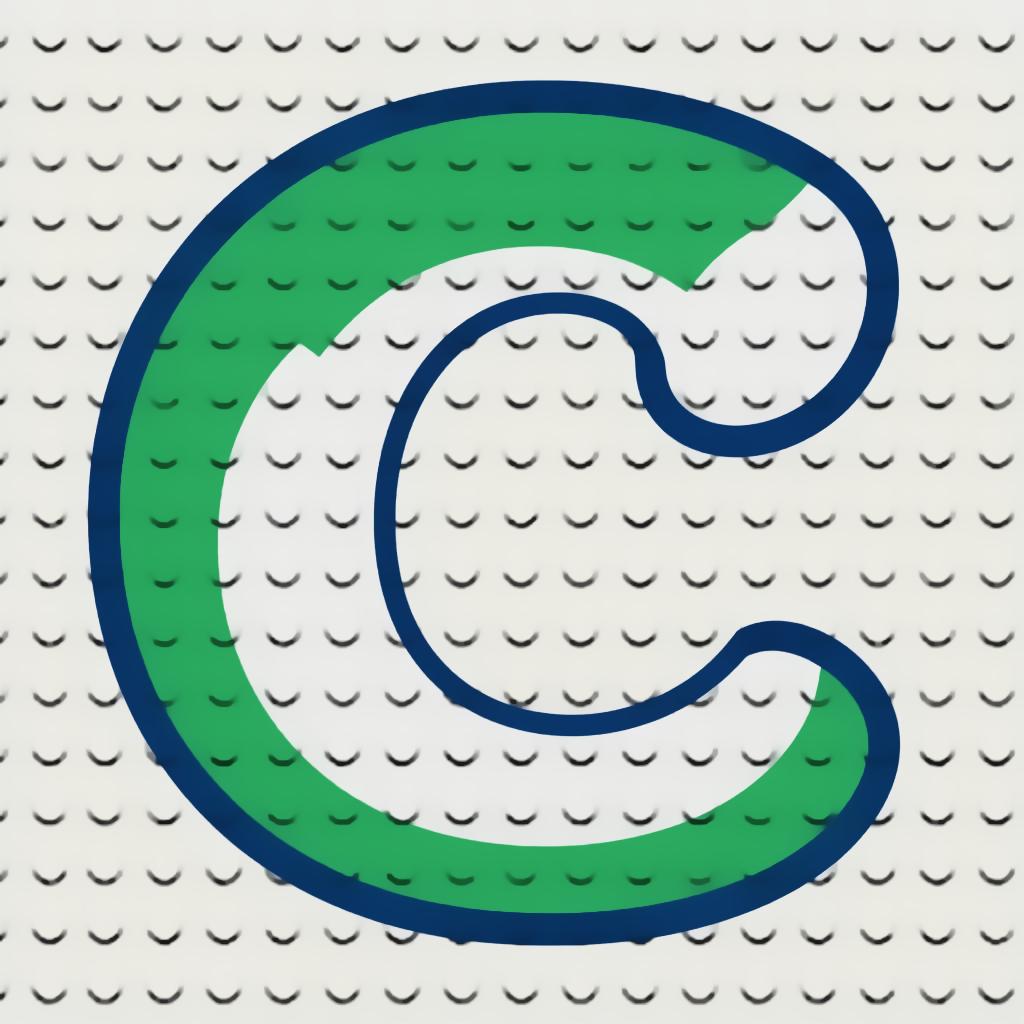} \\ \vspace{0.5mm}
        \includegraphics[width=\linewidth]{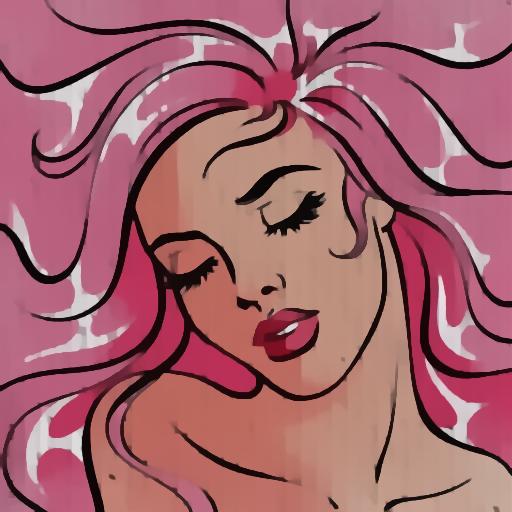} \\ 
        \vspace{-2mm}
		\caption*{\cite{xu2012structure-rtv}}
	\end{subfigure}
    \begin{subfigure}[c]{0.161\textwidth}
		\centering
        \includegraphics[width=\linewidth]{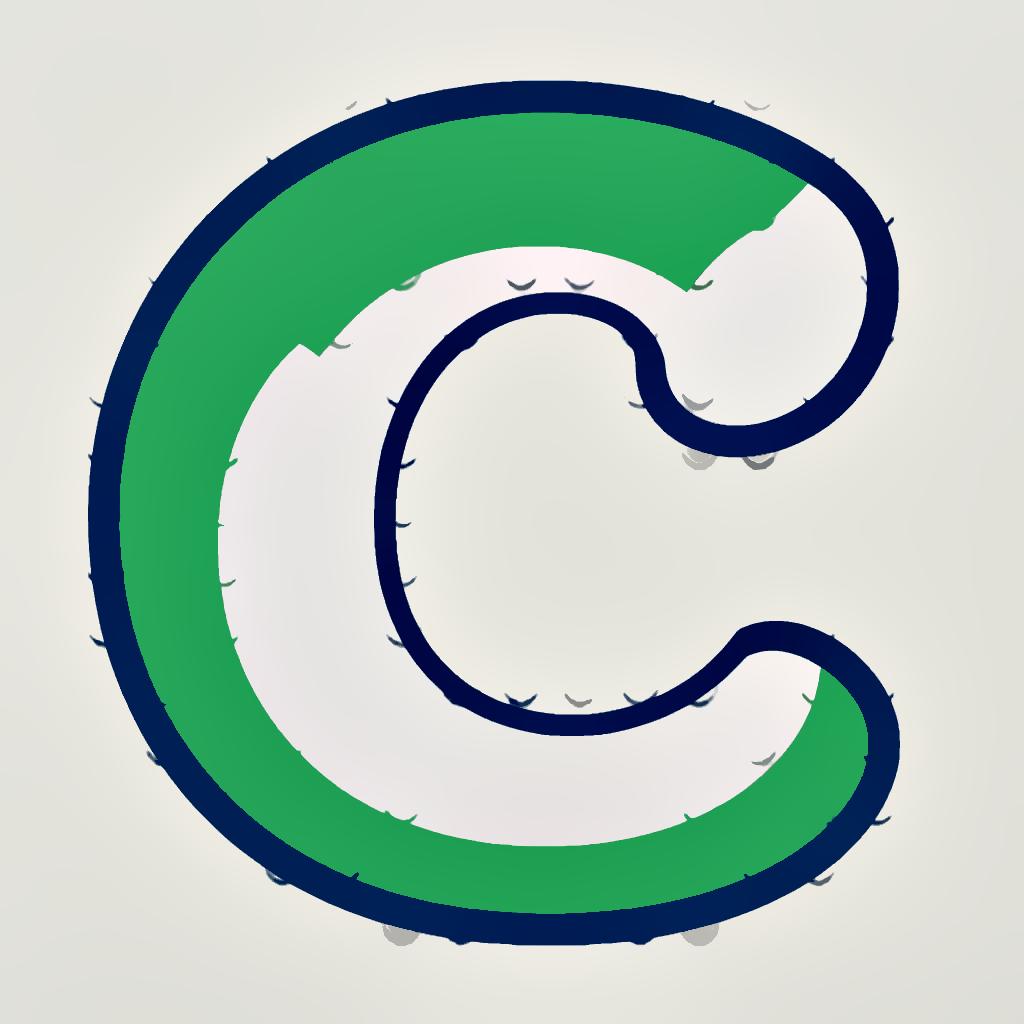} \\ \vspace{0.5mm}
        \includegraphics[width=\linewidth]{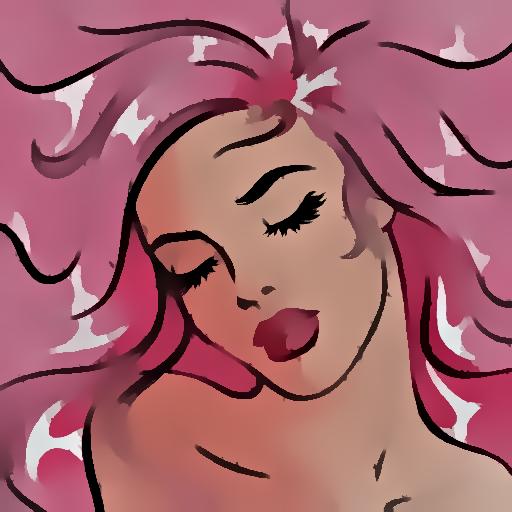} \\ 
        \vspace{-2mm}
		\caption*{\cite{zhang2023pyramid}}
	\end{subfigure}
    \begin{subfigure}[c]{0.161\textwidth}
		\centering
        \includegraphics[width=\linewidth]{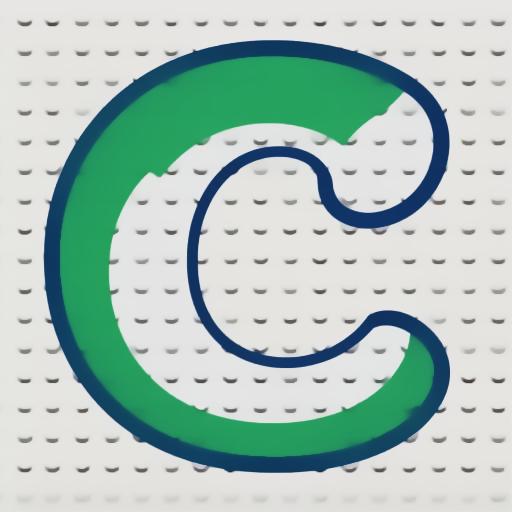} \\ \vspace{0.5mm}
        \includegraphics[width=\linewidth]{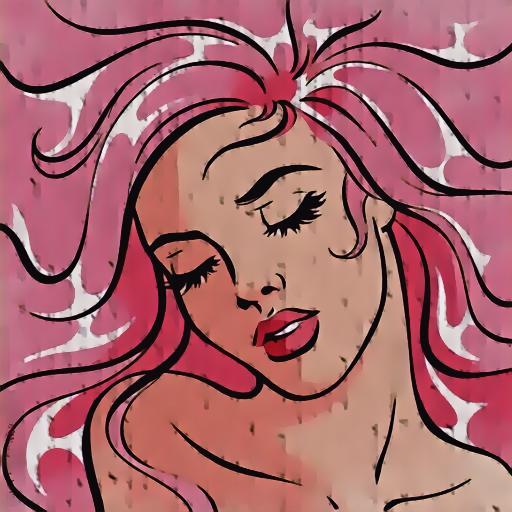} \\ 
        \vspace{-2mm}
		\caption*{\cite{zhang2025stf}}
	\end{subfigure}
    \begin{subfigure}[c]{0.161\textwidth}
		\centering
        \includegraphics[width=\linewidth]{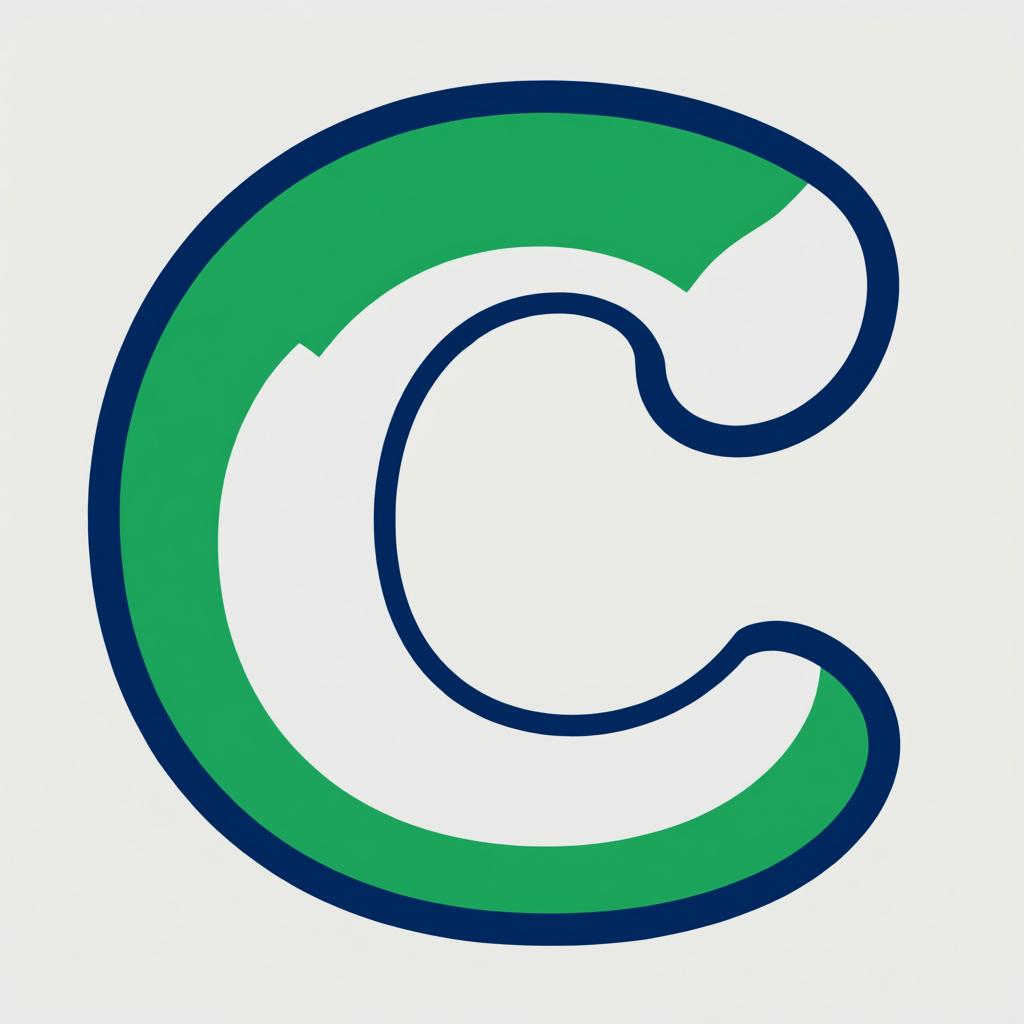} \\ \vspace{0.5mm}
        \includegraphics[width=\linewidth]{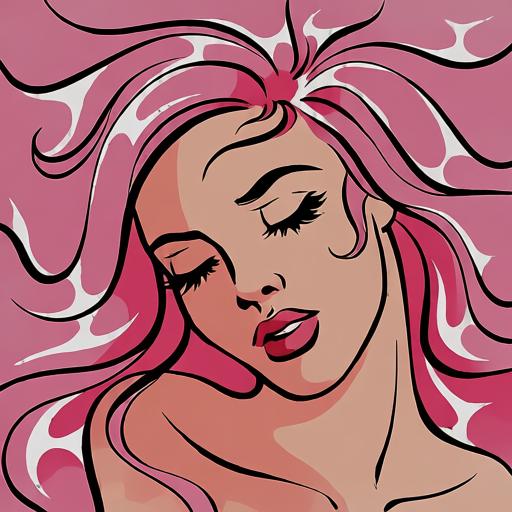} \\ 
        \vspace{-2mm}
		\caption*{Ours}
	\end{subfigure}
    \begin{subfigure}[c]{0.161\textwidth}
		\centering
        \includegraphics[width=\linewidth]{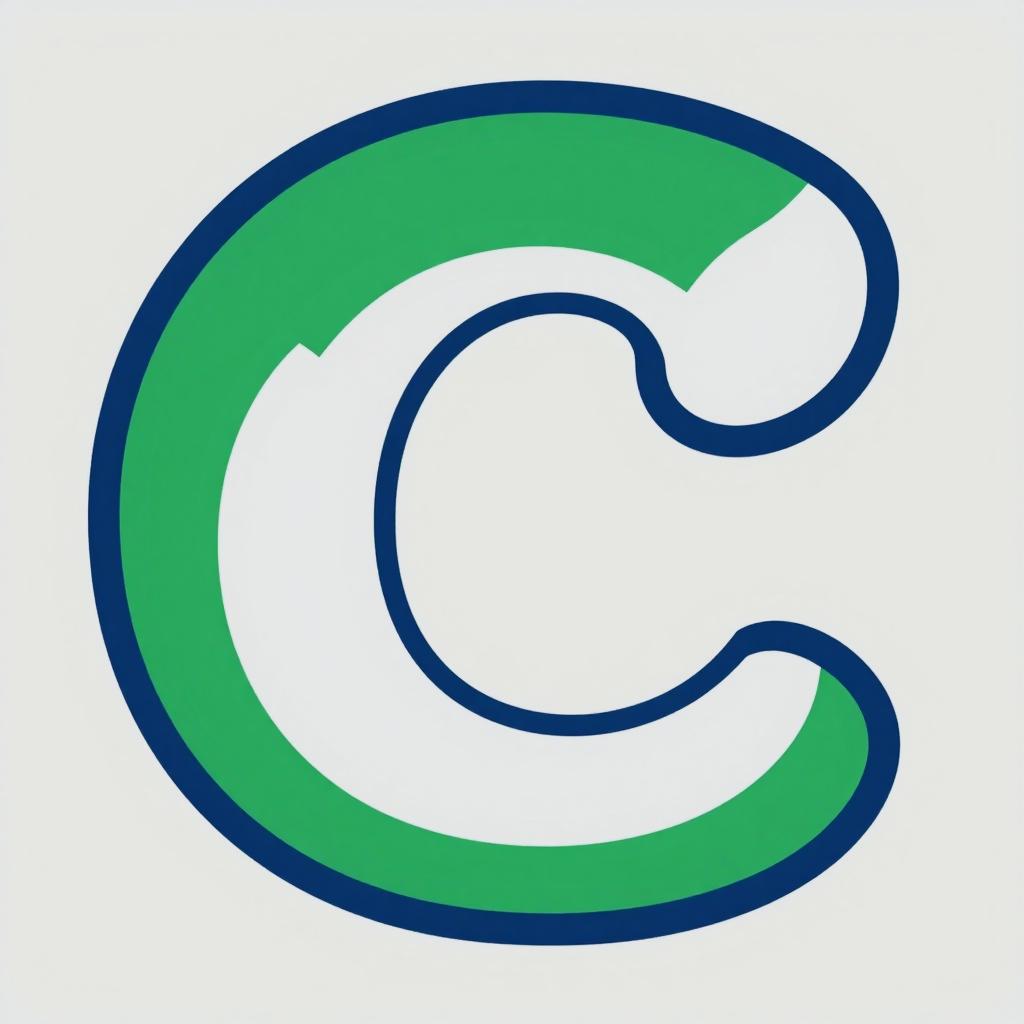} \\ \vspace{0.5mm}
        \includegraphics[width=\linewidth]{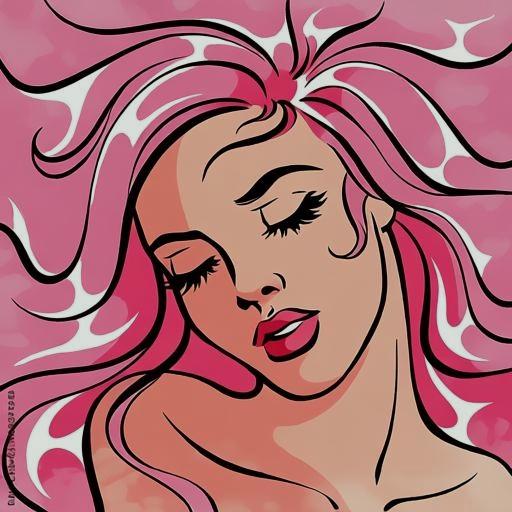} \\ 
        \vspace{-2mm}
		\caption*{GT}
	\end{subfigure}
    \vspace{-3mm}
	\caption{Comparison with previous methods on the synthetic dataset. Please see the supplementary material for more visual comparison results. }
	\label{fig:synthetic_comparison}
    \vspace{-2mm}
\end{figure*}

\section{Experiments}
\qing{\noindent \textbf{Evaluation datasets and metrics.} We evaluate our method on two datasets, including a synthetic paired dataset with ground truth (GT) as reference and a real-world dataset without GT. The synthetic dataset contains 500 image pairs, which is constructed by mixing structure-only images with textures as done in \citep{lu2018deep}, with the mixed image as input and the original structure-only image as GT. For the synthetic dataset, we employ PSNR and SSIM for evaluation. \zrj{To create the real-world dataset, we first crawl 10,000 images from Flickr by searching with the same keywords used in our training data synthesis, and then manually select 2,000 images with diverse textures. Note, due to lack of GT as reference, we provide only qualitative comparison on the real-world dataset.}
}

\subsection{Comparison with Previous Methods}
\qing{\noindent \textbf{Baselines.} We compare with various previous methods, including RTV \citep{xu2012structure-rtv}, BTF \citep{cho2014bilateral-btf}, ULS \citep{fan2018image}, GSF \citep{liu2021generalized}, PTF \citep{zhang2023pyramid}, and SSTF \citep{zhang2025stf}. For fair comparison, we produce their results using publicly available implementations or trained models provided by the authors with the recommended parameter settings.}

\begin{figure}[!t]
    \centering
    \begin{subfigure}[c]{0.24\linewidth}
		\centering
		\includegraphics[width=\textwidth]{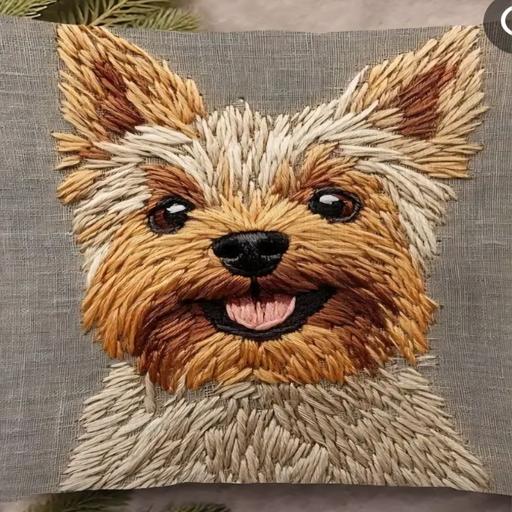} \\
        \vspace{-2mm}
		\caption*{Input}
    \end{subfigure}
    \begin{subfigure}[c]{0.24\linewidth}
		\centering
		\includegraphics[width=\textwidth]{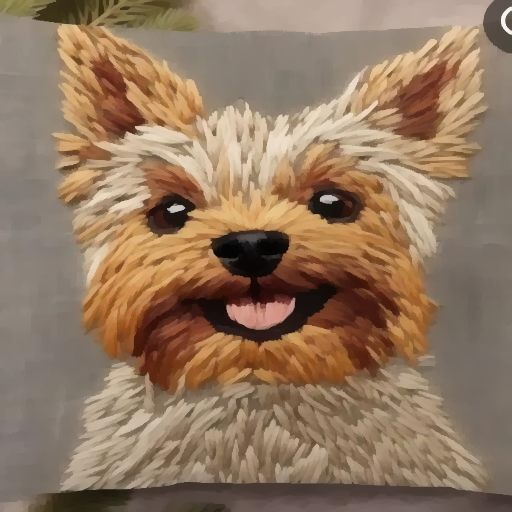} \\ \vspace{1pt}
		\includegraphics[width=\textwidth]{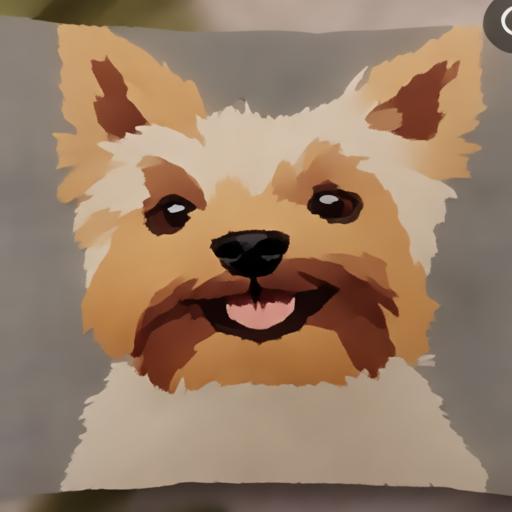} \\
        \vspace{-2mm}
        \caption*{10 image pairs}
    \end{subfigure}
    \begin{subfigure}[c]{0.24\linewidth}
		\centering
		\includegraphics[width=\textwidth]{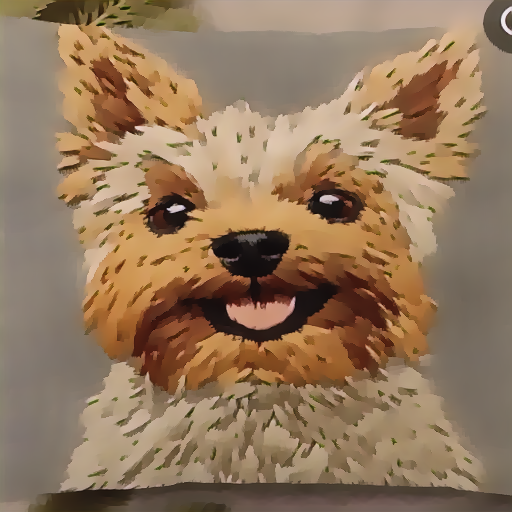} \\ \vspace{1pt}
		\includegraphics[width=\textwidth]{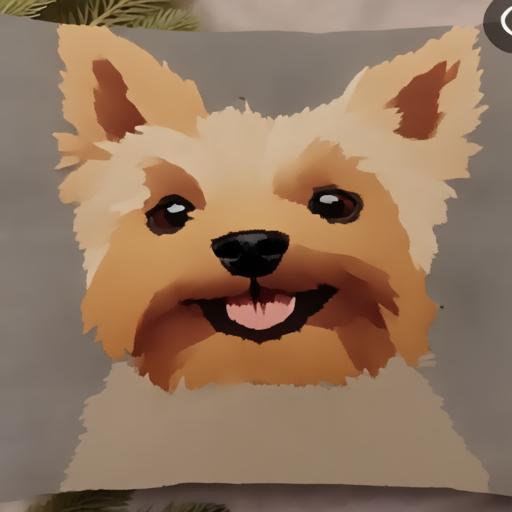} \\
        \vspace{-2mm}
        \caption*{100 image pairs}
    \end{subfigure}
    \begin{subfigure}[c]{0.24\linewidth}
		\centering
		\includegraphics[width=\textwidth]{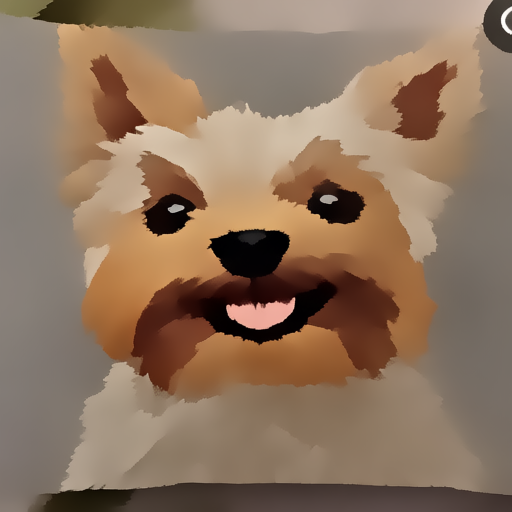} \\ \vspace{1pt}
        \includegraphics[width=\textwidth]{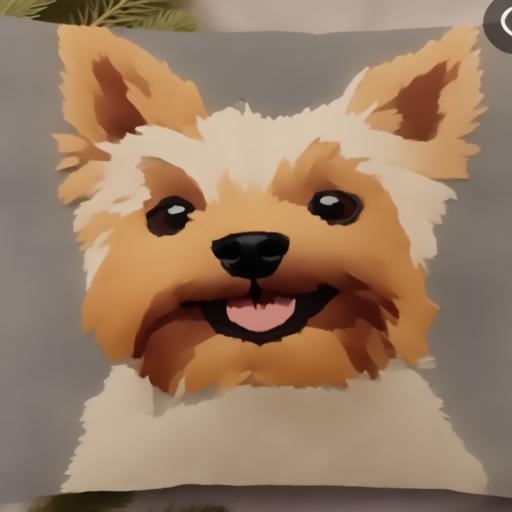} \\ 
        \vspace{-2mm}
        \caption*{1000 image pairs}
    \end{subfigure}
  \vspace{-3mm}
    \caption{Effect of varying numbers of image pairs in supervised fine-tuning. The top row shows the results produced with only supervised fine-tuning, while the bottom gives the final results with reinforcement fine-tuning. }
    \label{fig:ablation_size_dataset}
    \vspace{-2mm}
\end{figure}

\begin{figure*}[!t]
	\centering
	\begin{subfigure}[c]{0.161\textwidth}
		\centering
        \includegraphics[width=\linewidth]{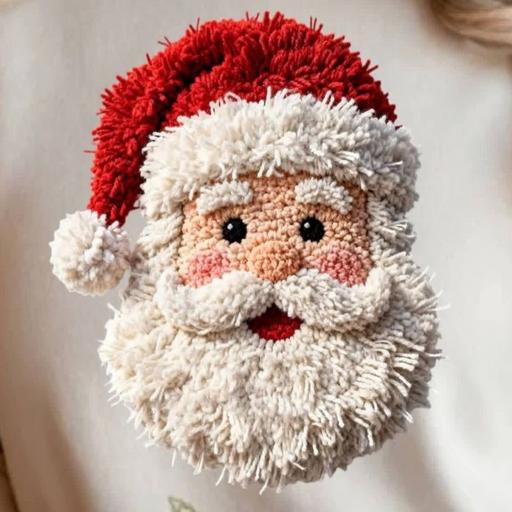} \\ \vspace{0.5mm}
        \includegraphics[width=\linewidth]{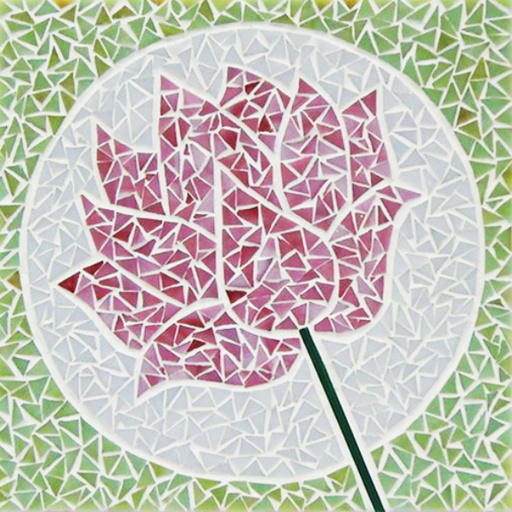} \\ 
        \vspace{-2mm}
		\caption*{Input}
	\end{subfigure}
    \begin{subfigure}[c]{0.161\textwidth}
		\centering
        \includegraphics[width=\linewidth]{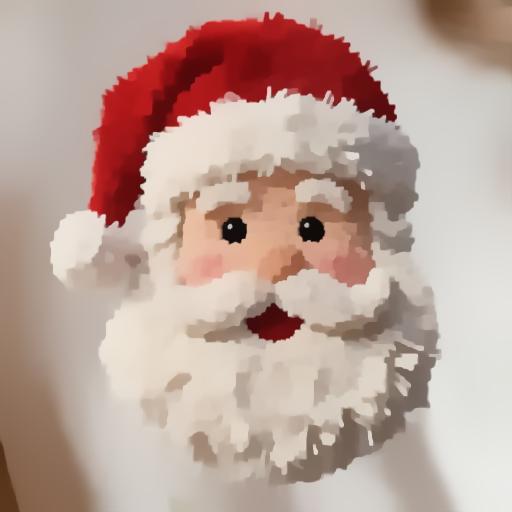} \\ \vspace{0.5mm}
        \includegraphics[width=\linewidth]{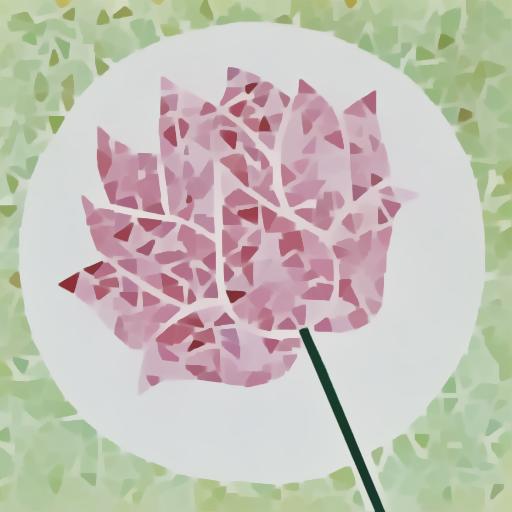} \\ 
        \vspace{-2mm}
		\caption*{\cite{xu2012structure-rtv}}
	\end{subfigure}
    \begin{subfigure}[c]{0.161\textwidth}
		\centering
        \includegraphics[width=\linewidth]{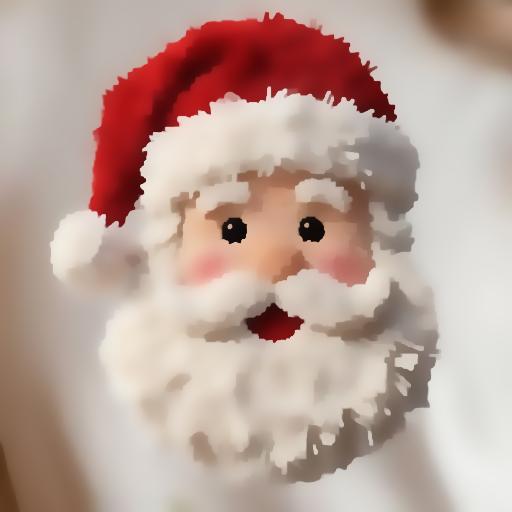} \\ \vspace{0.5mm}
        \includegraphics[width=\linewidth]{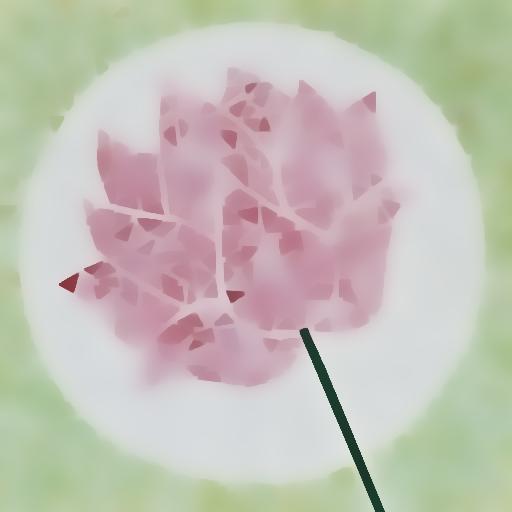} \\ 
        \vspace{-2mm}
		\caption*{\cite{cho2014bilateral-btf}}
	\end{subfigure}
    \begin{subfigure}[c]{0.161\textwidth}
		\centering
        \includegraphics[width=\linewidth]{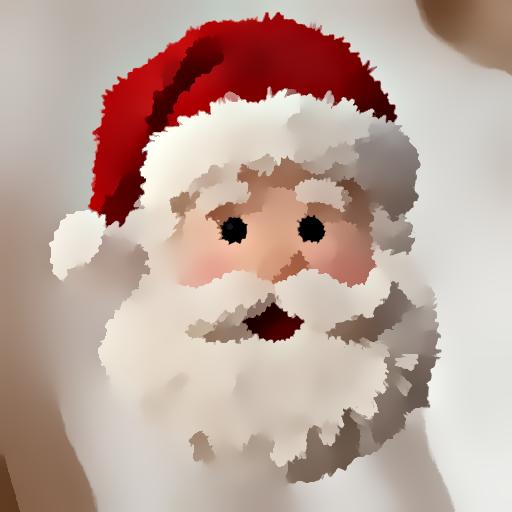} \\ \vspace{0.5mm}
        \includegraphics[width=\linewidth]{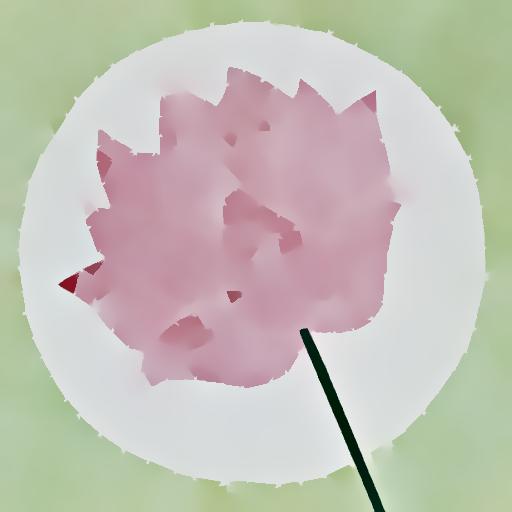} \\ 
        \vspace{-2mm}
		\caption*{\cite{zhang2023pyramid}}
	\end{subfigure}
    \begin{subfigure}[c]{0.161\textwidth}
		\centering
        \includegraphics[width=\linewidth]{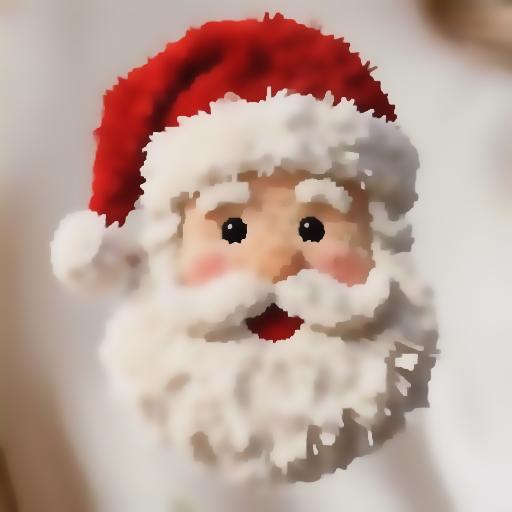} \\ \vspace{0.5mm}
         \includegraphics[width=\linewidth]{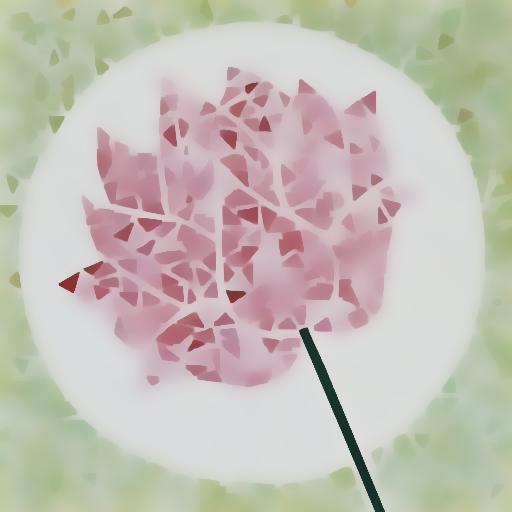} \\ 
        \vspace{-2mm}
		\caption*{\cite{zhang2025stf}}
	\end{subfigure}
    \begin{subfigure}[c]{0.161\textwidth}
		\centering
        \includegraphics[width=\linewidth]{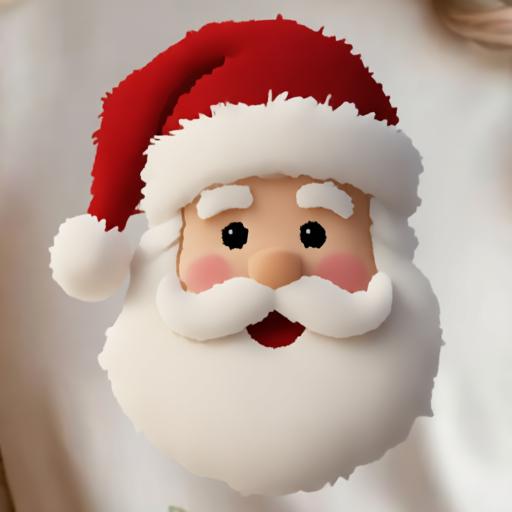} \\ \vspace{0.5mm}
        \includegraphics[width=\linewidth]{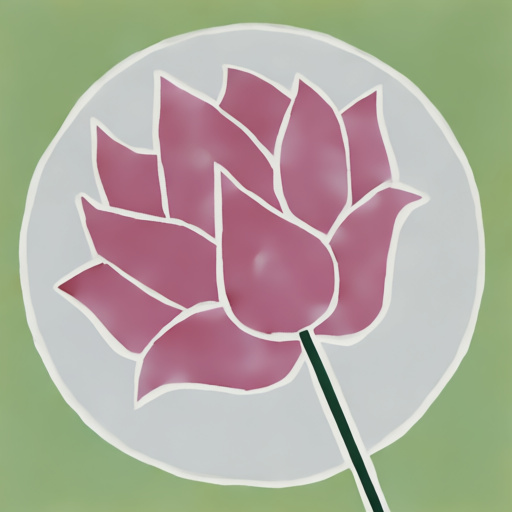} \\ 
        \vspace{-2mm}
		\caption*{Ours}
	\end{subfigure}
    \vspace{-3mm}
	\caption{Comparison with previous methods on the real-world dataset. Please see the supplementary material for more visual comparison results.}
	\label{fig:real_comparison}
\end{figure*}

\begin{figure*}[!t]
	\centering
	\begin{subfigure}[c]{0.161\textwidth}
		\centering
        \includegraphics[width=\linewidth]{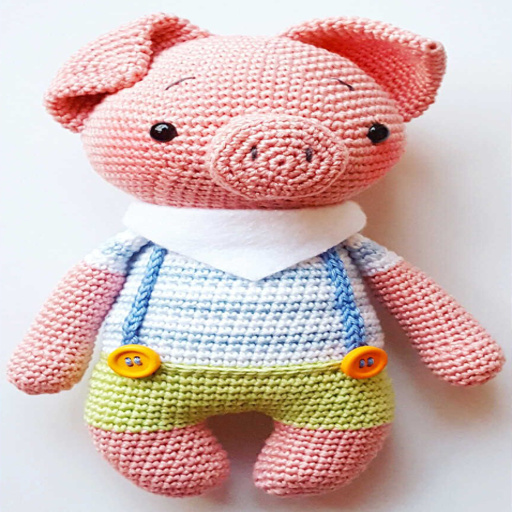} \\ \vspace{0.5mm}
        \includegraphics[width=\linewidth]{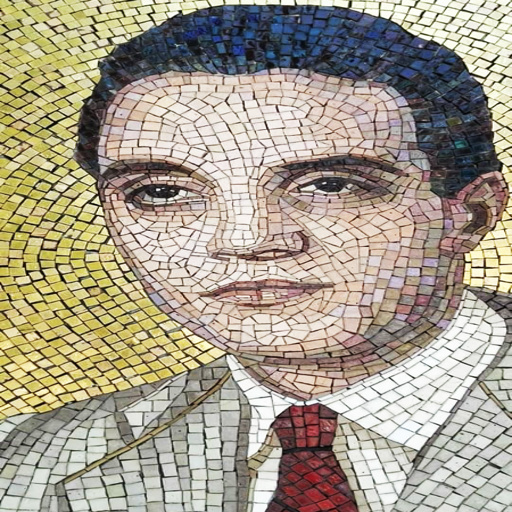} \\ \vspace{0.5mm}
        \includegraphics[width=\linewidth]{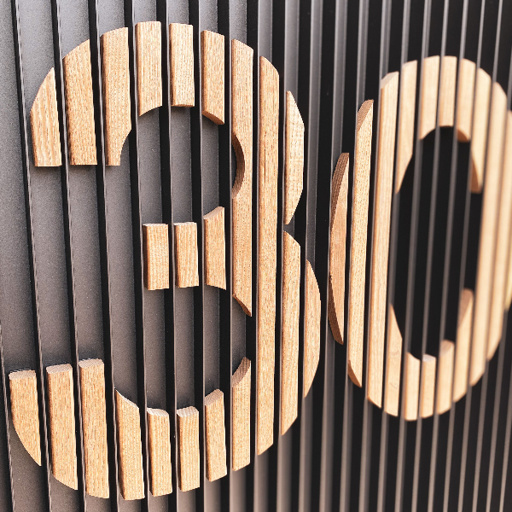} \\ \vspace{0.5mm}
        \includegraphics[width=\linewidth]{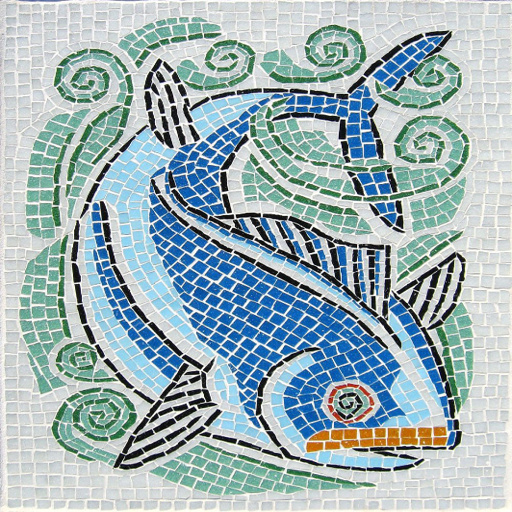} \\ \vspace{0.5mm}
        \includegraphics[width=\linewidth]{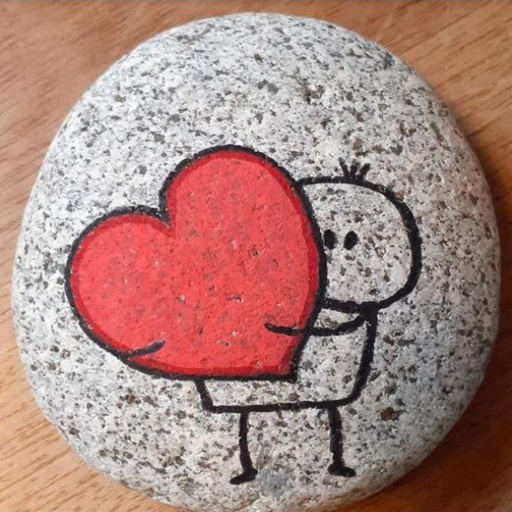} \\ \vspace{0.5mm}
        \includegraphics[width=\linewidth]{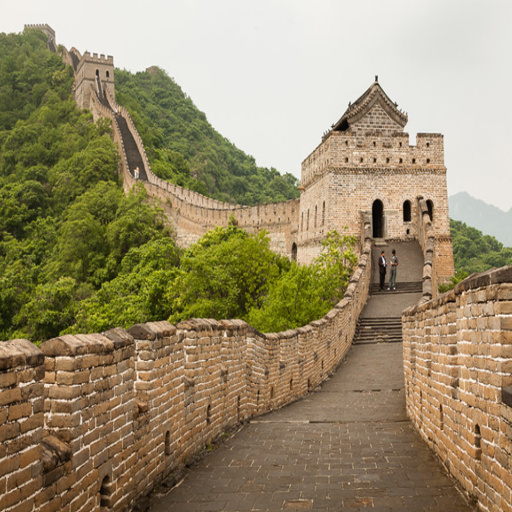} \\ \vspace{0.5mm}
        \includegraphics[width=\linewidth]{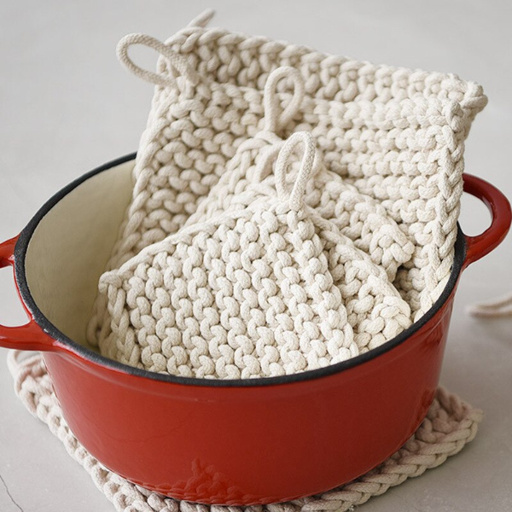}\\  \vspace{-2mm}
		\caption*{Input}
	\end{subfigure}
    \begin{subfigure}[c]{0.161\textwidth}
		\centering
        \includegraphics[width=\linewidth]{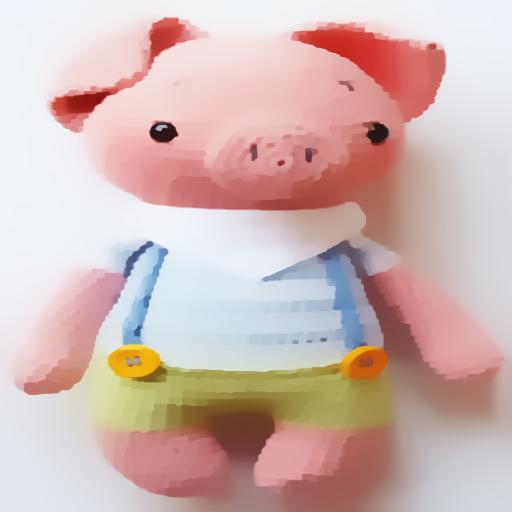} \\ \vspace{0.5mm}
         \includegraphics[width=\linewidth]{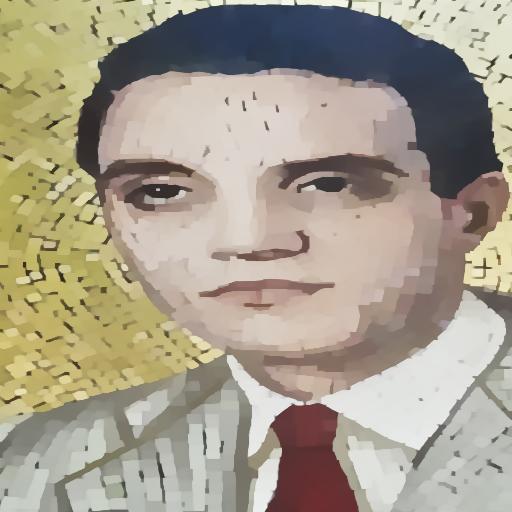}\\ \vspace{0.5mm}
        \includegraphics[width=\linewidth]{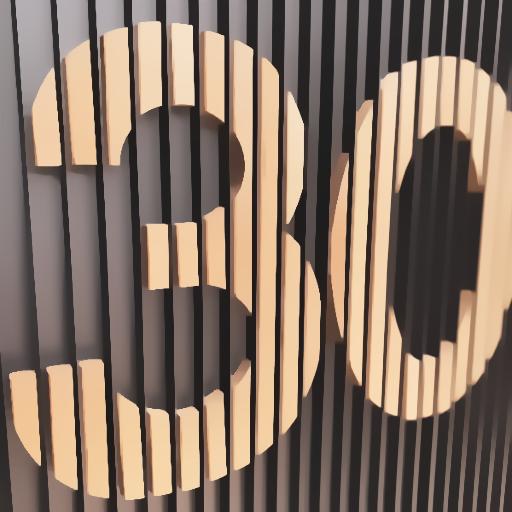} \\ \vspace{0.5mm}
        \includegraphics[width=\linewidth]{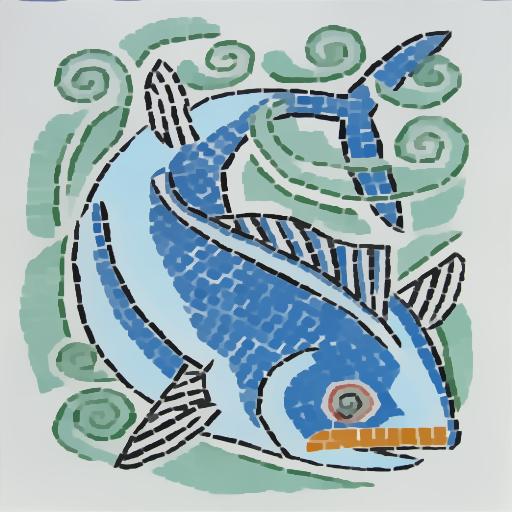} \\ \vspace{0.5mm}
        \includegraphics[width=\linewidth]{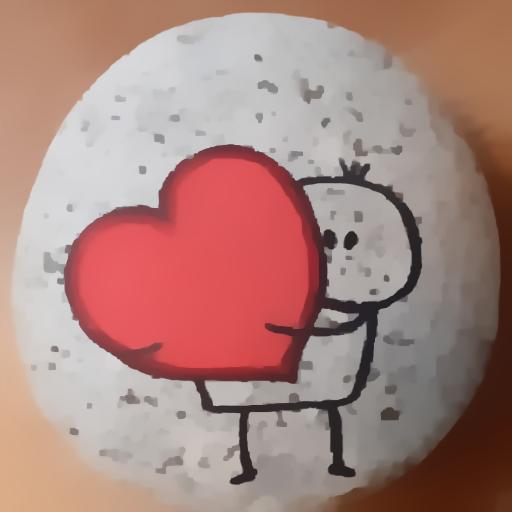} \\ \vspace{0.5mm}
        \includegraphics[width=\linewidth]{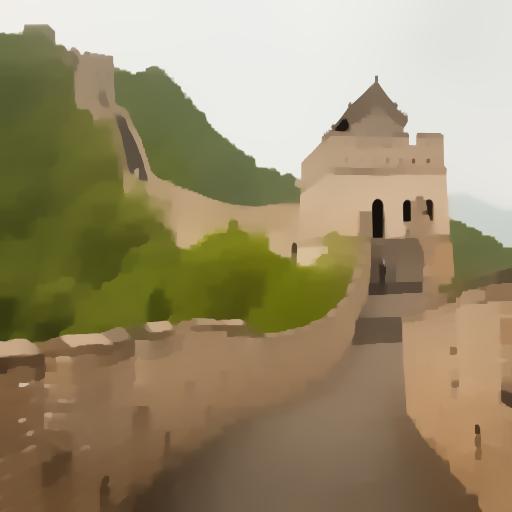} \\ \vspace{0.5mm}
        \includegraphics[width=\linewidth]{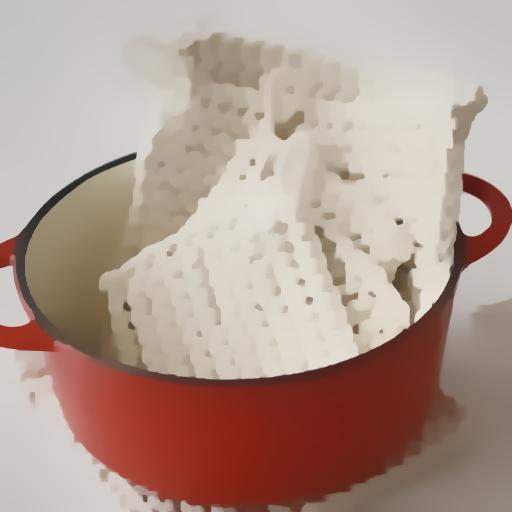} \\  \vspace{-2mm}
		\caption*{\cite{xu2012structure-rtv}}
	\end{subfigure}
    \begin{subfigure}[c]{0.161\textwidth}
		\centering
        \includegraphics[width=\linewidth]{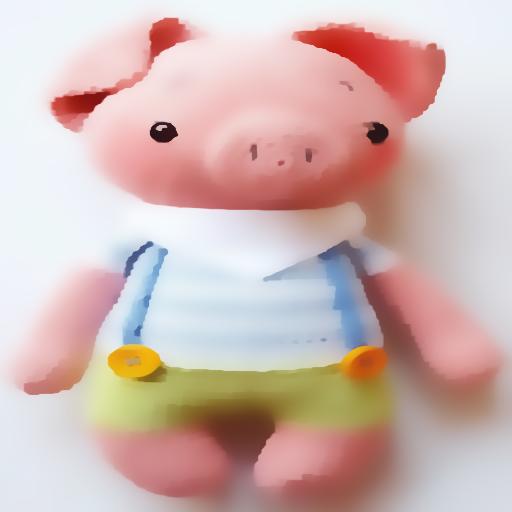} \\ \vspace{0.5mm}
         \includegraphics[width=\linewidth]{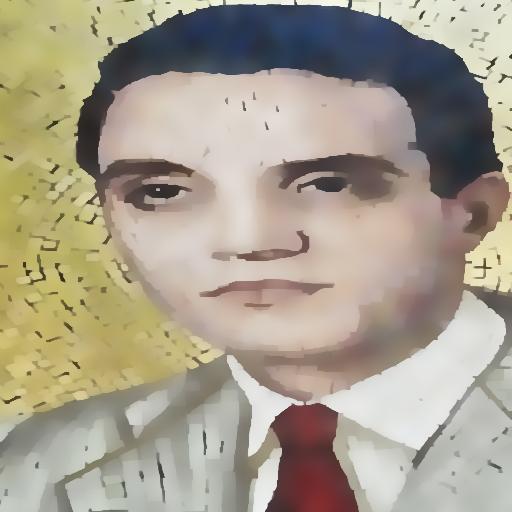} \\ \vspace{0.5mm}
        \includegraphics[width=\linewidth]{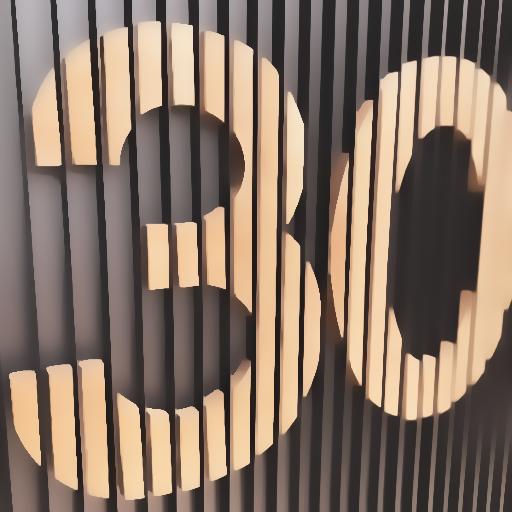} \\ \vspace{0.5mm}
        \includegraphics[width=\linewidth]{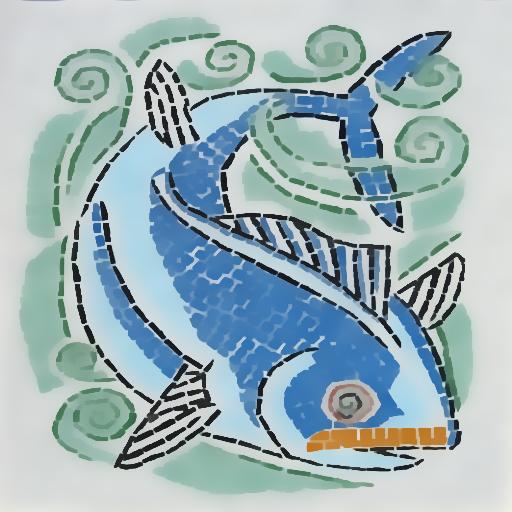} \\ \vspace{0.5mm}
        
        \includegraphics[width=\linewidth]{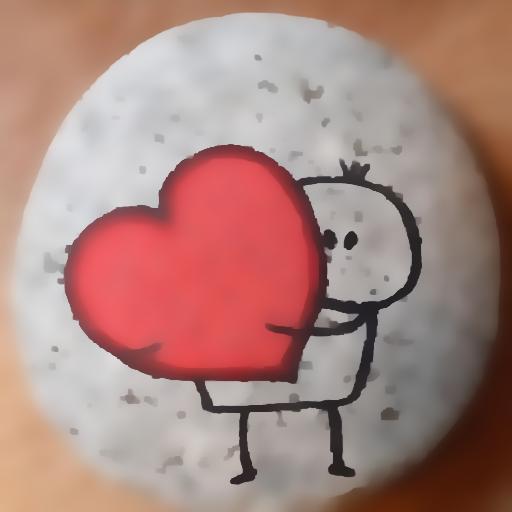} \\ \vspace{0.5mm}
        \includegraphics[width=\linewidth]{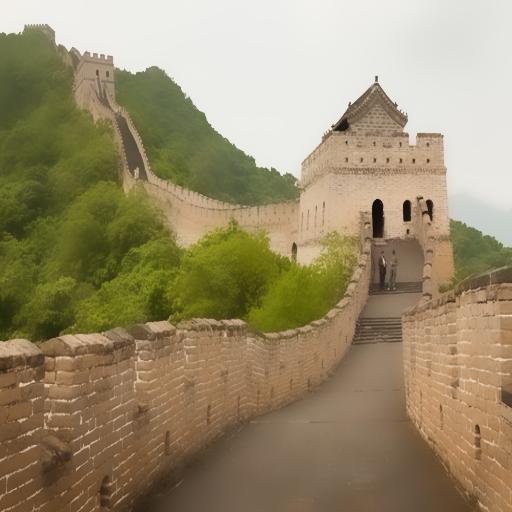} \\ \vspace{0.5mm}
        \includegraphics[width=\linewidth]{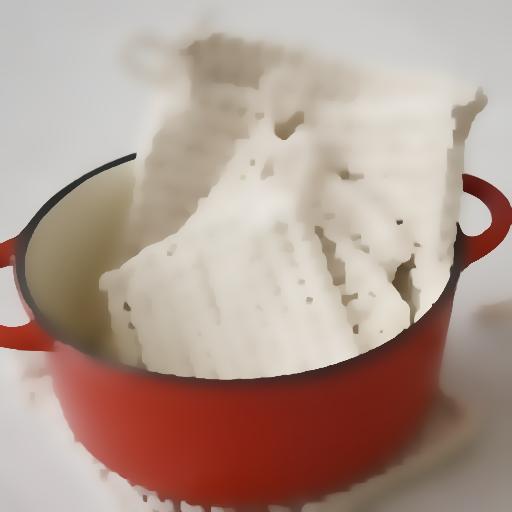} \\  \vspace{-2mm}
		\caption*{\cite{cho2014bilateral-btf}}
	\end{subfigure}
    \begin{subfigure}[c]{0.161\textwidth}
		\centering
        \includegraphics[width=\linewidth]{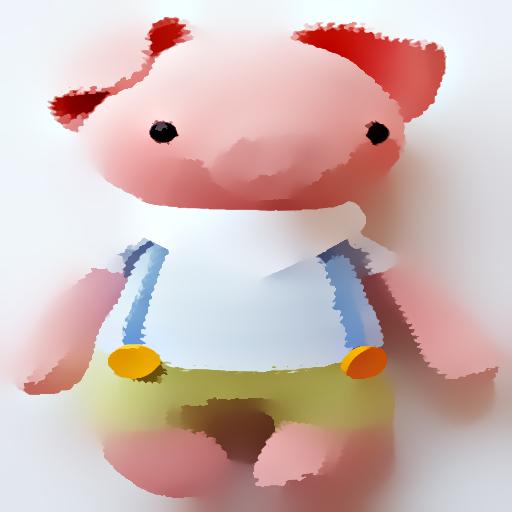} \\ \vspace{0.5mm}
         \includegraphics[width=\linewidth]{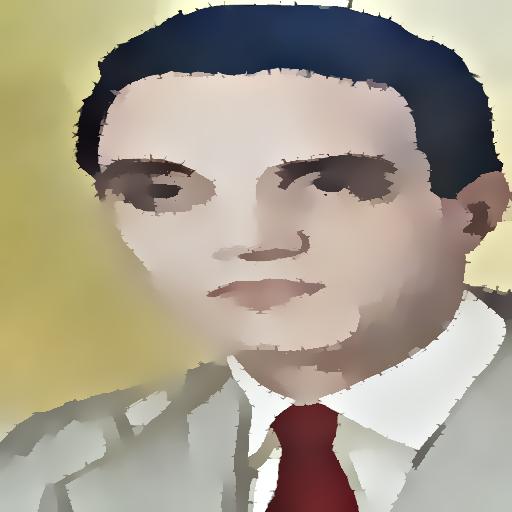} \\ \vspace{0.5mm}
        \includegraphics[width=\linewidth]{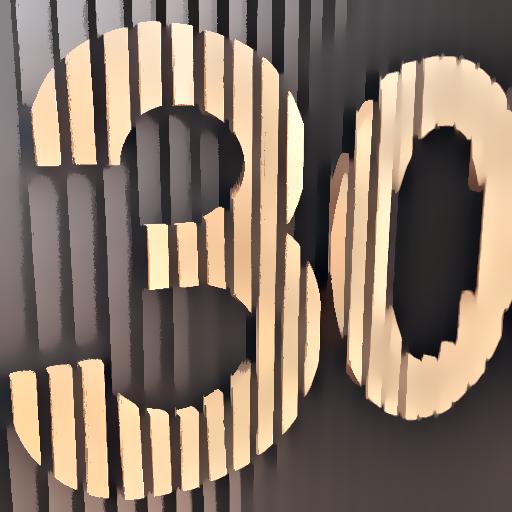} \\ \vspace{0.5mm}
        \includegraphics[width=\linewidth]{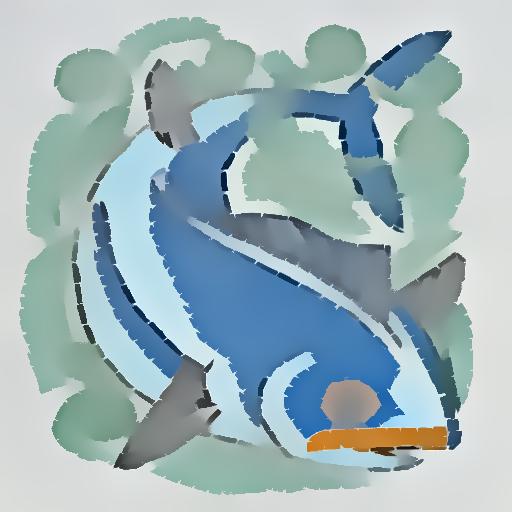} \\ \vspace{0.5mm}
        \includegraphics[width=\linewidth]{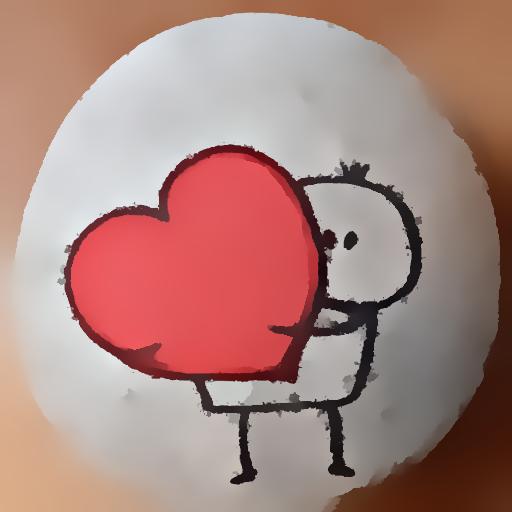} \\ \vspace{0.5mm}
        \includegraphics[width=\linewidth]{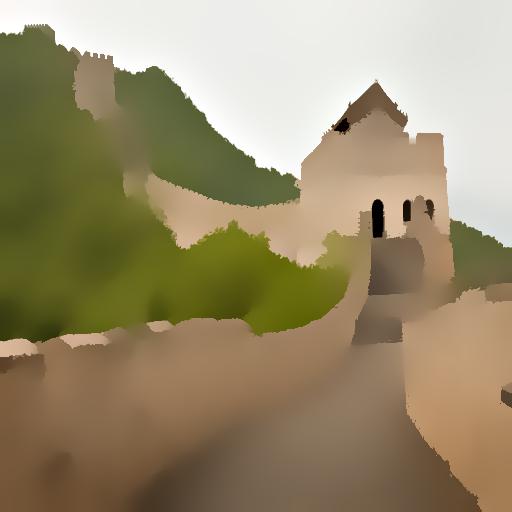} \\ \vspace{0.5mm}
        \includegraphics[width=\linewidth]{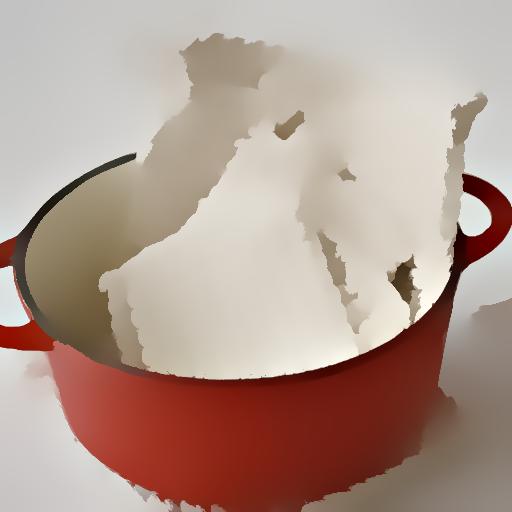} \\  \vspace{-2mm}
		\caption*{\cite{zhang2023pyramid}}
	\end{subfigure}
    \begin{subfigure}[c]{0.161\textwidth}
		\centering
        \includegraphics[width=\linewidth]{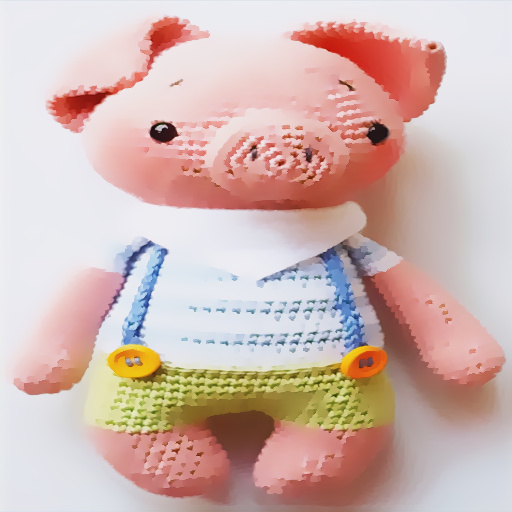} \\ \vspace{0.5mm}
        \includegraphics[width=\linewidth]{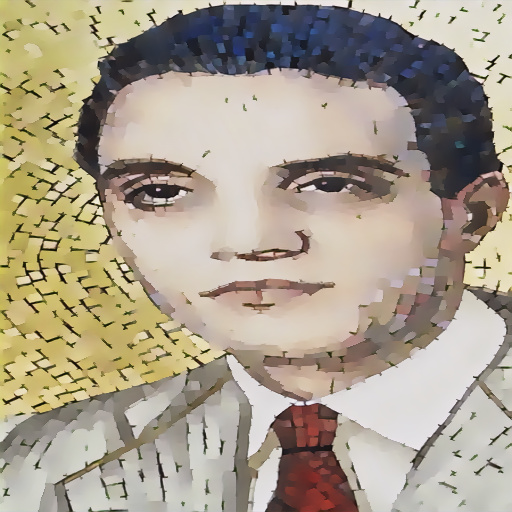} \\ \vspace{0.5mm}
        \includegraphics[width=\linewidth]{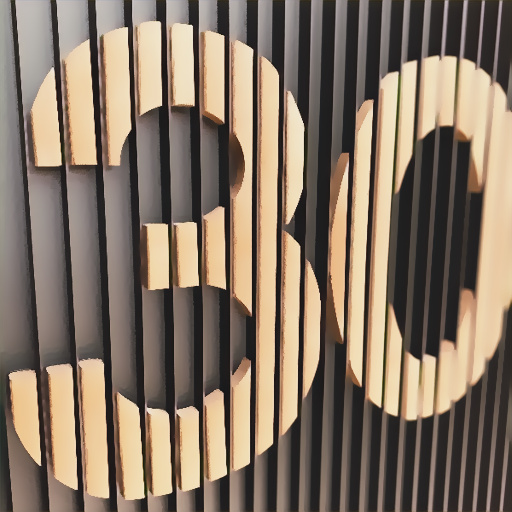} \\ \vspace{0.5mm}
        \includegraphics[width=\linewidth]{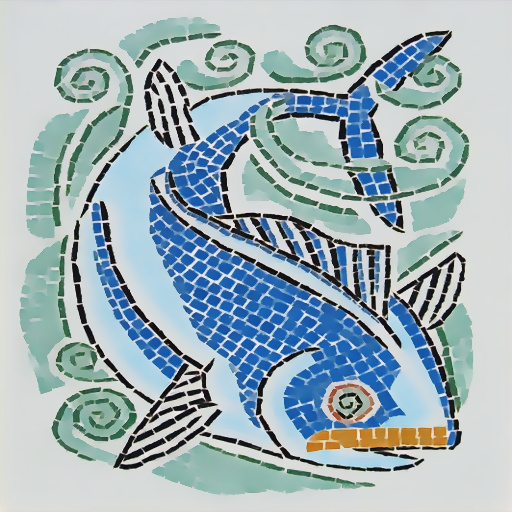} \\ \vspace{0.5mm}
        \includegraphics[width=\linewidth]{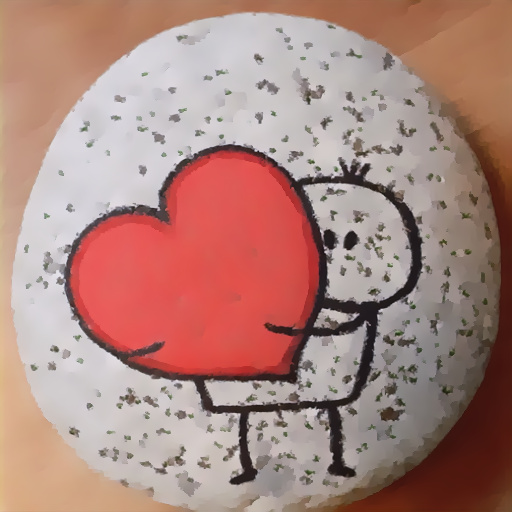} \\ \vspace{0.5mm}
        \includegraphics[width=\linewidth]{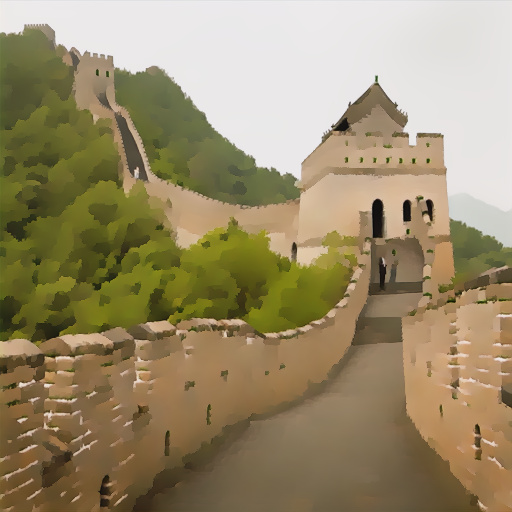} \\ \vspace{0.5mm}
        \includegraphics[width=\linewidth]{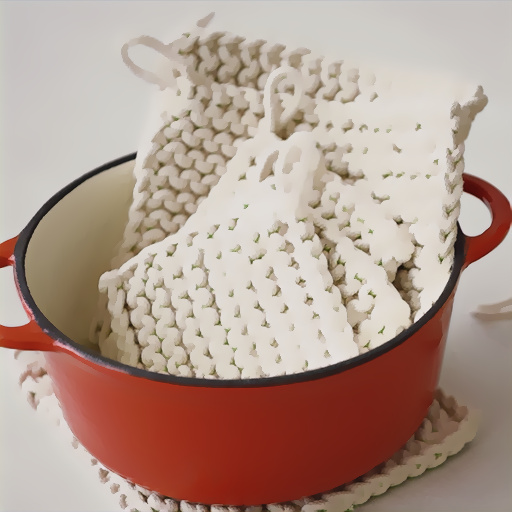} \\  \vspace{-2mm}
		\caption*{\cite{zhang2025stf}}
	\end{subfigure}
    \begin{subfigure}[c]{0.161\textwidth}
		\centering
        \includegraphics[width=\linewidth]{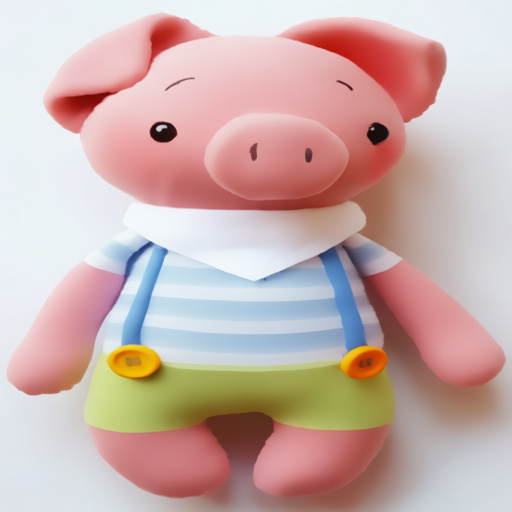} \\ \vspace{0.5mm}
        \includegraphics[width=\linewidth]{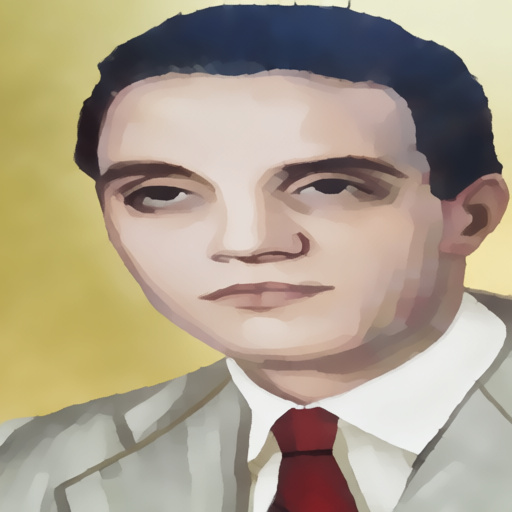} \\ \vspace{0.5mm}
        \includegraphics[width=\linewidth]{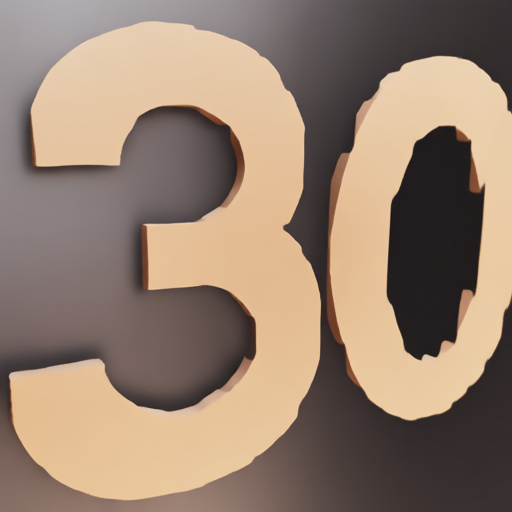} \\ \vspace{0.5mm}
        \includegraphics[width=\linewidth]{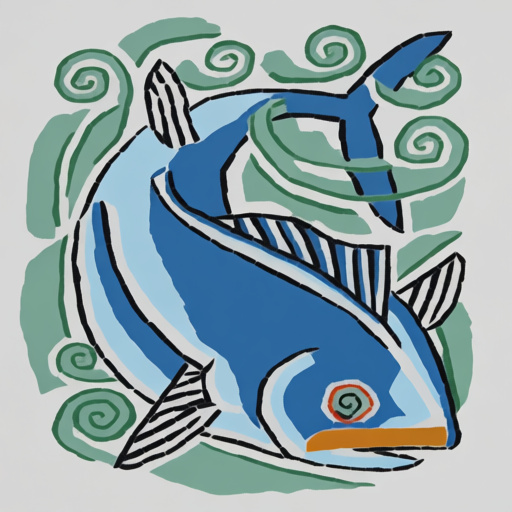} \\ \vspace{0.5mm}
        \includegraphics[width=\linewidth]{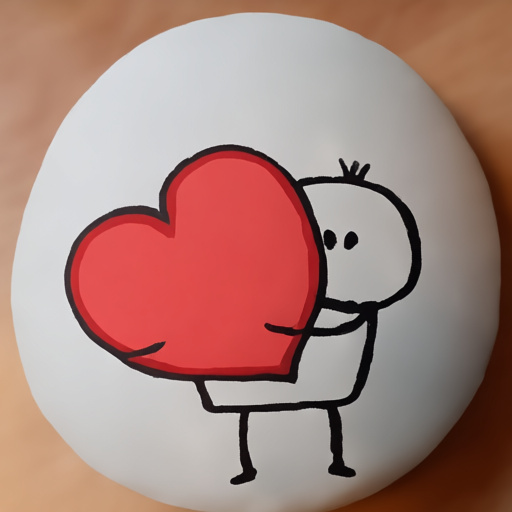} \\ \vspace{0.5mm}
        \includegraphics[width=\linewidth]{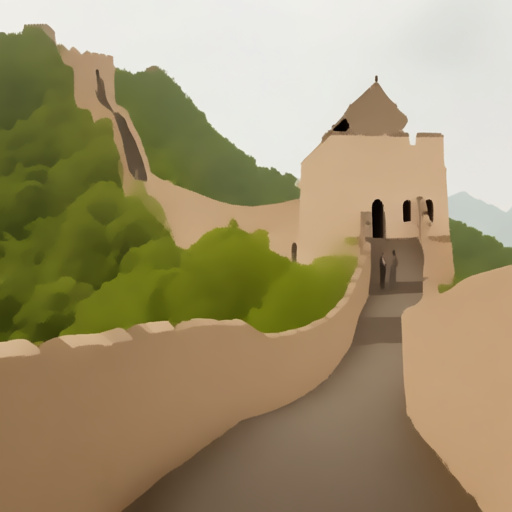} \\ \vspace{0.5mm}
        \includegraphics[width=\linewidth]{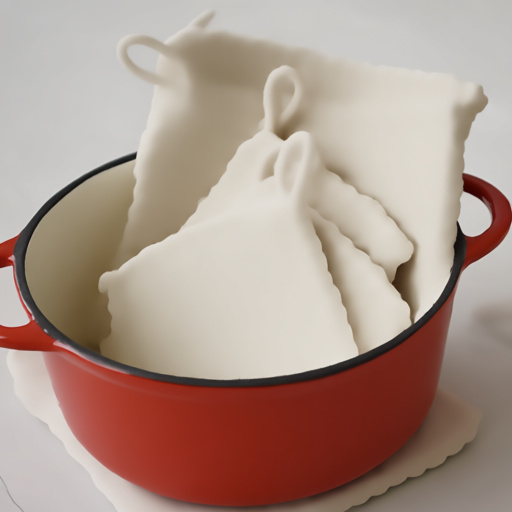} \\  \vspace{-2mm}
		\caption{Ours}
	\end{subfigure}
    \vspace{-2mm}
	\caption{More visual comparison with previous methods on the real-world dataset.}
	\label{fig:more_visual_comparison}
\end{figure*}

\vspace{0.5em}
\zrj{\noindent \textbf{Evaluation on synthetic and real-world datasets.} Table~\ref{table:quantitative_comparison} presents the quantitative comparison on the synthetic dataset with paired images. As shown, our method outperforms all the compared methods on both the PSNR and SSIM metrics. Figure~\ref{fig:synthetic_comparison} further presents visual comparison results, where we can see that our method produces better results visually closer to GT while the compared methods either fail to remove the texture or faithfully preserve the structure. Besides, we also evaluate our method on unlabeled real-world dataset. As shown in Figures~\ref{fig:real_comparison} and \ref{fig:more_visual_comparison}, our method clearly outperforms others in terms of texture removal and structure preservation. 
}

\begin{table}[t]
\centering
\caption{Quantitative comparison on the synthetic dataset.}
\vspace{-3mm}
\label{table:quantitative_comparison}
\resizebox{0.8\columnwidth}{!}{
\begin{tabular}{lcc}
\toprule 
\multicolumn{1}{l}{Method} & \multicolumn{1}{c}{PSNR$\uparrow$} & \multicolumn{1}{c}{SSIM$\uparrow$} \\ \midrule
RTV \cite{xu2012structure-rtv} & 25.57 & 0.840 \\
BTF \cite{cho2014bilateral-btf}  & 25.83 & 0.842 \\
ULS \cite{fan2018image} & 25.44 & 0.847 \\
GSF \cite{liu2021generalized} & 24.98 & 0.835 \\
PTF \cite{zhang2023pyramid} & 26.23  & 0.846 \\
SSTF \cite{zhang2025stf} & 25.71 & 0.838 \\ \midrule
Ours (full method) & \textbf{29.06} & \textbf{0.895} \\
Ours w/ only SFT& 24.75 & 0.830 \\
Ours w/ only RFT& 27.89 & 0.862 \\
Ours w/o $\mathcal{R}_{texture}$ & 24.37 & 0.848 \\
Ours w/o $\mathcal{R}_{structure}$ & 27.52 & 0.865 \\
Ours w/o $\mathcal{R}_{fidelity}$ & 28.48 & 0.883 \\
Ours w/ naive upsampling & 24.68 & 0.815 \\
Ours w/ JBU \cite{kopf2007joint} & 24.92 & 0.826 \\
Ours using SSIM in Eq.~\eqref{eq:structure_reward} & 28.13 & 0.874 \\
Ours w/ 100 pairs in SFT & 28.95 & 0.891 \\
Ours w/ 1000 pairs in SFT & 29.02 & 0.893 \\
\zrj{Only SFT w/ 1000 pairs (without RFT)} & 26.05 & 0.843 \\
\bottomrule
\end{tabular}}
\vspace{-2mm}
\end{table}

\subsection{More Analysis}
\begin{figure*}[!t]
	\centering
	\begin{subfigure}[c]{0.161\textwidth}
		\centering
        \includegraphics[width=\linewidth]{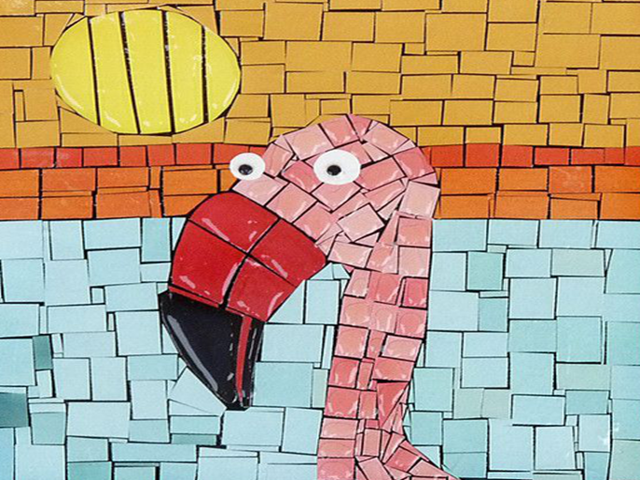}\\
        \vspace{-2mm}
		\caption*{Input}
	\end{subfigure}
    \begin{subfigure}[c]{0.161\textwidth}
		\centering
        \includegraphics[width=\linewidth]{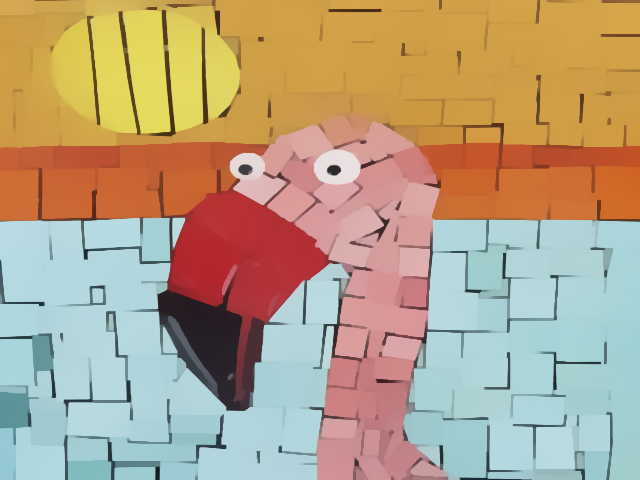}\\
        \vspace{-2mm}
		\caption*{\cite{xu2012structure-rtv}}
	\end{subfigure}
    \begin{subfigure}[c]{0.161\textwidth}
		\centering
        \includegraphics[width=\linewidth]{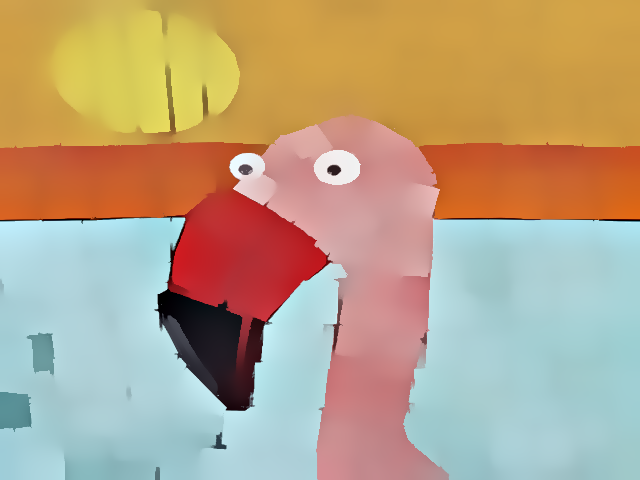}\\
        \vspace{-2mm}
		\caption*{\cite{zhang2023pyramid}}
	\end{subfigure}
    \begin{subfigure}[c]{0.161\textwidth}
		\centering
        \includegraphics[width=\linewidth]{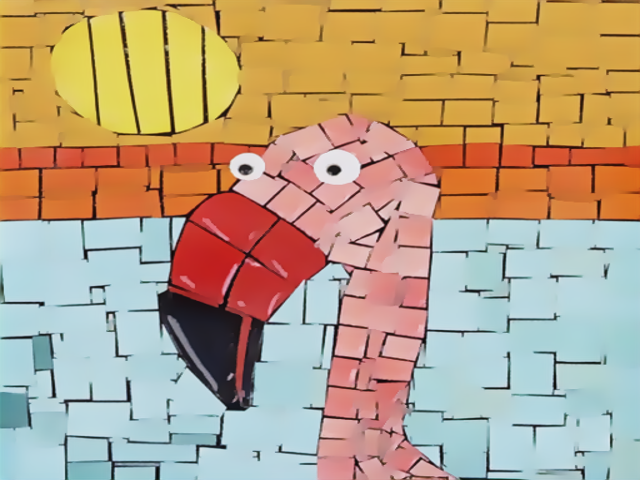}\\
        \vspace{-2mm}
		\caption*{\cite{zhang2025stf}}
	\end{subfigure}
    \begin{subfigure}[c]{0.161\textwidth}
		\centering
        \includegraphics[width=\linewidth]{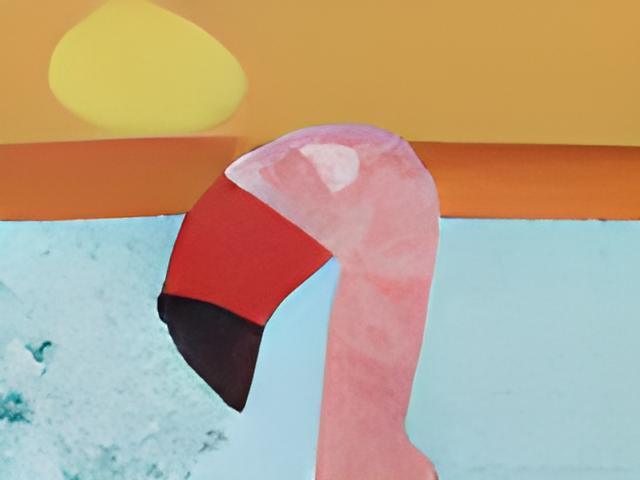}\\
        \vspace{-2mm}
		\caption*{Super-resolution model}
	\end{subfigure}
    \begin{subfigure}[c]{0.161\textwidth}
		\centering
        \includegraphics[width=\linewidth]{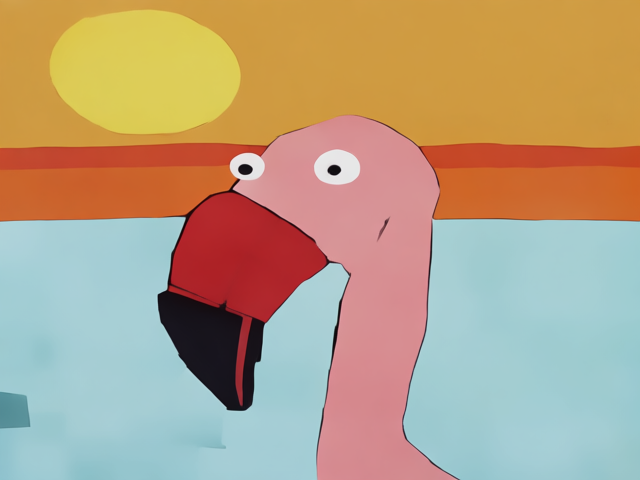}\\
        \vspace{-2mm}
		\caption*{Ours}
	\end{subfigure}
    \vspace{-3mm}
	\caption{Result of applying our adopted super-resolution model~\cite{wang2021real} to upsample the coarsest Gaussian pyramid level of the input image.} 
	\label{fig:ablation_sr}
    \vspace{-1mm}
\end{figure*}

\qing{\noindent \textbf{Ablation studies.} We conduct various ablation studies to evaluate the contribution of each design in our method. Specifically, we validate the necessity of the two-stage fine-tuning strategy in Figure~\ref{fig:two-stage-ablation}, demonstrate the effectiveness of our reward function in Figure~\ref{fig:reward_function}, examine in Figure~\ref{fig:ablation_sr} why our adopted super-resolution model is a better choice for upsampling in Eq.~\eqref{eq:reward_texture}, and also verify in Figure~\ref{fig:structure_preservation} why SSIM is not adopted for computing the structure preservation reward in Eq.~\eqref{eq:structure_reward}. Table~\ref{table:quantitative_comparison} also quantitatively demonstrates the effectiveness of our design by quantitative ablation studies. 
}

\begin{figure}[ht]
    \centering
    \begin{subfigure}[c]{0.32\linewidth}
		\centering
		\includegraphics[width=\textwidth]{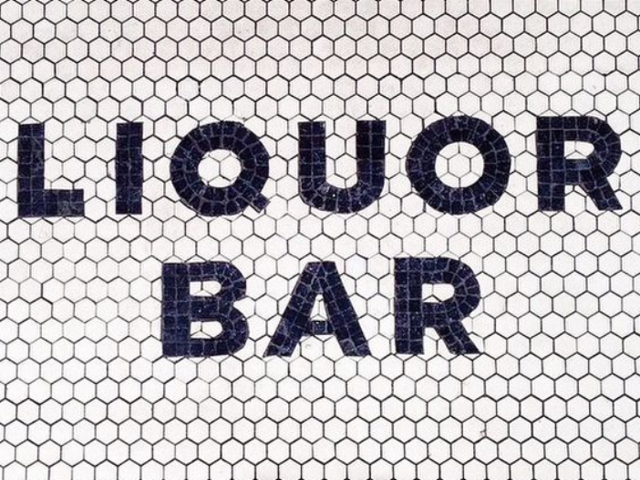} \\
        \vspace{-2mm}
		\caption*{Input}
    \end{subfigure}
    \begin{subfigure}[c]{0.32\linewidth}
		\centering
		\includegraphics[width=\textwidth]{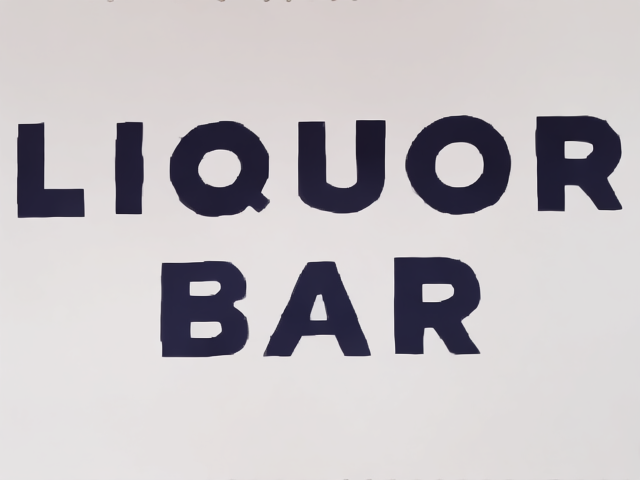} \\
        \vspace{-2mm}
		\caption*{Flux.1-Kontext }
    \end{subfigure}
    \begin{subfigure}[c]{0.32\linewidth}
		\centering
		\includegraphics[width=\textwidth]{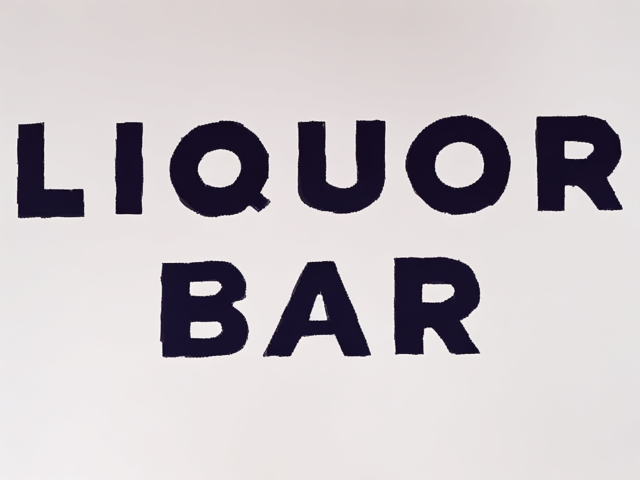} \\
        \vspace{-2mm}
		\caption*{Qwen-Image-Edit}
    \end{subfigure}
    \vspace{-3mm}
    \caption{\zrj{Effect of using different pre-trained generative models. Please see the supplementary for more results.}}
    \label{fig:different_models}
    \vspace{-2mm}
\end{figure}

\vspace{0.5em}
\qing{\noindent \textbf{Effect of using different pre-trained generative models.} Our texture filtering framework is applicable to any other open-source image generation foundation models. As shown in Figure~\ref{fig:different_models}, replacing Qwen-Image-Edit with Flux.1-Kontext also produces comparable result, demonstrating the scalability of our framework.} 

\begin{figure}[!t]
    \centering
    \begin{subfigure}[c]{0.24\linewidth}
		\centering
		\includegraphics[width=\textwidth]{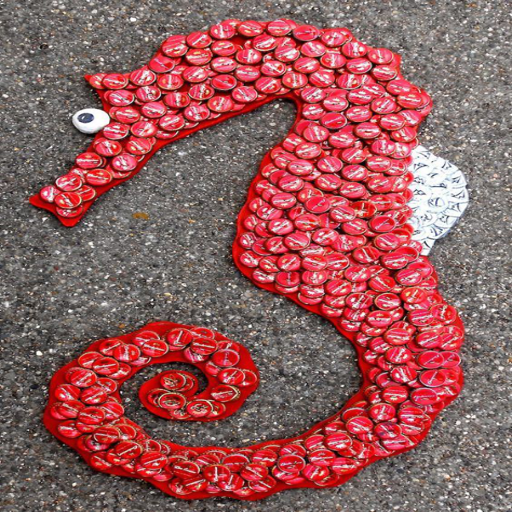} \\ \vspace{-2mm}
		\caption*{Input}
    \end{subfigure}
    \begin{subfigure}[c]{0.24\linewidth}
		\centering
		\includegraphics[width=\textwidth]{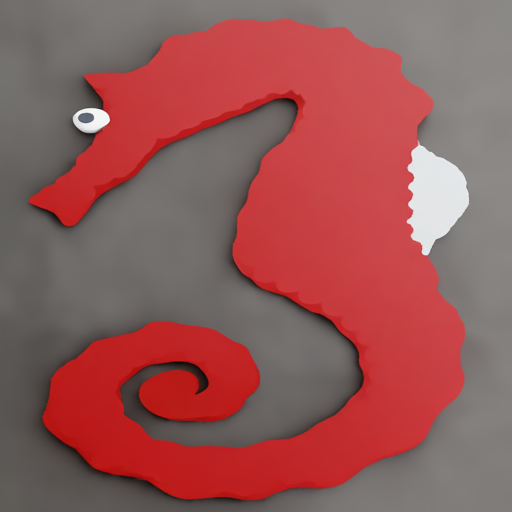} \\ \vspace{-2mm}
		\caption*{With RTV}
    \end{subfigure}
    \begin{subfigure}[c]{0.24\linewidth}
		\centering
		\includegraphics[width=\textwidth]{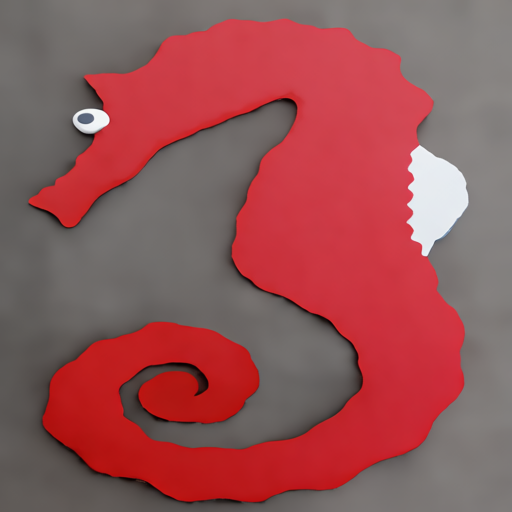} \\ \vspace{-2mm} 
		\caption*{With BTF}
    \end{subfigure}
    \begin{subfigure}[c]{0.24\linewidth}
		\centering
		\includegraphics[width=\textwidth]{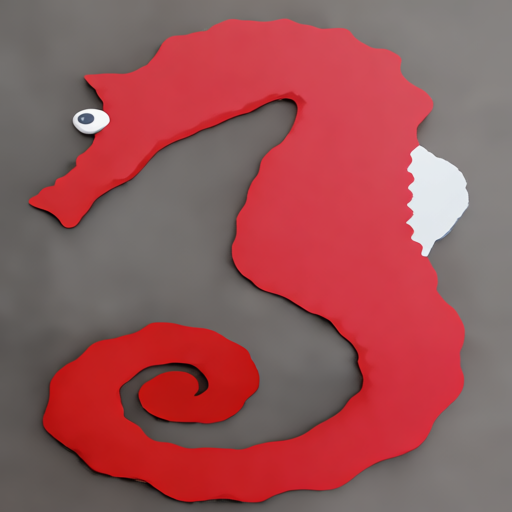} \\ \vspace{-2mm}
		\caption*{Ours}
    \end{subfigure}
    \vspace{-3mm}
    \caption{Effect of using different existing texture filtering methods to construct paired data for supervised fine-tuning.  }
    \label{fig:ablation_dataset_methods}
    \vspace{-2mm}
\end{figure}

\vspace{0.5em}
\qing{\noindent \textbf{Effect of training data scale for supervised fine-tuning.} We show in Table~\ref{table:quantitative_comparison}, as well as Figures~\ref{fig:ablation_size_dataset} and \ref{fig:ablation_dataset_methods} that the scale of the training data as well as the methods for constructing image pairs have very weak impact to the final texture filtering performance. 
}

\vspace{0.5em}
\qing{\noindent \textbf{Analysis of generalization ability.} As shown by visual comparisons in Figures~\ref{fig:synthetic_comparison} and \ref{fig:real_comparison}, as well as quantitative comparison in Table~\ref{table:quantitative_comparison}, our method demonstrates the best results on both the two unseen datasets, validating the generalizability of our method.
}


\begin{figure*}[!t]
	\centering
	\begin{subfigure}[c]{0.161\textwidth}
		\centering
        \includegraphics[width=\linewidth]{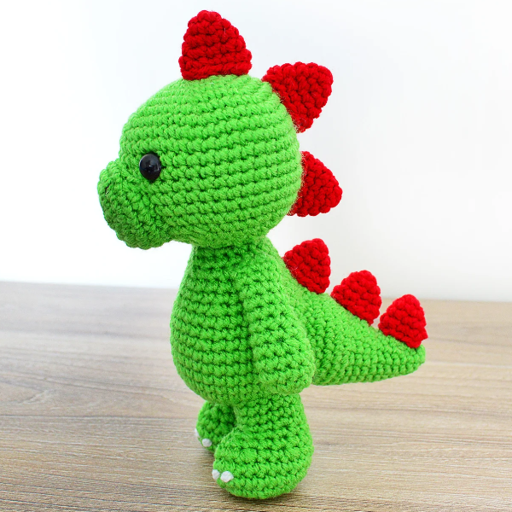}\\
        \vspace{-2mm}
		\caption*{Input}
	\end{subfigure}
    \begin{subfigure}[c]{0.161\textwidth}
		\centering
        \includegraphics[width=\linewidth]{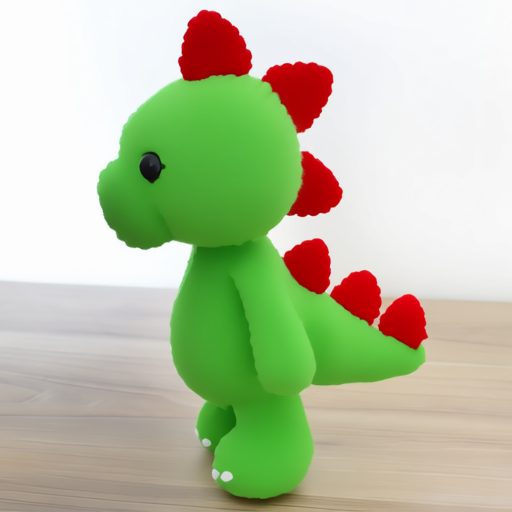}\\
        \vspace{-2mm}
		\caption*{$\lambda_1$=0.2, $\lambda_2$=0.6, $\lambda_3$=0.2}
	\end{subfigure}
    \begin{subfigure}[c]{0.161\textwidth}
		\centering
        \includegraphics[width=\linewidth]{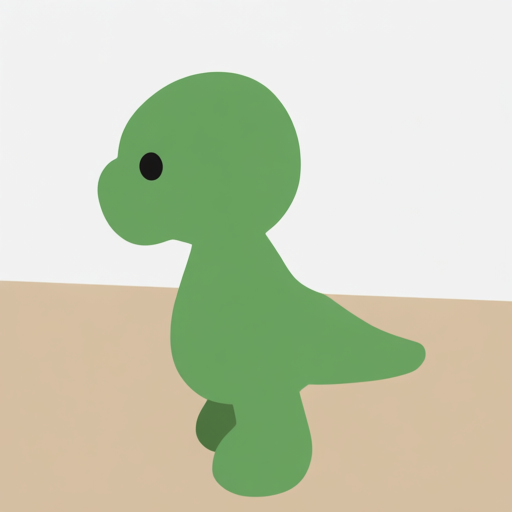}\\
        \vspace{-2mm}
		\caption*{$\lambda_1$=0.5, $\lambda_2$=0.6, $\lambda_3$=0.2}
	\end{subfigure}
    \begin{subfigure}[c]{0.161\textwidth}
		\centering
        \includegraphics[width=\linewidth]{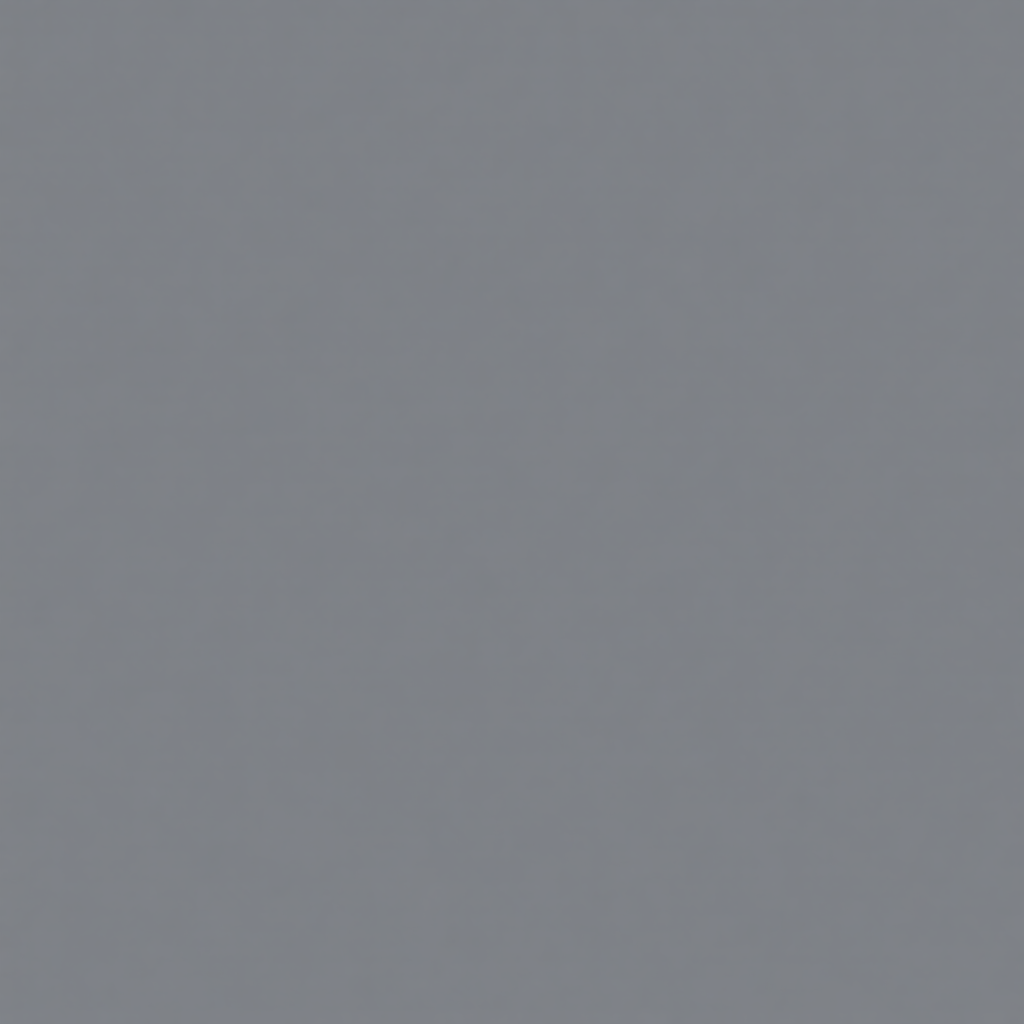}\\ 
        \vspace{-2mm}
		\caption*{$\lambda_1$=1.0, $\lambda_2$=0.6, $\lambda_3$=0.2}
	\end{subfigure}
    \begin{subfigure}[c]{0.161\textwidth}
		\centering
        \includegraphics[width=\linewidth]{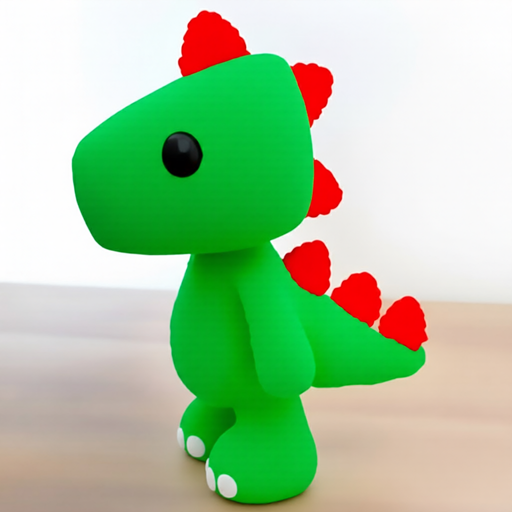}\\
        \vspace{-2mm}
		\caption*{$\lambda_1$=0.2, $\lambda_2$=0.1, $\lambda_3$=0.2}
	\end{subfigure}
    \begin{subfigure}[c]{0.161\textwidth}
		\centering
        \includegraphics[width=\linewidth]{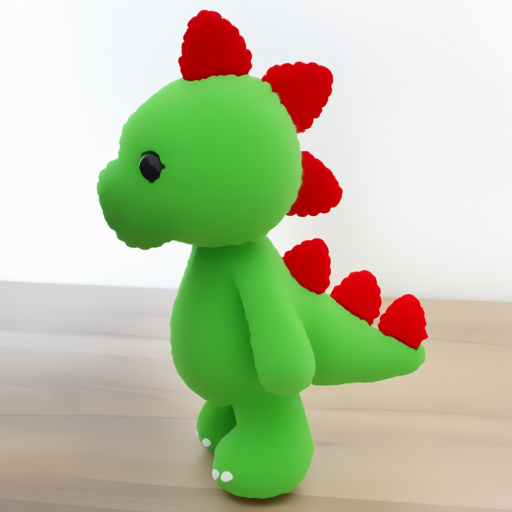}\\
        \vspace{-2mm}
		\caption*{$\lambda_1$=0.2, $\lambda_2$=0.6, $\lambda_3$=0.6}
	\end{subfigure}
    \vspace{-3mm}
	\caption{Results with different hyperparameter settings for the reward function. Note, We use $\lambda_1$=0.2, $\lambda_2$=0.6, $\lambda_3$=0.2 as our default setting in all experiments.} 
	\label{fig:different_param}
    \vspace{-1mm}
\end{figure*}

\vspace{0.5em} 
\noindent \textbf{Effect of different hyperparameters in reward function.}  As shown in Figure~\ref{fig:different_param}, larger $\lambda_1$ corresponds to stronger texture removal effect, and when setting $\lambda_1=1$, the model tends to generate a local optima in terms of an empty image without any content. On the other hand, higher $\lambda_2$ and $\lambda_3$ lead to better performance of structure preservation and image fidelity. 

\begin{figure}[ht]
    \centering
    \begin{subfigure}[c]{0.32\linewidth}
		\centering
		\includegraphics[width=\textwidth]{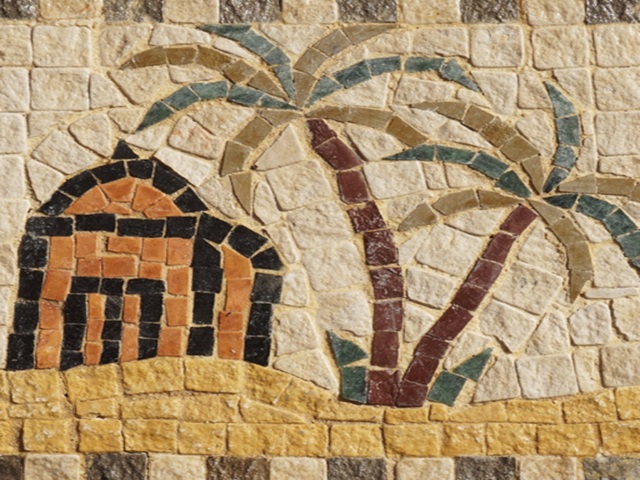} \\
        \vspace{-2mm}
		\caption*{Input}
    \end{subfigure}
    \begin{subfigure}[c]{0.32\linewidth}
		\centering
		\includegraphics[width=\textwidth]{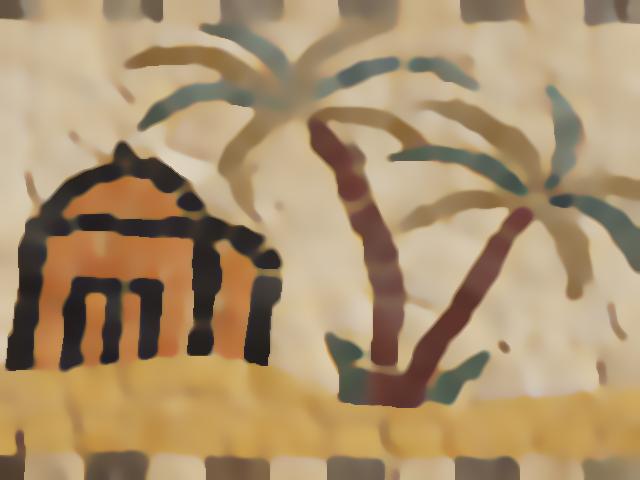} \\
        \vspace{-2mm}
		\caption*{Direct optimization}
    \end{subfigure}
    \begin{subfigure}[c]{0.32\linewidth}
		\centering
		\includegraphics[width=\textwidth]{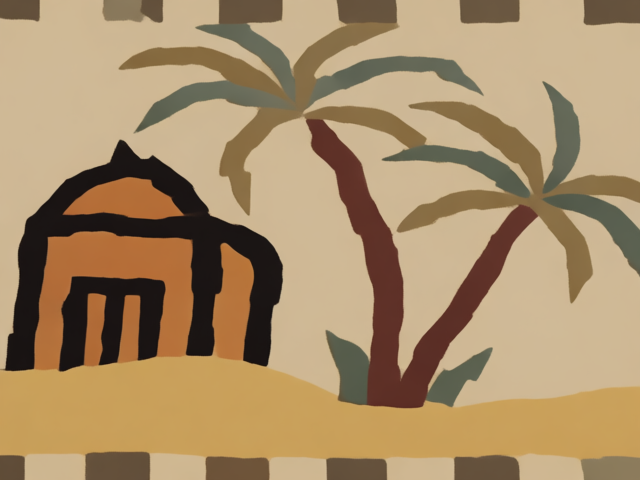} \\
        \vspace{-2mm}
		\caption*{Ours (with learning)}
    \end{subfigure}
    \vspace{-3mm}
    \caption{\zrj{Effect of direct optimization using our rewards without learning.}}
    \label{fig:optimization}
    \vspace{-2mm}
\end{figure}

\vspace{0.5em} 
\noindent \zrj{\textbf{Effect of direct optimization without learning.} As shown in Figure~\ref{fig:optimization}, though a direct optimization with our rewards without learning can also work, it fails to produce results as good as our method with two-stage training as the learning process allows us to leverage the powerful image prior learned by pre-trained models.}

\vspace{0.5em}
\qing{\noindent \textbf{Runtime performance.} We use 4 diffusion steps for inference on a single A800 GPU, which takes about 1.2 seconds to process a $512\times512$ image and about 3 seconds for a $1024\times1024$ image. This is more efficient than most previous methods. \zrj{Moreover, by incorporating distribution matching distillation (DMD)~\cite{yin2024one} and CacheDiT~\cite{cache-dit}, our method still has significant room for further acceleration.}
}

\begin{figure}[!t]
    \centering
    \begin{subfigure}[c]{0.325\linewidth}
		\centering
		\includegraphics[width=\textwidth]{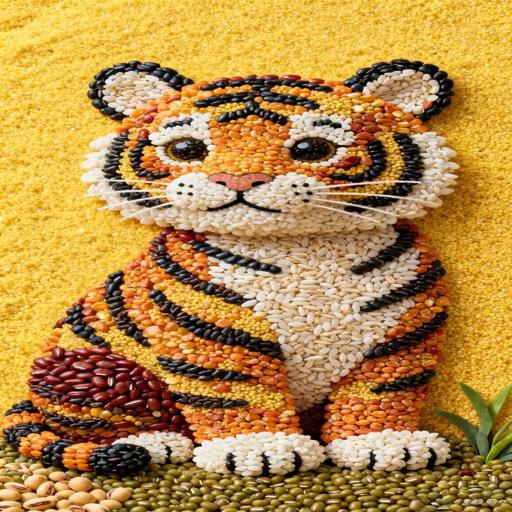} \\ \vspace{-2mm}
		\caption*{Input}
    \end{subfigure}
    \begin{subfigure}[c]{0.325\linewidth}
		\centering
		\includegraphics[width=\textwidth]{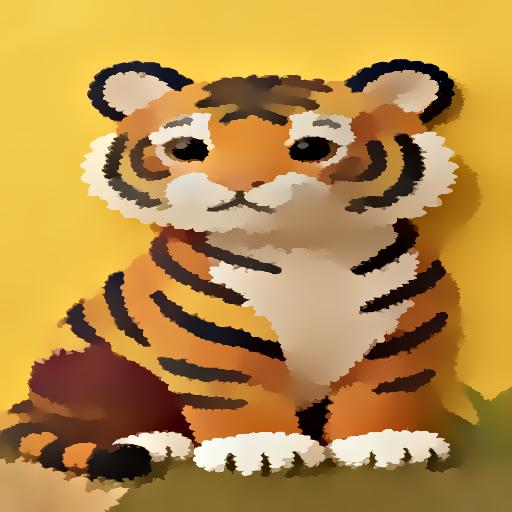} \\ \vspace{-2mm}
		\caption*{PTF \citep{zhang2023pyramid}}
    \end{subfigure}
    \begin{subfigure}[c]{0.325\linewidth}
		\centering
		\includegraphics[width=\textwidth]{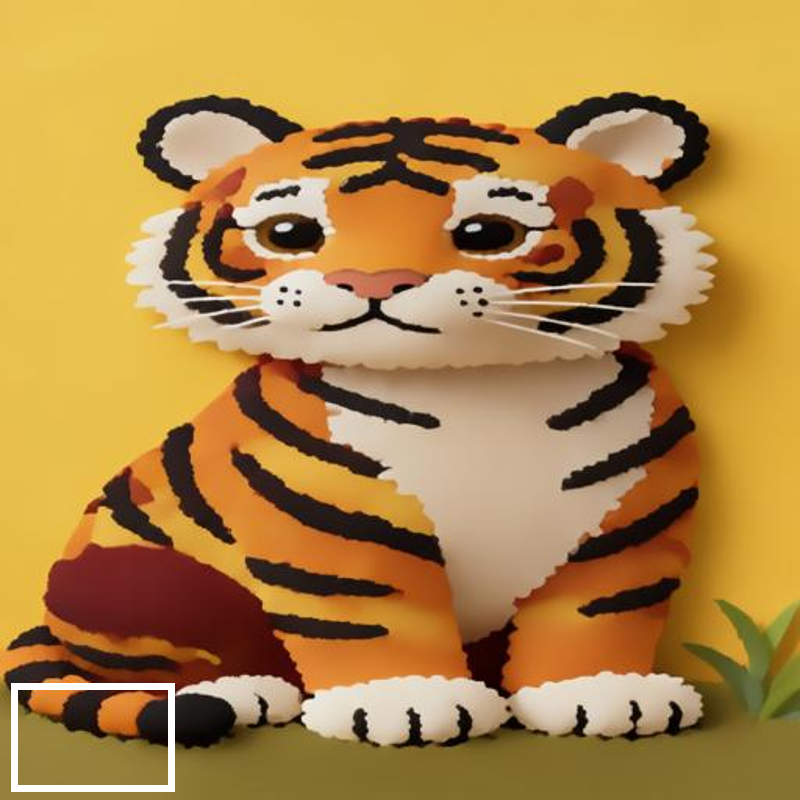} \\ \vspace{-2mm}
		\caption*{Ours}
    \end{subfigure}
    \vspace{-3mm}
    \caption{Failure case. Our method fails to preserve the color consistency for the bottom-left region (see the region indicated by the white box). }
    \label{fig:limitation}
    \vspace{-2mm}
\end{figure}

\begin{figure*}[!t]
	\centering
	\begin{subfigure}[c]{0.161\textwidth}
		\centering
        \includegraphics[width=\linewidth]{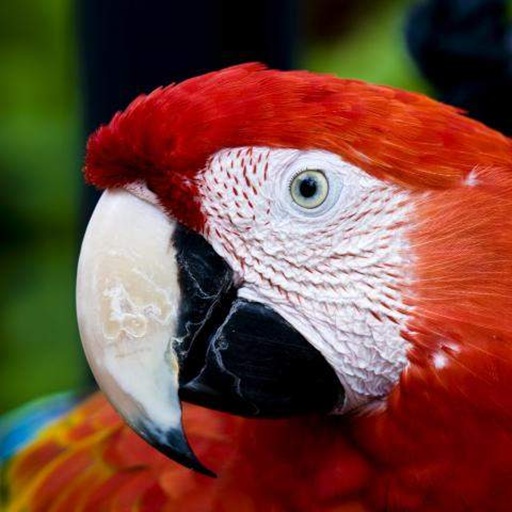}\\ \vspace{1pt}
		\includegraphics[width=\textwidth]{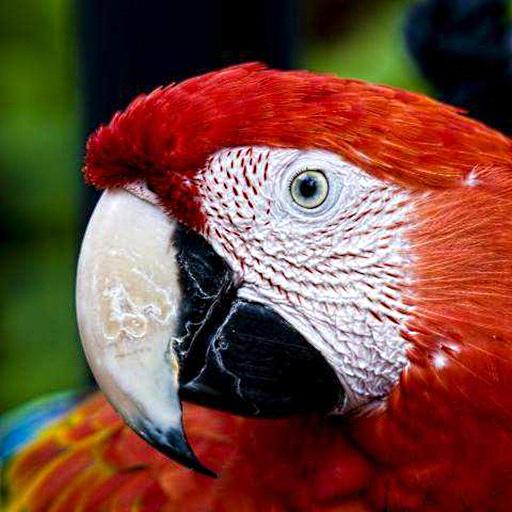} \\
        \vspace{-2mm}
		\caption*{Input \& WLS }
	\end{subfigure}
    \begin{subfigure}[c]{0.161\textwidth}
		\centering
        \includegraphics[width=\linewidth]{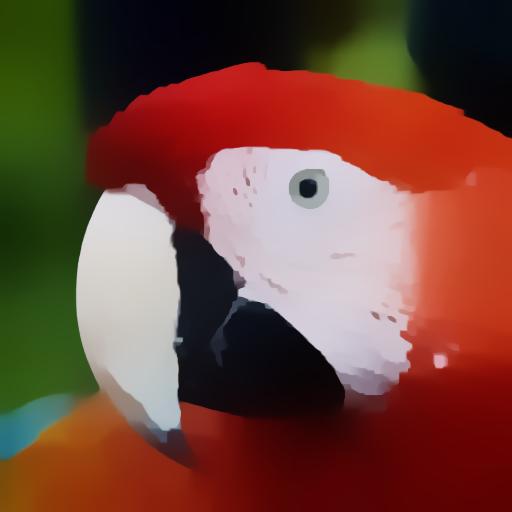}\\ \vspace{1pt}
		\includegraphics[width=\textwidth]{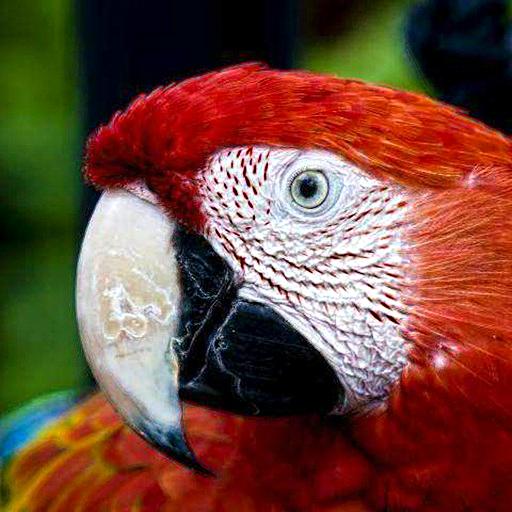} \\
        \vspace{-2mm}
		\caption*{\cite{xu2012structure-rtv}}
	\end{subfigure}
    \begin{subfigure}[c]{0.161\textwidth}
		\centering
        \includegraphics[width=\linewidth]{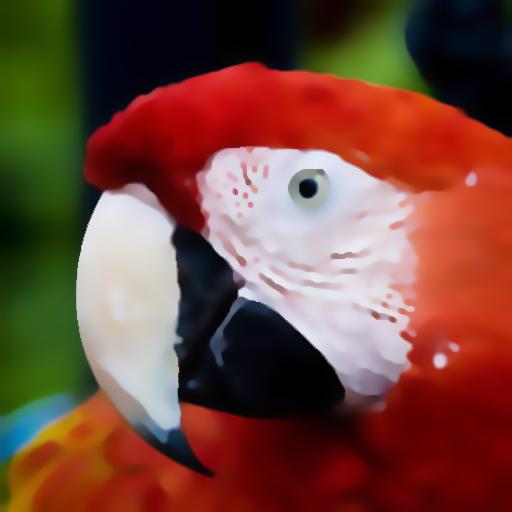}\\ \vspace{1pt}
		\includegraphics[width=\textwidth]{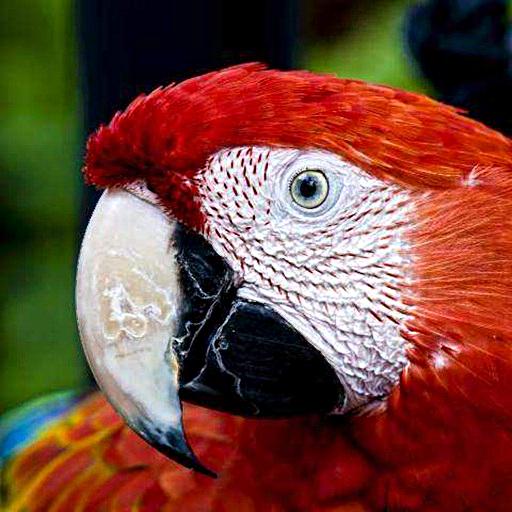} \\
        \vspace{-2mm}
		\caption*{\cite{cho2014bilateral-btf}}
	\end{subfigure}
    \begin{subfigure}[c]{0.161\textwidth}
		\centering
        \includegraphics[width=\linewidth]{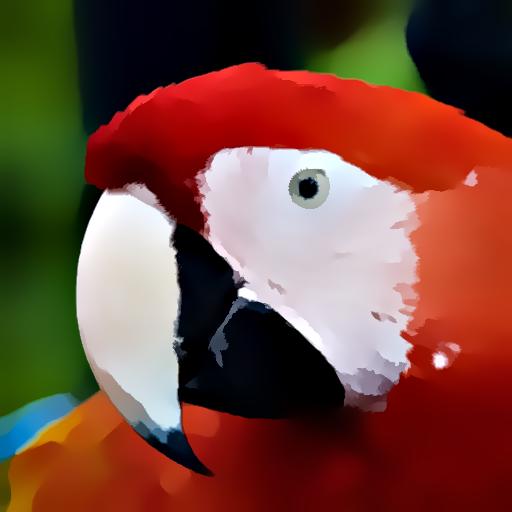}\\ \vspace{1pt}
		\includegraphics[width=\textwidth]{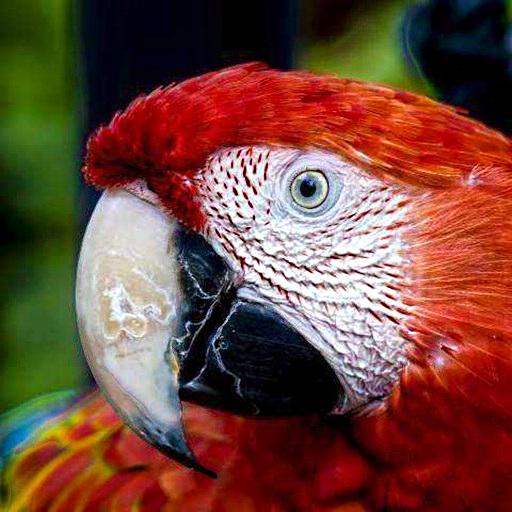} \\
        \vspace{-2mm}
		\caption*{\cite{zhang2023pyramid}}
	\end{subfigure}
    \begin{subfigure}[c]{0.161\textwidth}
		\centering
        \includegraphics[width=\linewidth]{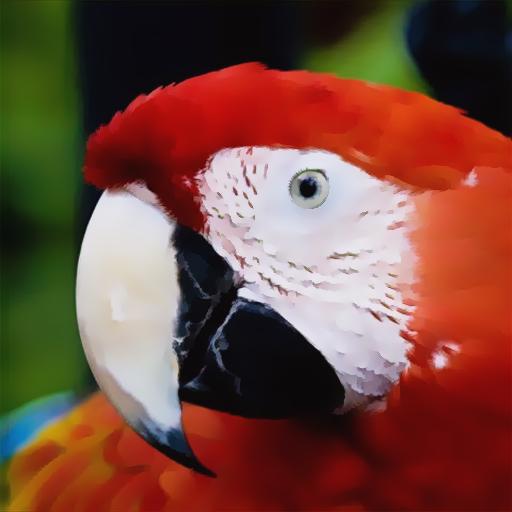}\\ \vspace{1pt}
		\includegraphics[width=\textwidth]{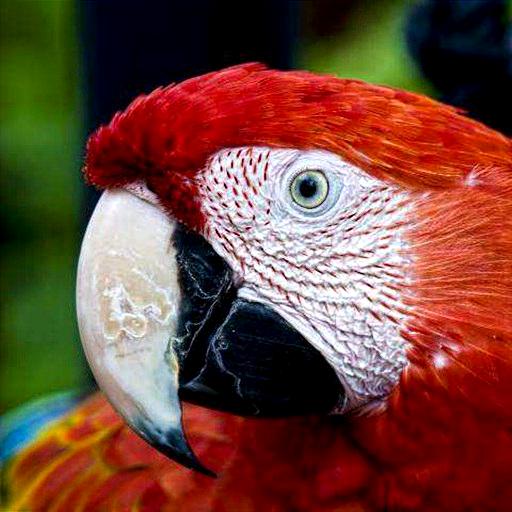} \\
        \vspace{-2mm}
		\caption*{\cite{zhang2025stf}}
	\end{subfigure}
    \begin{subfigure}[c]{0.161\textwidth}
		\centering
        \includegraphics[width=\linewidth]{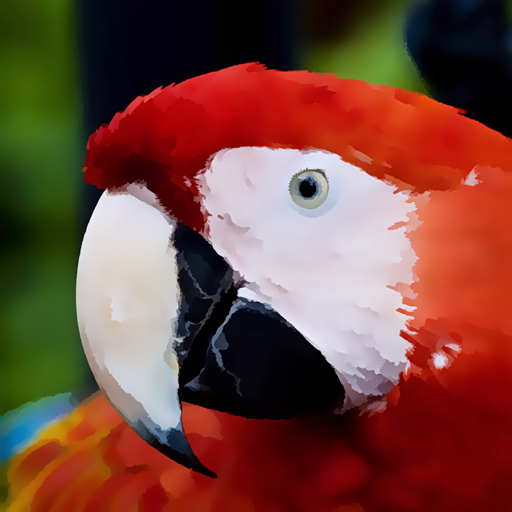}\\ \vspace{1pt}
		\includegraphics[width=\textwidth]{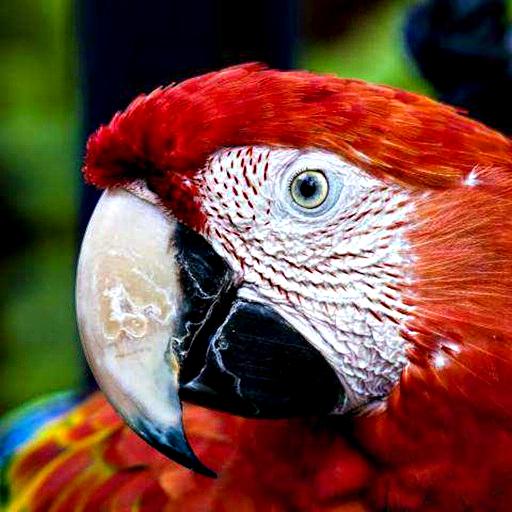} \\
        \vspace{-2mm}
		\caption*{Ours}
	\end{subfigure}
    \vspace{-3mm}
	\caption{Comparison with previous methods on detail enhancement. The top of the 1st column presents the input image, while the bottom gives the detail enhancement result of WLS \citep{farbman2008edge}. For other columns, the top and bottom shows the filtering and enhancement results, respectively. } 
	\label{fig:detail_enhancement}
    \vspace{-1mm}
\end{figure*}

\begin{figure*}[!t]
	\centering
	\begin{subfigure}[c]{0.161\textwidth}
		\centering
        \includegraphics[width=\linewidth]{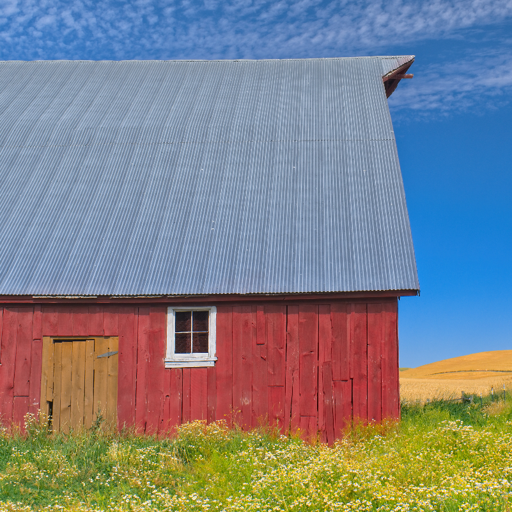}\\
        \vspace{-2mm}
		\caption*{Input}
	\end{subfigure}
    \begin{subfigure}[c]{0.161\textwidth}
		\centering
        \includegraphics[width=\linewidth]{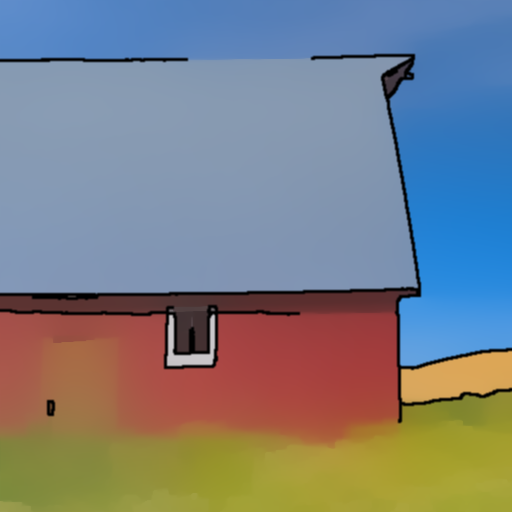}\\
        \vspace{-2mm}
		\caption*{\cite{xu2012structure-rtv}}
	\end{subfigure}
    \begin{subfigure}[c]{0.161\textwidth}
		\centering
        \includegraphics[width=\linewidth]{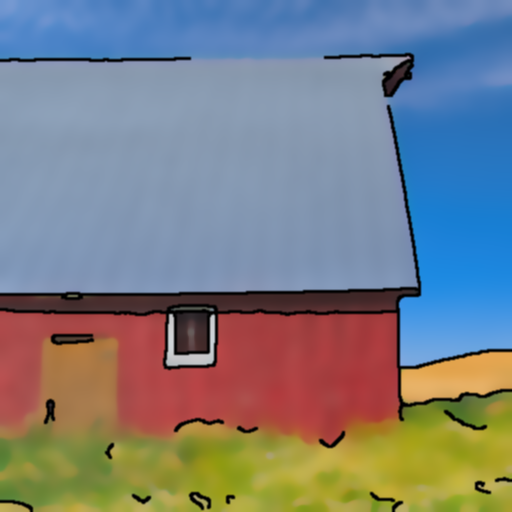}\\
        \vspace{-2mm}
		\caption*{\cite{cho2014bilateral-btf}}
	\end{subfigure}
    \begin{subfigure}[c]{0.161\textwidth}
		\centering
        \includegraphics[width=\linewidth]{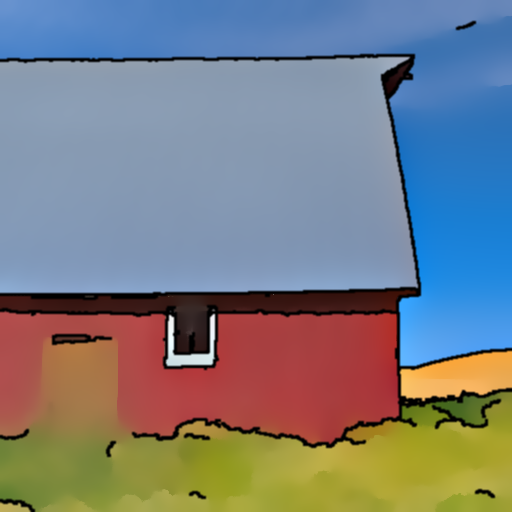}\\
        \vspace{-2mm}
		\caption*{\cite{zhang2023pyramid}}
	\end{subfigure}
    \begin{subfigure}[c]{0.161\textwidth}
		\centering
        \includegraphics[width=\linewidth]{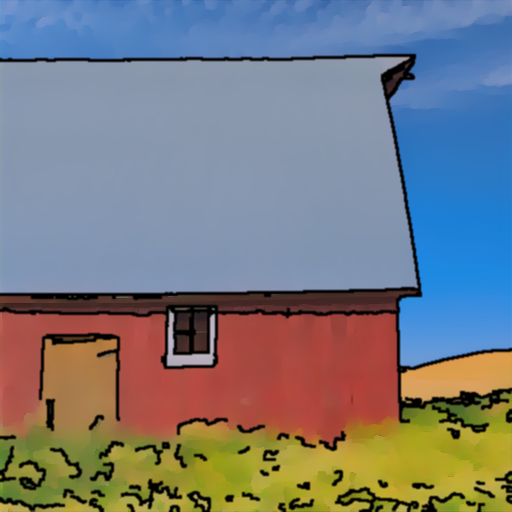}\\
        \vspace{-2mm}
		\caption*{\cite{zhang2025stf}}
	\end{subfigure}
    \begin{subfigure}[c]{0.161\textwidth}
		\centering
        \includegraphics[width=\linewidth]{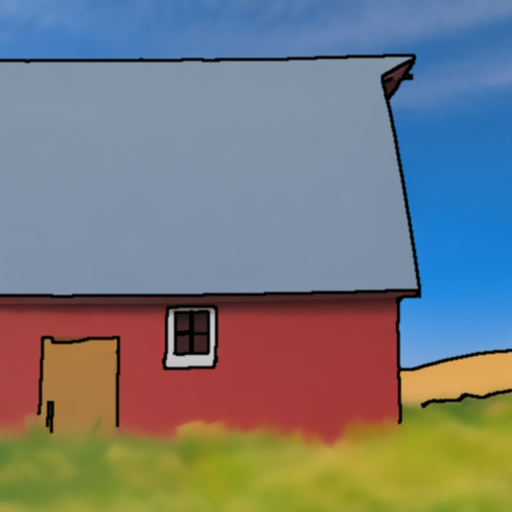}\\
        \vspace{-2mm}
		\caption*{Ours}
	\end{subfigure}
    \vspace{-3mm}
	\caption{Comparison with previous methods on image abstraction.} 
	\label{fig:image_abstraction}
    \vspace{-1mm}
\end{figure*}

\begin{figure*}[!t]
	\centering
	\begin{subfigure}[c]{0.161\textwidth}
		\centering
        \includegraphics[width=\linewidth]{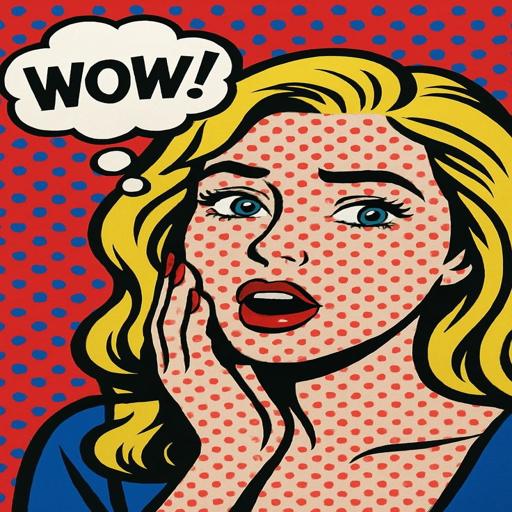}\\
        \vspace{-2mm}
		\caption*{Input}
	\end{subfigure}
    \begin{subfigure}[c]{0.161\textwidth}
		\centering
        \includegraphics[width=\linewidth]{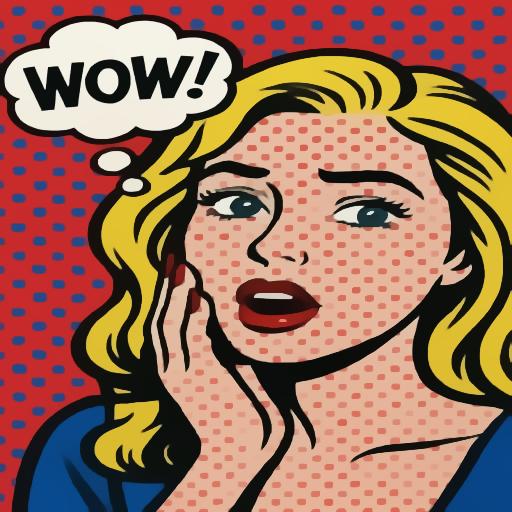}\\
        \vspace{-2mm}
		\caption*{\cite{xu2012structure-rtv}}
	\end{subfigure}
    \begin{subfigure}[c]{0.161\textwidth}
		\centering
        \includegraphics[width=\linewidth]{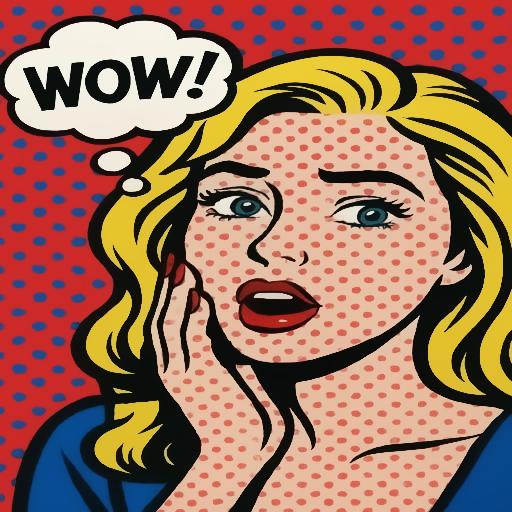}\\
        \vspace{-2mm}
		\caption*{\cite{cho2014bilateral-btf}}
	\end{subfigure}
    \begin{subfigure}[c]{0.161\textwidth}
		\centering
        \includegraphics[width=\linewidth]{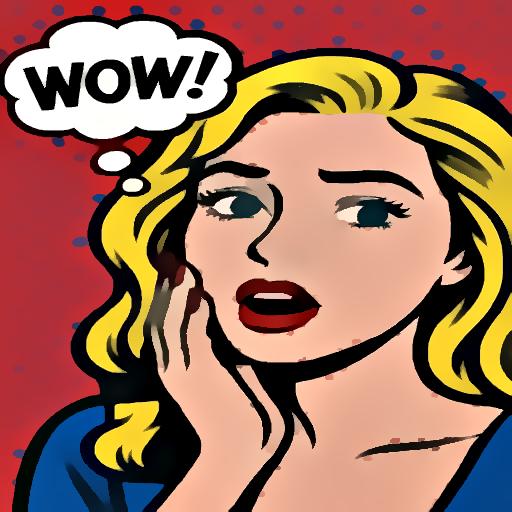}\\
        \vspace{-2mm}
		\caption*{\cite{zhang2023pyramid}}
	\end{subfigure}
    \begin{subfigure}[c]{0.161\textwidth}
		\centering
        \includegraphics[width=\linewidth]{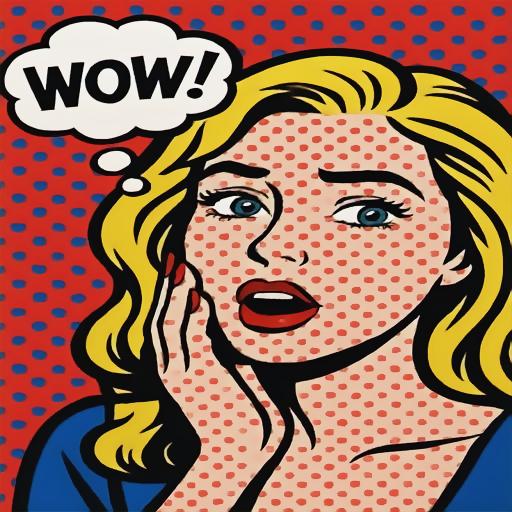}\\
        \vspace{-2mm}
		\caption*{\cite{zhang2025stf}}
	\end{subfigure}
    \begin{subfigure}[c]{0.161\textwidth}
		\centering
        \includegraphics[width=\linewidth]{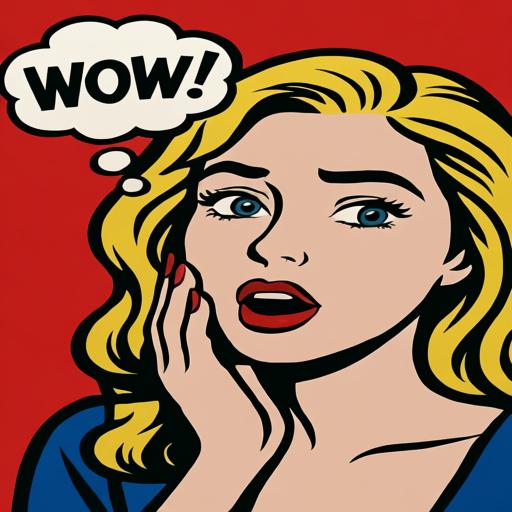}\\
        \vspace{-2mm}
		\caption*{Ours}
	\end{subfigure}
    \vspace{-3mm}
	\caption{Comparison with previous methods on inverse halftoning.} 
	\label{fig:inverse_halftoning}
    \vspace{-4mm}
\end{figure*}

\vspace{0.5em}
\qing{\noindent \textbf{Applications.} Our method also allows for various image manipulation applications, e.g., detail enhancement, image abstraction, and inverse-half toning. As shown in Figures~\ref{fig:detail_enhancement}-\ref{fig:inverse_halftoning}, our method produces impressive results for these three tasks, demonstrating the effectiveness and practicability of our method.
} 

\vspace{0.5em}
\noindent \textbf{Limitations and future work.} As shown in Figure~\ref{fig:limitation}, our method may erroneously remove some local color inconsistencies to obtain high-reward result as smooth as possible. Besides, it is still not so efficient. Addressing these two limitations will be our future work.

\section{Conclusion}
\qing{We have presented a generative texture filtering method. Unlike previous works, we propose to tackle the problem from an image generation perspective. Following this idea, we introduce a two-stage fine-tuning strategy for adapting a pre-trained generative model as a texture filtering method with strong performance and generalizability, and design a reward function to guide the reinforcement fine-tuning on large-scale unlabeled data. Extensive experiments demonstrate the effectiveness of our method. }

\bibliographystyle{ACM-Reference-Format}
\bibliography{paper_reference}


\begin{thebibliography}{49}


\ifx \showCODEN    \undefined \def \showCODEN     #1{\unskip}     \fi
\ifx \showISBNx    \undefined \def \showISBNx     #1{\unskip}     \fi
\ifx \showISBNxiii \undefined \def \showISBNxiii  #1{\unskip}     \fi
\ifx \showISSN     \undefined \def \showISSN      #1{\unskip}     \fi
\ifx \showLCCN     \undefined \def \showLCCN      #1{\unskip}     \fi
\ifx \shownote     \undefined \def \shownote      #1{#1}          \fi
\ifx \showarticletitle \undefined \def \showarticletitle #1{#1}   \fi
\ifx \showURL      \undefined \def \showURL       {\relax}        \fi
\providecommand\bibfield[2]{#2}
\providecommand\bibinfo[2]{#2}
\providecommand\natexlab[1]{#1}
\providecommand\showeprint[2][]{arXiv:#2}

\bibitem[Aubry et~al\mbox{.}(2014)]%
        {aubry2014fast}
\bibfield{author}{\bibinfo{person}{Mathieu Aubry}, \bibinfo{person}{Sylvain Paris}, \bibinfo{person}{Samuel~W Hasinoff}, \bibinfo{person}{Jan Kautz}, {and} \bibinfo{person}{Fr{\'e}do Durand}.} \bibinfo{year}{2014}\natexlab{}.
\newblock \showarticletitle{Fast local laplacian filters: Theory and applications}.
\newblock \bibinfo{journal}{\emph{ACM Transactions on Graphics (TOG)}} \bibinfo{volume}{33}, \bibinfo{number}{5} (\bibinfo{year}{2014}), \bibinfo{pages}{1--14}.
\newblock


\bibitem[Bao et~al\mbox{.}(2013)]%
        {bao2013tree}
\bibfield{author}{\bibinfo{person}{Linchao Bao}, \bibinfo{person}{Yibing Song}, \bibinfo{person}{Qingxiong Yang}, \bibinfo{person}{Hao Yuan}, {and} \bibinfo{person}{Gang Wang}.} \bibinfo{year}{2013}\natexlab{}.
\newblock \showarticletitle{Tree filtering: Efficient structure-preserving smoothing with a minimum spanning tree}.
\newblock \bibinfo{journal}{\emph{IEEE Transactions on Image Processing}} \bibinfo{volume}{23}, \bibinfo{number}{2} (\bibinfo{year}{2013}), \bibinfo{pages}{555--569}.
\newblock


\bibitem[Bi et~al\mbox{.}(2015)]%
        {bi20151}
\bibfield{author}{\bibinfo{person}{Sai Bi}, \bibinfo{person}{Xiaoguang Han}, {and} \bibinfo{person}{Yizhou Yu}.} \bibinfo{year}{2015}\natexlab{}.
\newblock \showarticletitle{An L1 image transform for edge-preserving smoothing and scene-level intrinsic decomposition}.
\newblock \bibinfo{journal}{\emph{ACM Transactions on Graphics}} \bibinfo{volume}{34}, \bibinfo{number}{4} (\bibinfo{year}{2015}), \bibinfo{pages}{1--12}.
\newblock


\bibitem[Chen et~al\mbox{.}(2007)]%
        {chen2007real}
\bibfield{author}{\bibinfo{person}{Jiawen Chen}, \bibinfo{person}{Sylvain Paris}, {and} \bibinfo{person}{Fr{\'e}do Durand}.} \bibinfo{year}{2007}\natexlab{}.
\newblock \showarticletitle{Real-time edge-aware image processing with the bilateral grid}.
\newblock \bibinfo{journal}{\emph{ACM Transactions on Graphics}} \bibinfo{volume}{26}, \bibinfo{number}{3} (\bibinfo{year}{2007}), \bibinfo{pages}{103}.
\newblock


\bibitem[Cho et~al\mbox{.}(2014)]%
        {cho2014bilateral-btf}
\bibfield{author}{\bibinfo{person}{Hojin Cho}, \bibinfo{person}{Hyunjoon Lee}, \bibinfo{person}{Henry Kang}, {and} \bibinfo{person}{Seungyong Lee}.} \bibinfo{year}{2014}\natexlab{}.
\newblock \showarticletitle{Bilateral texture filtering}.
\newblock \bibinfo{journal}{\emph{ACM Transactions on Graphics (TOG)}} \bibinfo{volume}{33}, \bibinfo{number}{4} (\bibinfo{year}{2014}), \bibinfo{pages}{1--8}.
\newblock


\bibitem[Comanici et~al\mbox{.}(2025)]%
        {comanici2025gemini}
\bibfield{author}{\bibinfo{person}{Gheorghe Comanici}, \bibinfo{person}{Eric Bieber}, \bibinfo{person}{Mike Schaekermann}, \bibinfo{person}{Ice Pasupat}, \bibinfo{person}{Noveen Sachdeva}, \bibinfo{person}{Inderjit Dhillon}, \bibinfo{person}{Marcel Blistein}, \bibinfo{person}{Ori Ram}, \bibinfo{person}{Dan Zhang}, \bibinfo{person}{Evan Rosen}, {et~al\mbox{.}}} \bibinfo{year}{2025}\natexlab{}.
\newblock \showarticletitle{Gemini 2.5: Pushing the frontier with advanced reasoning, multimodality, long context, and next generation agentic capabilities}.
\newblock \bibinfo{journal}{\emph{arXiv preprint arXiv:2507.06261}} (\bibinfo{year}{2025}).
\newblock


\bibitem[Criminisi et~al\mbox{.}(2010)]%
        {criminisi2010geodesic}
\bibfield{author}{\bibinfo{person}{Antonio Criminisi}, \bibinfo{person}{Toby Sharp}, \bibinfo{person}{Carsten Rother}, {and} \bibinfo{person}{Patrick P{\'e}rez}.} \bibinfo{year}{2010}\natexlab{}.
\newblock \showarticletitle{Geodesic image and video editing.}
\newblock \bibinfo{journal}{\emph{ACM Transactions on Graphics}} \bibinfo{volume}{29}, \bibinfo{number}{5} (\bibinfo{year}{2010}), \bibinfo{pages}{134--1}.
\newblock


\bibitem[Du et~al\mbox{.}(2016)]%
        {du2016two}
\bibfield{author}{\bibinfo{person}{Hui Du}, \bibinfo{person}{Xiaogang Jin}, {and} \bibinfo{person}{Philip~J Willis}.} \bibinfo{year}{2016}\natexlab{}.
\newblock \showarticletitle{Two-level joint local laplacian texture filtering}.
\newblock \bibinfo{journal}{\emph{The Visual Computer}} \bibinfo{volume}{32}, \bibinfo{number}{12} (\bibinfo{year}{2016}), \bibinfo{pages}{1537--1548}.
\newblock


\bibitem[Durand and Dorsey(2002)]%
        {durand2002fast}
\bibfield{author}{\bibinfo{person}{Fr{\'e}do Durand} {and} \bibinfo{person}{Julie Dorsey}.} \bibinfo{year}{2002}\natexlab{}.
\newblock \showarticletitle{Fast bilateral filtering for the display of high-dynamic-range images}.
\newblock \bibinfo{journal}{\emph{ACM Transactions on Graphics}} \bibinfo{volume}{21}, \bibinfo{number}{3} (\bibinfo{year}{2002}), \bibinfo{pages}{257--266}.
\newblock


\bibitem[Fan et~al\mbox{.}(2018)]%
        {fan2018image}
\bibfield{author}{\bibinfo{person}{Qingnan Fan}, \bibinfo{person}{Jiaolong Yang}, \bibinfo{person}{David Wipf}, \bibinfo{person}{Baoquan Chen}, {and} \bibinfo{person}{Xin Tong}.} \bibinfo{year}{2018}\natexlab{}.
\newblock \showarticletitle{Image smoothing via unsupervised learning}.
\newblock \bibinfo{journal}{\emph{ACM Transactions on Graphics}} \bibinfo{volume}{37}, \bibinfo{number}{6} (\bibinfo{year}{2018}), \bibinfo{pages}{1--14}.
\newblock


\bibitem[Farbman et~al\mbox{.}(2008)]%
        {farbman2008edge}
\bibfield{author}{\bibinfo{person}{Zeev Farbman}, \bibinfo{person}{Raanan Fattal}, \bibinfo{person}{Dani Lischinski}, {and} \bibinfo{person}{Richard Szeliski}.} \bibinfo{year}{2008}\natexlab{}.
\newblock \showarticletitle{Edge-preserving decompositions for multi-scale tone and detail manipulation}.
\newblock \bibinfo{journal}{\emph{ACM Transactions on Graphics}} \bibinfo{volume}{27}, \bibinfo{number}{3} (\bibinfo{year}{2008}), \bibinfo{pages}{1--10}.
\newblock


\bibitem[Fattal(2009)]%
        {fattal2009edge}
\bibfield{author}{\bibinfo{person}{Raanan Fattal}.} \bibinfo{year}{2009}\natexlab{}.
\newblock \showarticletitle{Edge-avoiding wavelets and their applications}.
\newblock \bibinfo{journal}{\emph{ACM Transactions on Graphics}} \bibinfo{volume}{28}, \bibinfo{number}{3} (\bibinfo{year}{2009}), \bibinfo{pages}{1--10}.
\newblock


\bibitem[Gastal and Oliveira(2011)]%
        {gastal2011domain}
\bibfield{author}{\bibinfo{person}{Eduardo~SL Gastal} {and} \bibinfo{person}{Manuel~M Oliveira}.} \bibinfo{year}{2011}\natexlab{}.
\newblock \showarticletitle{Domain transform for edge-aware image and video processing}.
\newblock \bibinfo{journal}{\emph{ACM Transactions on Graphics}} \bibinfo{volume}{30}, \bibinfo{number}{4} (\bibinfo{year}{2011}), \bibinfo{pages}{1--12}.
\newblock


\bibitem[Gastal and Oliveira(2012)]%
        {gastal2012adaptive}
\bibfield{author}{\bibinfo{person}{Eduardo~SL Gastal} {and} \bibinfo{person}{Manuel~M Oliveira}.} \bibinfo{year}{2012}\natexlab{}.
\newblock \showarticletitle{Adaptive manifolds for real-time high-dimensional filtering}.
\newblock \bibinfo{journal}{\emph{ACM Transactions on Graphics}} \bibinfo{volume}{31}, \bibinfo{number}{4} (\bibinfo{year}{2012}), \bibinfo{pages}{1--13}.
\newblock


\bibitem[Hu et~al\mbox{.}(2022)]%
        {hu2022lora}
\bibfield{author}{\bibinfo{person}{Edward~J Hu}, \bibinfo{person}{Yelong Shen}, \bibinfo{person}{Phillip Wallis}, \bibinfo{person}{Zeyuan Allen-Zhu}, \bibinfo{person}{Yuanzhi Li}, \bibinfo{person}{Shean Wang}, \bibinfo{person}{Lu Wang}, \bibinfo{person}{Weizhu Chen}, {et~al\mbox{.}}} \bibinfo{year}{2022}\natexlab{}.
\newblock \showarticletitle{Lora: Low-rank adaptation of large language models.}
\newblock \bibinfo{journal}{\emph{ICLR}} \bibinfo{volume}{1}, \bibinfo{number}{2} (\bibinfo{year}{2022}), \bibinfo{pages}{3}.
\newblock


\bibitem[Hurst et~al\mbox{.}(2024)]%
        {hurst2024gpt}
\bibfield{author}{\bibinfo{person}{Aaron Hurst}, \bibinfo{person}{Adam Lerer}, \bibinfo{person}{Adam~P Goucher}, \bibinfo{person}{Adam Perelman}, \bibinfo{person}{Aditya Ramesh}, \bibinfo{person}{Aidan Clark}, \bibinfo{person}{AJ Ostrow}, \bibinfo{person}{Akila Welihinda}, \bibinfo{person}{Alan Hayes}, \bibinfo{person}{Alec Radford}, {et~al\mbox{.}}} \bibinfo{year}{2024}\natexlab{}.
\newblock \showarticletitle{Gpt-4o system card}.
\newblock \bibinfo{journal}{\emph{arXiv preprint arXiv:2410.21276}} (\bibinfo{year}{2024}).
\newblock


\bibitem[Jeon et~al\mbox{.}(2016)]%
        {jeon2016scale}
\bibfield{author}{\bibinfo{person}{Junho Jeon}, \bibinfo{person}{Hyunjoon Lee}, \bibinfo{person}{Henry Kang}, {and} \bibinfo{person}{Seungyong Lee}.} \bibinfo{year}{2016}\natexlab{}.
\newblock \showarticletitle{Scale-aware structure-preserving texture filtering}.
\newblock \bibinfo{journal}{\emph{Computer Graphics Forum}} \bibinfo{volume}{35}, \bibinfo{number}{7} (\bibinfo{year}{2016}), \bibinfo{pages}{77--86}.
\newblock


\bibitem[Jiang et~al\mbox{.}(2025)]%
        {zhang2025stf}
\bibfield{author}{\bibinfo{person}{Hao Jiang}, \bibinfo{person}{Rongjia Zheng}, \bibinfo{person}{Yongwei Nie}, \bibinfo{person}{Chunxia Xiao}, \bibinfo{person}{Wei-Shi Zheng}, {and} \bibinfo{person}{Qing Zhang}.} \bibinfo{year}{2025}\natexlab{}.
\newblock \showarticletitle{Self-supervised Texture Filtering}.
\newblock \bibinfo{journal}{\emph{ACM Transactions on Graphics}} \bibinfo{volume}{44}, \bibinfo{number}{5} (\bibinfo{year}{2025}), \bibinfo{numpages}{13}~pages.
\newblock


\bibitem[Karacan et~al\mbox{.}(2013)]%
        {karacan2013structure}
\bibfield{author}{\bibinfo{person}{Levent Karacan}, \bibinfo{person}{Erkut Erdem}, {and} \bibinfo{person}{Aykut Erdem}.} \bibinfo{year}{2013}\natexlab{}.
\newblock \showarticletitle{Structure-preserving image smoothing via region covariances}.
\newblock \bibinfo{journal}{\emph{ACM Transactions on Graphics}} \bibinfo{volume}{32}, \bibinfo{number}{6} (\bibinfo{year}{2013}), \bibinfo{pages}{1--11}.
\newblock


\bibitem[Kass and Solomon(2010)]%
        {kass2010smoothed}
\bibfield{author}{\bibinfo{person}{Michael Kass} {and} \bibinfo{person}{Justin Solomon}.} \bibinfo{year}{2010}\natexlab{}.
\newblock \showarticletitle{Smoothed local histogram filters}.
\newblock \bibinfo{journal}{\emph{ACM Transactions on Graphics}} \bibinfo{volume}{29}, \bibinfo{number}{4} (\bibinfo{year}{2010}), \bibinfo{pages}{1--10}.
\newblock


\bibitem[Kim et~al\mbox{.}(2018)]%
        {kim2018structure}
\bibfield{author}{\bibinfo{person}{Youngjung Kim}, \bibinfo{person}{Bumsub Ham}, \bibinfo{person}{Minh~N Do}, {and} \bibinfo{person}{Kwanghoon Sohn}.} \bibinfo{year}{2018}\natexlab{}.
\newblock \showarticletitle{Structure-texture image decomposition using deep variational priors}.
\newblock \bibinfo{journal}{\emph{IEEE Transactions on Image Processing}} \bibinfo{volume}{28}, \bibinfo{number}{6} (\bibinfo{year}{2018}), \bibinfo{pages}{2692--2704}.
\newblock


\bibitem[Kopf et~al\mbox{.}(2007)]%
        {kopf2007joint}
\bibfield{author}{\bibinfo{person}{Johannes Kopf}, \bibinfo{person}{Michael~F Cohen}, \bibinfo{person}{Dani Lischinski}, {and} \bibinfo{person}{Matt Uyttendaele}.} \bibinfo{year}{2007}\natexlab{}.
\newblock \showarticletitle{Joint bilateral upsampling}.
\newblock \bibinfo{journal}{\emph{ACM Transactions on Graphics}} \bibinfo{volume}{26}, \bibinfo{number}{3} (\bibinfo{year}{2007}), \bibinfo{pages}{96}.
\newblock


\bibitem[Labs et~al\mbox{.}(2025)]%
        {labs2025flux}
\bibfield{author}{\bibinfo{person}{Black~Forest Labs}, \bibinfo{person}{Stephen Batifol}, \bibinfo{person}{Andreas Blattmann}, \bibinfo{person}{Frederic Boesel}, \bibinfo{person}{Saksham Consul}, \bibinfo{person}{Cyril Diagne}, \bibinfo{person}{Tim Dockhorn}, \bibinfo{person}{Jack English}, \bibinfo{person}{Zion English}, \bibinfo{person}{Patrick Esser}, {et~al\mbox{.}}} \bibinfo{year}{2025}\natexlab{}.
\newblock \showarticletitle{FLUX. 1 Kontext: Flow Matching for In-Context Image Generation and Editing in Latent Space}.
\newblock \bibinfo{journal}{\emph{arXiv preprint arXiv:2506.15742}} (\bibinfo{year}{2025}).
\newblock


\bibitem[Lipman et~al\mbox{.}(2022)]%
        {lipman2022flow}
\bibfield{author}{\bibinfo{person}{Yaron Lipman}, \bibinfo{person}{Ricky~TQ Chen}, \bibinfo{person}{Heli Ben-Hamu}, \bibinfo{person}{Maximilian Nickel}, {and} \bibinfo{person}{Matt Le}.} \bibinfo{year}{2022}\natexlab{}.
\newblock \showarticletitle{Flow matching for generative modeling}.
\newblock \bibinfo{journal}{\emph{arXiv preprint arXiv:2210.02747}} (\bibinfo{year}{2022}).
\newblock


\bibitem[Liu et~al\mbox{.}(2016)]%
        {liu2016learning}
\bibfield{author}{\bibinfo{person}{Sifei Liu}, \bibinfo{person}{Jinshan Pan}, {and} \bibinfo{person}{Ming-Hsuan Yang}.} \bibinfo{year}{2016}\natexlab{}.
\newblock \showarticletitle{Learning recursive filters for low-level vision via a hybrid neural network}. In \bibinfo{booktitle}{\emph{ECCV}}. \bibinfo{pages}{560--576}.
\newblock


\bibitem[Liu et~al\mbox{.}(2020)]%
        {liu2020real}
\bibfield{author}{\bibinfo{person}{Wei Liu}, \bibinfo{person}{Pingping Zhang}, \bibinfo{person}{Xiaolin Huang}, \bibinfo{person}{Jie Yang}, \bibinfo{person}{Chunhua Shen}, {and} \bibinfo{person}{Ian Reid}.} \bibinfo{year}{2020}\natexlab{}.
\newblock \showarticletitle{Real-time image smoothing via iterative least squares}.
\newblock \bibinfo{journal}{\emph{ACM Transactions on Graphics}} \bibinfo{volume}{39}, \bibinfo{number}{3} (\bibinfo{year}{2020}), \bibinfo{pages}{1--24}.
\newblock


\bibitem[Liu et~al\mbox{.}(2021)]%
        {liu2021generalized}
\bibfield{author}{\bibinfo{person}{Wei Liu}, \bibinfo{person}{Pingping Zhang}, \bibinfo{person}{Yinjie Lei}, \bibinfo{person}{Xiaolin Huang}, \bibinfo{person}{Jie Yang}, {and} \bibinfo{person}{Michael Ng}.} \bibinfo{year}{2021}\natexlab{}.
\newblock \showarticletitle{A generalized framework for edge-preserving and structure-preserving image smoothing}.
\newblock \bibinfo{journal}{\emph{IEEE Transactions on Pattern Analysis and Machine Intelligence}} \bibinfo{volume}{44}, \bibinfo{number}{10} (\bibinfo{year}{2021}), \bibinfo{pages}{6631--6648}.
\newblock


\bibitem[Liu et~al\mbox{.}(2022)]%
        {liu2022flow}
\bibfield{author}{\bibinfo{person}{Xingchao Liu}, \bibinfo{person}{Chengyue Gong}, {and} \bibinfo{person}{Qiang Liu}.} \bibinfo{year}{2022}\natexlab{}.
\newblock \showarticletitle{Flow straight and fast: Learning to generate and transfer data with rectified flow}.
\newblock \bibinfo{journal}{\emph{arXiv preprint arXiv:2209.03003}} (\bibinfo{year}{2022}).
\newblock


\bibitem[Lu et~al\mbox{.}(2022)]%
        {lu2022dpm}
\bibfield{author}{\bibinfo{person}{Cheng Lu}, \bibinfo{person}{Yuhao Zhou}, \bibinfo{person}{Fan Bao}, \bibinfo{person}{Jianfei Chen}, \bibinfo{person}{Chongxuan Li}, {and} \bibinfo{person}{Jun Zhu}.} \bibinfo{year}{2022}\natexlab{}.
\newblock \showarticletitle{Dpm-solver: A fast ode solver for diffusion probabilistic model sampling in around 10 steps}.
\newblock \bibinfo{journal}{\emph{NeurIPS}}  \bibinfo{volume}{35} (\bibinfo{year}{2022}), \bibinfo{pages}{5775--5787}.
\newblock


\bibitem[Lu et~al\mbox{.}(2018)]%
        {lu2018deep}
\bibfield{author}{\bibinfo{person}{Kaiyue Lu}, \bibinfo{person}{Shaodi You}, {and} \bibinfo{person}{Nick Barnes}.} \bibinfo{year}{2018}\natexlab{}.
\newblock \showarticletitle{Deep texture and structure aware filtering network for image smoothing}. In \bibinfo{booktitle}{\emph{ECCV}}. \bibinfo{pages}{217--233}.
\newblock


\bibitem[Paris and Durand(2006)]%
        {paris2006fast}
\bibfield{author}{\bibinfo{person}{Sylvain Paris} {and} \bibinfo{person}{Fr{\'e}do Durand}.} \bibinfo{year}{2006}\natexlab{}.
\newblock \showarticletitle{A fast approximation of the bilateral filter using a signal processing approach}. In \bibinfo{booktitle}{\emph{ECCV}}. \bibinfo{pages}{568--580}.
\newblock


\bibitem[Paris et~al\mbox{.}(2011)]%
        {paris2011local}
\bibfield{author}{\bibinfo{person}{Sylvain Paris}, \bibinfo{person}{Samuel~W Hasinoff}, {and} \bibinfo{person}{Jan Kautz}.} \bibinfo{year}{2011}\natexlab{}.
\newblock \showarticletitle{Local laplacian filters: edge-aware image processing with a laplacian pyramid.}
\newblock \bibinfo{journal}{\emph{ACM Transactions on Graphics}} \bibinfo{volume}{30}, \bibinfo{number}{4} (\bibinfo{year}{2011}), \bibinfo{pages}{68}.
\newblock


\bibitem[Perona and Malik(1990)]%
        {perona1990scale}
\bibfield{author}{\bibinfo{person}{Pietro Perona} {and} \bibinfo{person}{Jitendra Malik}.} \bibinfo{year}{1990}\natexlab{}.
\newblock \showarticletitle{Scale-space and edge detection using anisotropic diffusion}.
\newblock \bibinfo{journal}{\emph{IEEE Transactions on Pattern Analysis and Machine Intelligence}} \bibinfo{volume}{12}, \bibinfo{number}{7} (\bibinfo{year}{1990}), \bibinfo{pages}{629--639}.
\newblock


\bibitem[Subr et~al\mbox{.}(2009)]%
        {subr2009edge}
\bibfield{author}{\bibinfo{person}{Kartic Subr}, \bibinfo{person}{Cyril Soler}, {and} \bibinfo{person}{Fr{\'e}do Durand}.} \bibinfo{year}{2009}\natexlab{}.
\newblock \showarticletitle{Edge-preserving multiscale image decomposition based on local extrema}.
\newblock \bibinfo{journal}{\emph{ACM Transactions on Graphics}} \bibinfo{volume}{28}, \bibinfo{number}{5} (\bibinfo{year}{2009}), \bibinfo{pages}{1--9}.
\newblock


\bibitem[Team(2025)]%
        {team2025zimage}
\bibfield{author}{\bibinfo{person}{Z-Image Team}.} \bibinfo{year}{2025}\natexlab{}.
\newblock \showarticletitle{Z-Image: An Efficient Image Generation Foundation Model with Single-Stream Diffusion Transformer}.
\newblock \bibinfo{journal}{\emph{arXiv preprint arXiv:2511.22699}} (\bibinfo{year}{2025}).
\newblock


\bibitem[Tomasi and Manduchi(1998)]%
        {tomasi1998bilateral}
\bibfield{author}{\bibinfo{person}{Carlo Tomasi} {and} \bibinfo{person}{Roberto Manduchi}.} \bibinfo{year}{1998}\natexlab{}.
\newblock \showarticletitle{Bilateral Filtering for Gray and Color Images}. In \bibinfo{booktitle}{\emph{ICCV}}. \bibinfo{pages}{839--839}.
\newblock


\bibitem[Vipshop(2025)]%
        {cache-dit}
\bibfield{author}{\bibinfo{person}{Vipshop}.} \bibinfo{year}{2025}\natexlab{}.
\newblock \bibinfo{title}{Cache-DiT: A PyTorch-native Inference Engine with Hybrid Cache Acceleration and Massive Parallelism for DiTs.}
\newblock
\urldef\tempurl%
\url{https://github.com/vipshop/cache-dit.git}
\showURL{%
\tempurl}


\bibitem[Wang et~al\mbox{.}(2021)]%
        {wang2021real}
\bibfield{author}{\bibinfo{person}{Xintao Wang}, \bibinfo{person}{Liangbin Xie}, \bibinfo{person}{Chao Dong}, {and} \bibinfo{person}{Ying Shan}.} \bibinfo{year}{2021}\natexlab{}.
\newblock \showarticletitle{Real-esrgan: Training real-world blind super-resolution with pure synthetic data}. In \bibinfo{booktitle}{\emph{ICCV}}. \bibinfo{pages}{1905--1914}.
\newblock


\bibitem[Wang et~al\mbox{.}(2004)]%
        {wang2004image}
\bibfield{author}{\bibinfo{person}{Zhou Wang}, \bibinfo{person}{Alan~C Bovik}, \bibinfo{person}{Hamid~R Sheikh}, {and} \bibinfo{person}{Eero~P Simoncelli}.} \bibinfo{year}{2004}\natexlab{}.
\newblock \showarticletitle{Image quality assessment: from error visibility to structural similarity}.
\newblock \bibinfo{journal}{\emph{IEEE Transactions on Image Processing}} \bibinfo{volume}{13}, \bibinfo{number}{4} (\bibinfo{year}{2004}), \bibinfo{pages}{600--612}.
\newblock


\bibitem[Weiss(2006)]%
        {weiss2006fast}
\bibfield{author}{\bibinfo{person}{Ben Weiss}.} \bibinfo{year}{2006}\natexlab{}.
\newblock \showarticletitle{Fast median and bilateral filtering}.
\newblock \bibinfo{journal}{\emph{ACM Transactions on Graphics}} \bibinfo{volume}{25}, \bibinfo{number}{3} (\bibinfo{year}{2006}), \bibinfo{pages}{519--526}.
\newblock


\bibitem[Wu et~al\mbox{.}(2025)]%
        {wu2025qwenimagetechnicalreport}
\bibfield{author}{\bibinfo{person}{Chenfei Wu}, \bibinfo{person}{Jiahao Li}, \bibinfo{person}{Jingren Zhou}, \bibinfo{person}{Junyang Lin}, \bibinfo{person}{Kaiyuan Gao}, \bibinfo{person}{Kun Yan}, \bibinfo{person}{Sheng ming Yin}, \bibinfo{person}{Shuai Bai}, \bibinfo{person}{Xiao Xu}, \bibinfo{person}{Yilei Chen}, \bibinfo{person}{Yuxiang Chen}, \bibinfo{person}{Zecheng Tang}, \bibinfo{person}{Zekai Zhang}, \bibinfo{person}{Zhengyi Wang}, \bibinfo{person}{An Yang}, \bibinfo{person}{Bowen Yu}, \bibinfo{person}{Chen Cheng}, \bibinfo{person}{Dayiheng Liu}, \bibinfo{person}{Deqing Li}, \bibinfo{person}{Hang Zhang}, \bibinfo{person}{Hao Meng}, \bibinfo{person}{Hu Wei}, \bibinfo{person}{Jingyuan Ni}, \bibinfo{person}{Kai Chen}, \bibinfo{person}{Kuan Cao}, \bibinfo{person}{Liang Peng}, \bibinfo{person}{Lin Qu}, \bibinfo{person}{Minggang Wu}, \bibinfo{person}{Peng Wang}, \bibinfo{person}{Shuting Yu}, \bibinfo{person}{Tingkun Wen}, \bibinfo{person}{Wensen Feng}, \bibinfo{person}{Xiaoxiao Xu}, \bibinfo{person}{Yi
  Wang}, \bibinfo{person}{Yichang Zhang}, \bibinfo{person}{Yongqiang Zhu}, \bibinfo{person}{Yujia Wu}, \bibinfo{person}{Yuxuan Cai}, {and} \bibinfo{person}{Zenan Liu}.} \bibinfo{year}{2025}\natexlab{}.
\newblock \bibinfo{title}{Qwen-Image Technical Report}.
\newblock
\showeprint[arxiv]{2508.02324}~[cs.CV]
\urldef\tempurl%
\url{https://arxiv.org/abs/2508.02324}
\showURL{%
\tempurl}


\bibitem[Xu et~al\mbox{.}(2011)]%
        {xu2011image}
\bibfield{author}{\bibinfo{person}{Li Xu}, \bibinfo{person}{Cewu Lu}, \bibinfo{person}{Yi Xu}, {and} \bibinfo{person}{Jiaya Jia}.} \bibinfo{year}{2011}\natexlab{}.
\newblock \showarticletitle{Image smoothing via {L0} gradient minimization}.
\newblock \bibinfo{journal}{\emph{ACM Transactions on Graphics}} \bibinfo{volume}{30}, \bibinfo{number}{6} (\bibinfo{year}{2011}), \bibinfo{pages}{1--12}.
\newblock


\bibitem[Xu et~al\mbox{.}(2015)]%
        {xu2015deep}
\bibfield{author}{\bibinfo{person}{Li Xu}, \bibinfo{person}{Jimmy Ren}, \bibinfo{person}{Qiong Yan}, \bibinfo{person}{Renjie Liao}, {and} \bibinfo{person}{Jiaya Jia}.} \bibinfo{year}{2015}\natexlab{}.
\newblock \showarticletitle{Deep edge-aware filters}. In \bibinfo{booktitle}{\emph{ICML}}. \bibinfo{pages}{1669--1678}.
\newblock


\bibitem[Xu et~al\mbox{.}(2012)]%
        {xu2012structure-rtv}
\bibfield{author}{\bibinfo{person}{Li Xu}, \bibinfo{person}{Qiong Yan}, \bibinfo{person}{Yang Xia}, {and} \bibinfo{person}{Jiaya Jia}.} \bibinfo{year}{2012}\natexlab{}.
\newblock \showarticletitle{Structure extraction from texture via relative total variation}.
\newblock \bibinfo{journal}{\emph{ACM transactions on graphics (TOG)}} \bibinfo{volume}{31}, \bibinfo{number}{6} (\bibinfo{year}{2012}), \bibinfo{pages}{1--10}.
\newblock


\bibitem[Yin et~al\mbox{.}(2024)]%
        {yin2024one}
\bibfield{author}{\bibinfo{person}{Tianwei Yin}, \bibinfo{person}{Micha{\"e}l Gharbi}, \bibinfo{person}{Richard Zhang}, \bibinfo{person}{Eli Shechtman}, \bibinfo{person}{Fredo Durand}, \bibinfo{person}{William~T Freeman}, {and} \bibinfo{person}{Taesung Park}.} \bibinfo{year}{2024}\natexlab{}.
\newblock \showarticletitle{One-step diffusion with distribution matching distillation}. In \bibinfo{booktitle}{\emph{CVPR}}. \bibinfo{pages}{6613--6623}.
\newblock


\bibitem[Zhang et~al\mbox{.}(2015)]%
        {zhang2015segment}
\bibfield{author}{\bibinfo{person}{Feihu Zhang}, \bibinfo{person}{Longquan Dai}, \bibinfo{person}{Shiming Xiang}, {and} \bibinfo{person}{Xiaopeng Zhang}.} \bibinfo{year}{2015}\natexlab{}.
\newblock \showarticletitle{Segment graph based image filtering: fast structure-preserving smoothing}. In \bibinfo{booktitle}{\emph{ICCV}}. \bibinfo{pages}{361--369}.
\newblock


\bibitem[Zhang et~al\mbox{.}(2023)]%
        {zhang2023pyramid}
\bibfield{author}{\bibinfo{person}{Qing Zhang}, \bibinfo{person}{Hao Jiang}, \bibinfo{person}{Yongwei Nie}, {and} \bibinfo{person}{Wei-Shi Zheng}.} \bibinfo{year}{2023}\natexlab{}.
\newblock \showarticletitle{Pyramid Texture Filtering}.
\newblock \bibinfo{journal}{\emph{ACM Transactions on Graphics (Proceedings of ACM SIGGRAPH 2023)}} \bibinfo{volume}{42}, \bibinfo{number}{4} (\bibinfo{year}{2023}), \bibinfo{pages}{1--11}.
\newblock


\bibitem[Zhang et~al\mbox{.}(2014)]%
        {zhang2014rolling}
\bibfield{author}{\bibinfo{person}{Qi Zhang}, \bibinfo{person}{Xiaoyong Shen}, \bibinfo{person}{Li Xu}, {and} \bibinfo{person}{Jiaya Jia}.} \bibinfo{year}{2014}\natexlab{}.
\newblock \showarticletitle{Rolling guidance filter}. In \bibinfo{booktitle}{\emph{ECCV}}. \bibinfo{pages}{815--830}.
\newblock


\bibitem[Zheng et~al\mbox{.}(2025)]%
        {zheng2025diffusionnft}
\bibfield{author}{\bibinfo{person}{Kaiwen Zheng}, \bibinfo{person}{Huayu Chen}, \bibinfo{person}{Haotian Ye}, \bibinfo{person}{Haoxiang Wang}, \bibinfo{person}{Qinsheng Zhang}, \bibinfo{person}{Kai Jiang}, \bibinfo{person}{Hang Su}, \bibinfo{person}{Stefano Ermon}, \bibinfo{person}{Jun Zhu}, {and} \bibinfo{person}{Ming-Yu Liu}.} \bibinfo{year}{2025}\natexlab{}.
\newblock \showarticletitle{Diffusionnft: Online diffusion reinforcement with forward process}.
\newblock \bibinfo{journal}{\emph{arXiv preprint arXiv:2509.16117}} (\bibinfo{year}{2025}).
\newblock


\end{thebibliography}

\end{document}